\documentclass[pmlr,twocolumn,10pt]{jmlr} 

\usepackage{booktabs}
\usepackage{siunitx}

\usepackage[switch]{lineno}

\usepackage{hyperref}       
\usepackage{url}            
\usepackage{booktabs}       
\usepackage{amsfonts}       
\usepackage{nicefrac}       
\usepackage{microtype}      
\usepackage{lipsum}
\usepackage{fancyhdr}       
\usepackage{graphicx}       
\graphicspath{{media/}}     
\usepackage{booktabs}
\usepackage{rotating}
\usepackage{graphicx}
\usepackage{listings}
\usepackage{xcolor}

\usepackage{amsmath}
\usepackage{csquotes} 
\usepackage{comment}
\usepackage{natbib}
\usepackage{hyperref}
\usepackage[capitalize]{cleveref}
\usepackage{pgfplotstable}
\usepackage{booktabs}
\usepackage{longtable}
\usepackage{array}
\usepackage{tikz}
\usetikzlibrary{positioning}
\usepackage{soul}

\usepackage{booktabs}
\usepackage{csvsimple}
\usepackage{array}
\usepackage{siunitx}

\usepackage{amssymb,amsmath,amsfonts,mathtools,dsfont,thmtools}

\pgfplotsset{compat=1.18}

\usepackage{shortcuts}

\theorembodyfont{\upshape}
\theoremheaderfont{\scshape}
\theorempostheader{:}
\theoremsep{\newline}

\jmlrvolume{333}
\jmlryear{2026}
\jmlrsubmitted{LEAVE UNSET}
\jmlrpublished{LEAVE UNSET}
\jmlrworkshop{Conference on Health, Inference, and Learning (CHIL) 2026} 

\title{Revisiting Performance Claims for Chest X-Ray Models \\
 \ Using Clinical Context}

\author{%
 \Name{Andrew Wang} \Email{andrew\_wang3@brown.edu}\\
 \addr Brown University
 \AND
 \Name{Jiashuo Zhang} \Email{jzhan427@jhu.edu}\\
 \addr Johns Hopkins University
 \AND
 \Name{Michael Oberst} \Email{moberst@jhu.edu}\\
 \addr Johns Hopkins University
}

\begin{document}

\maketitle

\begin{abstract}
Public datasets of Chest X-Rays (CXRs) have long been a popular benchmark for developing machine learning (ML) computer vision models in healthcare. However, the reported strong average-case performance of these models do not necessarily reflect their actual utility when used in heterogeneous clinical settings, potentially masking weaker performance in medically significant scenarios. In this work we use clinical context to provide a more holistic evaluation of models for CXR diagnosis. In particular, we use discharge summaries, recorded prior to each CXR, to derive a ``pre-CXR'' probability of each CXR label, as a proxy for existing contextual knowledge available to clinicians when interpreting CXRs.  
We use this measure to probe model performance along two dimensions:  First, using a stratified analysis, we show that models tend to have lower performance (as measured by AUROC and other metrics) among individuals with higher pre-CXR probability.  Second, by controlling for pre-CXR probability via matching and re-weighting, we demonstrate that performance degrades when the correlation is broken between prior context and the current CXR label, suggesting that model performance is highly sensitive to the underlying distribution of clinical context. Specifically, cases with high pre-test probabilities present a fundamentally more difficult visual classification task, highlighting a gap in clinical utility when models are applied to high-risk cohorts.
\end{abstract}

\paragraph*{Data and Code Availability}
We use data from the MIMIC-CXR \citep{Johnson2019-0c} and MIMIC-IV \citep{Johnson2023_MIMIC-IV} datasets, which are available on \texttt{physionet.org} via credentialed access under a data-use agreement. Code is publicly available at \url{https://github.com/oberst-lab/revisiting-cxr-performance}. 

\paragraph*{Institutional Review Board (IRB)}
This study uses de-identified, publicly available datasets under the PhysioNet Credentialed Health Data Use Agreement 1.5.0. Institutional Review Board (IRB) approval was not required for this study.

\section{Introduction}
\label{sec:intro}
Machine learning (ML) systems have shown impressive results in disease diagnosis from medical images, including on large public Chest X-Ray (CXR) datasets like MIMIC-CXR \citep{Johnson2019-0c}.  However, average-case evaluations do not fully capture the clinical utility of a model’s predictions.  First, these aggregate evaluations may mask differences in relevant subpopulations \citep{Chexclusion, Oakden-Rayner2020-06}, and clinical end-users of model predictions often need to understand the populations where a model should (or should not) be expected to perform well.
Second, even on a well-defined subpopulation, standard ``AI-alone'' performance metrics may fail to capture the intrinsic diagnostic ability of models, and their ability to complement human decision-makers. Radiologists may not interpret CXRs in isolation, but also consider patient history and other contextual details. The clinical utility of a CXR model thus depends in part on its ability to rely on direct visual indicators of disease, rather than inferring prior clinical context already known to the clinician. For instance, \citet{Badgeley2019-ab} found that the performance of a hip-fracture detection model was no better than random, once factors like scanner type and manufacturer were controlled for. Similarly, \citet{Oakden-Rayner2020-06} show that apparent high performance at pneumothorax detection can be driven in part by high performance on cases where a chest drain is already present, a potentially less salient clinical population where the relevant condition is already being treated.

\begin{figure*}[t]
	\centering
	\includegraphics[width=0.9\linewidth]{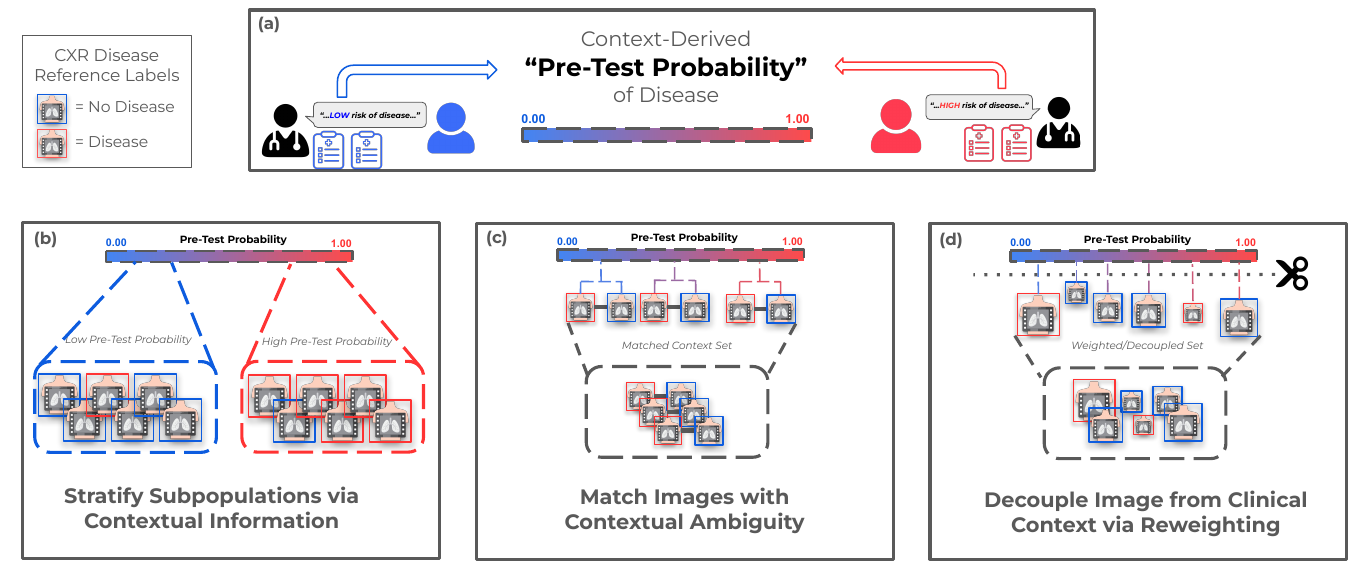}
	\caption{Overview of our evaluation framework
    (a): We use clinical context contained within discharge summaries to derive a pre-test probability estimate of disease risk, given knowledge obtained before a CXR is ordered.
    (b): We identify subpopulations of the evaluation using context from prior notes (such as pre-test probability, or prior mention of the disease label). We then evaluate performance on the resulting disjoint groups.
    (c): We create an evaluation set by matching positive/negative image pairs with similar context-derived pre-test probabilities. We then compare vision model performance on this balanced test set versus the original test set
(d): We statistically decouple the image label from the contextual information via reweighting the evaluation set.}
	\label{fig:overview}
\end{figure*}

In this paper, we use prior clinical notes as a proxy for the ``clinical context'' of a diagnostic task, as captured by the ``pre-test'' or ``pre-CXR'' probability, before the CXR is actually taken, of each CXR disease label.  We argue that this contextual information provides insights into the clinical utility of CXR models.  Throughout, we use the MIMIC-CXR / MIMIC-IV datasets \citep{Johnson2019-0c,Johnson2023_MIMIC-IV} as a case study, though we expect that our approach could be broadly applied on any longitudinal health records dataset containing both medical images and prior clinical notes.

First, we show that prior discharge summaries (from prior hospital admissions) often contain sufficient information to predict disease labels in CXRs from future visits, even without access to the images themselves. We find evidence that this predictive signal can be derived (at least in part) from medically relevant terms in prior notes (e.g., the words ``clavicle'' and ``rib'' have high importance in predicting the label of ``fracture'' on future CXRs). 

Second, we find that vision model performance varies significantly across different levels of the resulting pre-test probability derived from prior notes.  The CXR models we study 
show significant degradation in performance (as measured by AUROC and other metrics) for those individuals with a higher prior probability of disease. A similar divergence in performance appears with a simple stratification of patients into those with and without an explicit prior mention of certain disease-relevant phrases. These results suggest that prior clinical text is a useful axis along which to understand variation in CXR model performance. 

Finally, we consider performance of CXR models when controlling for pre-test probability.  Our analysis is motivated by the existing work that has shown that vision models can recover surprising amounts of contextual information (e.g., scanner type \citep{Badgeley2019-ab}, patient demographics \citep{GICHOYA2022e406}, etc) from images, even if this information may not be readily distinguishable by humans. As such, it is not always clear whether strong performance arises from clinically relevant visual indicators of disease, or from information that is redundant between the context and the image itself.  We probe this dependence in two ways: First, we demonstrate that model performance degrades substantially on a matched population subset where positive and negative examples are selected to have near-identical values of pre-test probability. Second, we demonstrate that performance degrades, though to a lesser degree, on a re-weighted population where the context-derived pre-test probability is rendered marginally independent of the label.

Taken together, these results suggest that model performance on standard CXR tasks is dependent on clinical context:  Stratified analysis reveals under-performance in those cases where prior knowledge suggests higher risk of diagnosis, and matching / reweighting suggest that model performance may depend in part on inference of pre-CXR clinical context,
although our analysis cannot rule out other causes, such as high intrinsic difficulty of distinguishing positive and negative high-risk cases.
Our framework is summarized in \cref{fig:overview}.

To summarize, our contributions are as follows 
\begin{itemize}
	\item{We demonstrate the ability to accurately predict future CXR disease labels using prior clinical notes, providing a ``pre-test probability'' that approximates the pre-imaging disease risk.}
	\item{We identify performance disparities of standard clinical vision models across subpopulations defined by the context-derived pre-test probability, and across subpopulations defined by prior mentions of medically relevant terms.}
    \item{We investigate the relationship between visual diagnostic signal and clinical context using a reweighted and context-matched evaluation framework, finding degradation in performance under both analyses, with degradation most severe in the matched evaluation.}
\end{itemize}

\section{Related Work}
 Recent studies have explored various approaches to improve medical imaging models and their evaluation using clinical context, though none address the specific challenge of characterizing model performance via prior medical history.  \citet{Juodelyte2023} systematically analyzes dataset confounders in CXR, while \citet{Olesen2024} develops slice discovery methods to expose biased subpopulations, but neither incorporates clinical notes as stratification signals. Closest to our focus, \citet{Saab2024} demonstrates the value of workflow notes for EEG subgroup analysis, and \citet{Yang2025} reveals persistent demographic biases in vision-language models, but neither use longitudinal medical context to explain failures. Similarly, \citet{Subbaswamy2024} automates under-performance detection but requires manual feature engineering in their experiments, and does not make use of medical context as a stratifying feature in the evaluation set. On the model development side, several works incorporate radiological notes for label refinements (\citet{SyedaMahmood2021, Kim2022}) or model training (\citet{mhf2023, Monajatipoor2021}), but treat these texts as auxiliary inputs rather than tools to interrogate performance.  Closest to our perspective, \citet{Badgeley2019-ab} showed that non-clinical correlates can inflate apparent performance, highlighting the need to disentangle true signal from contextual shortcuts when evaluating diagnostic models. Similarly, \citet{GICHOYA2022e406} illustrates the ability of vision models to recover patient demographic information from images, while \citet{makar2022causallymotivatedshortcutremoval} proposed enforcing conditional independence to reduce reliance on correlated features in ML classifiers. However, none of these works directly addresses the correlations between clinical context and imaging and its impact on vision model performance, which is the central focus of our study.

\section{Data and Experimental Setup}

\subsection{Dataset Construction \& Vision Model Training}
To construct a cohort of CXR studies with prior context, we made use of the MIMIC-CXR database, which includes CXR studies with structured diagnosis labels from the CheXpert labeling tool for 14 different labels. We do not report the ‘No Finding’ label metrics due to lack of relevance in our downstream evaluation.\footnote{We excluded the ``No Finding'' label from analysis as many of our experiments (e.g., stratifying based on prior mentions of a disease label) do not make sense in the context of the ``No Finding'' label.}  The MIMIC-CXR dataset provides linkages to the MIMIC-IV dataset, which we used to identify prior admissions for each patient, and pull the corresponding discharge summaries for those admissions.  Further details are given in~\cref{sec:Experimental Details}.

\begin{figure}
	\centering
	\includegraphics[width=0.9\linewidth]{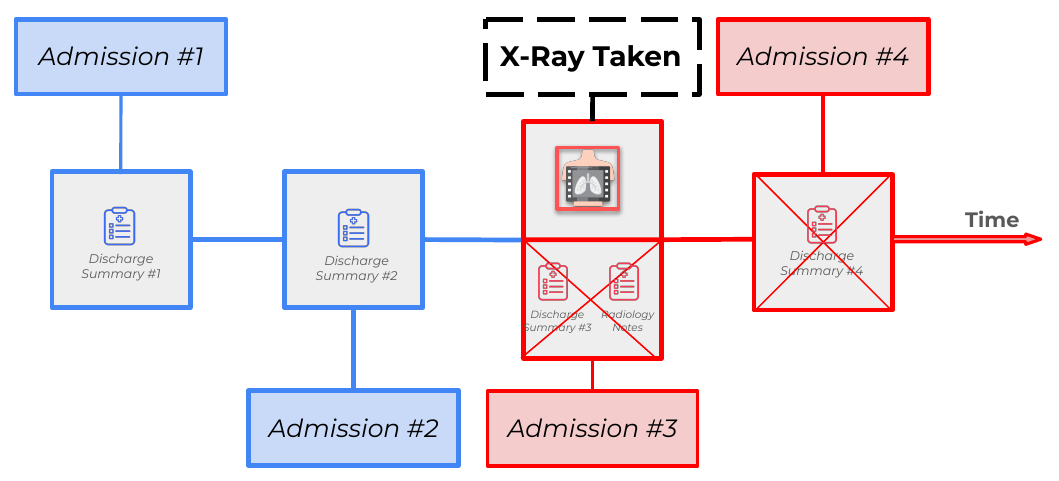}
	\caption{Visualization of ``prior clinical context'' used in this paper.  For a given CXR study, we use all discharge summaries from admissions that occur before the current admission (blue). We do not use any information (discharge summaries, radiology reports, or otherwise) from the current or any future admission (red).}
	\label{fig:data_processing}
\end{figure}

We consider a single instance of the multilabel classification problem to be predicting labels for a single chest CXR study, where it is possible for multiple labels to be positive for a given CXR. As such, we defined prior context for a study as all of the associated ER visits and associated clinical notes whose discharge times and chart times, respectively, occur before the time of the CXR study (\cref{fig:data_processing}). We also restricted to individuals that had prior admissions and discharge summaries, in order to construct an evaluation set where all images had associated prior clinical context.  We defined all CXR labels not explicitly corresponding to a positive diagnosis (\enquote{no label}, \enquote{uncertain}, among others) as a negative diagnosis in our label processing, following the procedure in \citet{Seyyed-Kalantari2020-97}. To prevent data leakage, we performed an 80-10-10 split by subject ID, ensuring that no patient’s images or notes appeared across multiple splits. This yielded 167,291 images for training, 21,239 for validation, and 20,770 for testing. Full details of pre-processing, database joins, and methods to train the vision model used for experiments are provided in~\cref{sec:Experimental Details}. We broadly observed identical performance trends across both DenseNet121 and ResNet50 backbones, and all subsequent results in the main paper and Appendix are presented using the DenseNet121 backbone, unless otherwise specified.

\subsection{Quantifying Pre-Test Probability from Prior Notes}
\label{subsec:Clinical}

\paragraph{Training Classifiers using Language Model (LM) Embeddings} To capture the clinical context contained in prior notes, we evaluated a diverse set of text representations and classification heads. We first embedded all prior notes (after basic preprocessing) using six distinct open-weight language models: Mistral-7B-v0.1 (\citet{jiang2023mistral7b}), PubMedBERT (\citet{Gu_2021}), BERT (\citet{devlin2019bertpretrainingdeepbidirectional}), ClinicalBERT (\citet{huang2020clinicalbertmodelingclinicalnotes}), BioLinkBERT (\citet{yasunaga2022linkbert}), and RoBERTa (\citet{DBLP:journals/corr/abs-1907-11692}). Treating these embeddings as fixed feature representations, we trained classifiers to predict the image label from the text of prior discharge summaries alone. We performed a  hyperparameter sweep using 5-fold stratified group cross-validation to select the optimal model class and hyperparameters from a set of 11 standard classification algorithms in \texttt{sklearn}. The \enquote{subject\_id} was designated as the grouping variable to ensure that all data from a single patient resided within the same fold during cross-validation, preventing patient-level data leakage between folds. For each label, we selected the combination of language model encoder, classification architecture, and hyperparameters that yielded the highest mean cross-validation AUROC. Finally, the selected classifiers were calibrated using Platt scaling (sigmoid) prior to downstream analysis. See \cref{sec:Experimental Details} for full details on text pre-processing, embedding extraction, the hyperparameter search space, and the best performing configurations.

\paragraph{Interpreting Predictive Signal with Bag of Words (BoW) Classifiers}\label{subsec:Interpretable Bag of Words Text Classifiers}
In addition to the models described above, which used language-model embeddings, we also constructed a bag-of-words (BoW) representation from these notes using \texttt{CountVectorizer} in \texttt{sklearn}, and evaluated seven standard classifiers (full training details in \cref{tab:bow_classifiers}, \cref{tab:bow_hp_configs}), chosen for their interpretable features / coefficients, conducting hyperparameter and model selection as described above. 
Following the selection of the best classifier, we examined the top 10 most important features via feature importance measures specific to each model class (full list of feature importances in \cref{tab:feature_importances}).
\\
\\
Finally, we also trained XGBoost models on these BoW representations as well. Clinical text was pre-processed and tokenized in the same manner as in \ref{sec:appendix_text}. Model hyperparameters were optimized through a 60-iteration randomized search using 5-fold Stratified Group K-Fold cross-validation, preventing data leakage across splits. Finally, to ensure reliable pre-test probability estimates, the best-performing model from the cross-validation phase was post-hoc calibrated using Isotonic Regression. 

\subsection{Stratified Performance Analyses}
\label{subsec:Stratified and Matched Analyses}

\paragraph{Stratification via Pre-Test Probability}
\label{subsec:Stratification Experimental Setup}

Using the classification models trained on prior text, we obtained the predicted probabilities for each CXR label, for each example in our evaluation set. These label predictions were then binned by quantiles (bottom $25\%$, middle $50\%$, top $25\%$) to form disjoint subpopulations in the evaluation set.

\paragraph{Stratification via Prior Mentions}
\label{subsec:prior_mentions}
 As an alternative to using pre-test probability, we utilized the phrase list for each label from CheXpert \citep{irvin2019chexpert} to define prior mentions of the label in prior clinical notes.  After adapting their phrase list to our context\footnote{Since the CheXpert labeler was specifically designed for radiology reports, some phrases in their list were not directly applicable to the clinical notes in our study. For example, for the ``Pneumonia" label, the original corresponding phases included ``infection" and ``infectious", but these terms are commonly used in clinical notes to describe other infectious diseases and do not demonstrate a strong correlation with pneumonia specifically. We therefore adapted and refined selected terms to ensure precise identification of relevant mentions in our clinical context. Details of the specific phrases used are presented in \cref{tab:chexpert_phrases}.}, we defined disjoint populations based on the presence of any relevant term, and the absence of all relevant terms.

\begin{figure*}[t]
\centering
\includegraphics[width=0.8\linewidth]{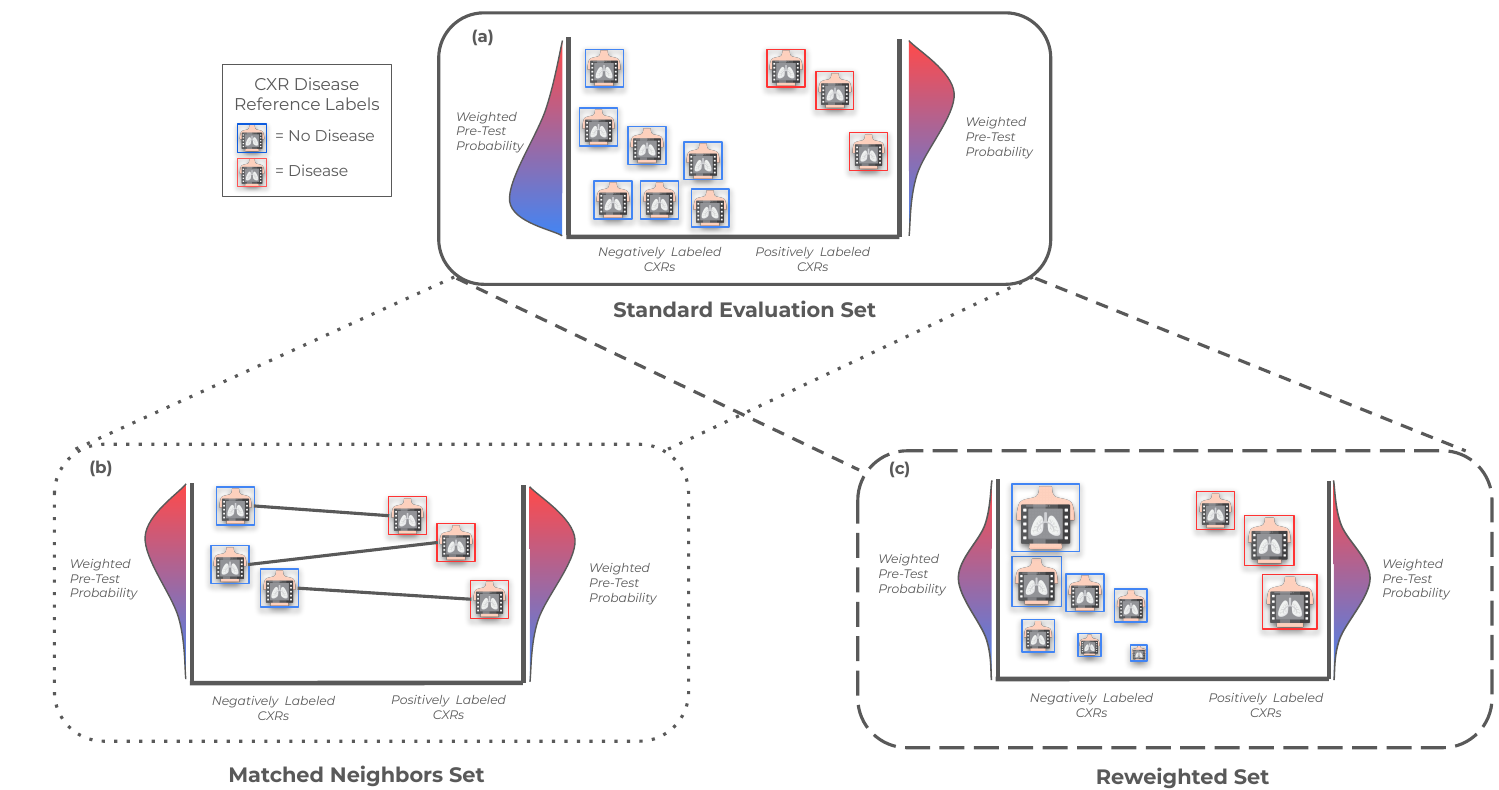}
	\caption{Details of Matched Neighbors vs Reweighted Evaluation Set
    (a): Original evaluation set:  Negatively and positively labeled CXRs in the original evaluation set with their associated distributions of pre-test probabilities. All images and their associated pre-test probabilities are equally weighted.
    (b): Matched Neighbors set:  Our procedure matches positive and negative examples 1-to-1 based on pre-test probability.  In practice, given low prevalence of positive labels, this procedure tends to retain the subset of negatively labeled CXRs whose associated pre-test probability distribution more closely resembles that of the positively labeled CXRs, which are unchanged. 
    (c): Reweighted set:  After reweighting of images and their associated pre-test probabilities, the weighted distribution of pre-test probabilities become comparable across the positive and negative reweighted populations.}
\label{fig:matching-eval}
\end{figure*}

\subsection{Controlling for Pre-Test Probability}\label{subsec:controlling_context}
To assess the extent to which vision models depend on inference of pre-CXR context, rather than direct visual cues, we develop two complementary evaluation strategies to control for pre-test probability:
\paragraph{Matched Neighbor Analysis}\label{par:methods_matching}We constructed a context-balanced test set by selecting pairs of positive and negative examples (as determined by their ground truth labels) with similar pre-test probabilities. 
Positive (label = 1) and negative (label = 0) examples were separated, and the Hungarian algorithm applied (via \texttt{linear\_sum\_assignment} in SciPy) to perform 1:1 nearest-neighbor matching between positive and negative examples. The cost matrix was defined by the absolute difference between the pre-test probabilities of each positive–negative pair. This procedure yielded matched pairs with highly similar text-based probabilities but opposite ground truth labels. As illustrated in~\cref{fig:matching-eval}, this matched set represents a setting where the clinical context is balanced across class labels, making it impossible to predict the label by inferring prior clinical context alone. 
In practice, due to the limited number of positive cases, we found that this procedure tends to select all of the original positive examples, retaining only a subset (matched on pre-test probability) of the negative examples for each label.  Hence, in contrast to the method described below, the resulting distribution of pre-test probability is unchanged among cases where $Y = 1$, and is substantially modified among cases where $Y = 0$.

\paragraph{Inverse Probability Weighting (IPW)}
\label{par:methods_ipw}
To rigorously assess whether the vision model utilizes visual signals independent of clinical context, we estimate performance on a theoretical target distribution where diagnostic labels are statistically independent of the prior clinical context. We formalize this as follows:

\begin{definition}[Reweighting Distributions]
\label{def:target_dist}
Let $P(X, Y, C)$ denote the observed data distribution, where $X$ is the image, $Y \in \{0, 1\}$ is the diagnostic label, and $C$ is the pre-test probability derived from the prior clinical context. In the observational setting, $Y$ and $C$ are correlated. We define a target distribution $Q(X, Y, C)$ that satisfies two conditions:
\begin{enumerate}
\item \textbf{Independence of Context and Label:} $Q(Y, C) = Q(Y)Q(C) = P(Y)P(C)$. This represents a scenario where clinical history provides no information about the current diagnosis, but the marginal distributions of $Y$ and $C$ are unchanged.
\item \textbf{Invariance of Imaging Mechanism:} $Q(X \mid Y, C) = P(X \mid Y, C)$. Conditioned on context and the true label, the distribution of images is unchanged.  
\end{enumerate}

\end{definition}

We note that standard performance metrics evaluated on the target distribution $Q$ can be estimated via weighted metrics on the observed distribution $P$ (\citet{NIPS2007_be83ab3e}). We provide the formal derivation for the importance weights and the estimators for the performance metrics in Appendix \ref{sec:proofs}, but to provide intuition here, we note the following basic result, which is an application of standard methods in importance sampling to our context.
\begin{restatable}[Expected Equivalence]{proposition}{MetricEquiv}
\label{prop:metric_equiv}
For an observation $(x, y, c)$, let the importance weight $w(y, c)$ be defined as the ratio of the marginal disease prevalence to the context-conditional probability:
\begin{equation} 
\label{eq:weight_def}w(y, c) = \frac{P(Y=y)}{P(Y=y \mid C=c)}
\end{equation}
Then, given a distribution $P$ and a corresponding distribution $Q$ (as defined in Def.~\ref{def:target_dist}), where $Q(X, Y, C) > 0 \implies P(X, Y, C) > 0$, then for any measurable function $g(X, Y, C)$:
\begin{equation*}
E_P[w(y, c) g(X, Y, C)] = E_Q[g(X, Y, C)]
\end{equation*}
\end{restatable}
In other words, for any metric that can be expressed as an expectation over the population $Q$, we can re-express that metric as a reweighted average over the population $P$.  In Appendix~\ref{sec:proofs} we illustrate how this result can be used to derive re-weighting estimators for  our performance metrics of interest (Sensitivity, Specificity, AUROC, ACE, and BSS).

\paragraph{Implementation}
In practice, we estimate the weights using the training set and the text-only classifiers. The marginal prevalence $\hat{P}(Y=1)$ is estimated from the training set distribution. The conditional probability $\hat{P}(Y=1 \mid C=c_i)$ is obtained from the calibrated text-only classifier described in Section \ref{subsec:Clinical}. To prevent numerical instability from extreme weights, text-derived probabilities are clipped to the range $[0.001, 0.999]$ prior to weight calculation. Finally, the weights are normalized such that their sum equals the total number of samples in the evaluation set ($\sum w_i = N$). All weighted metrics are computed using the \texttt{sample\_weight} parameter in standard \texttt{scikit-learn} implementations.

\subsection{Metrics and Statistical Uncertainty}
We estimate performance on subpopulations with stratified bootstrapping. For each bootstrap iteration, we sample (with replacement) while maintaining the label distribution by stratifying on the ground truth labels. We calculate the area under the receiver operating characteristic curve (AUROC) for each bootstrap sample for each subgroup using the \texttt{MultilabelAUROC} metric across our 13 labels. The two metrics are subtracted to obtain our final difference in AUROC metric between the subgroups, and this process is repeated for a total of $10,000$ bootstrap iterations. Finally, we compute the mean difference and the corresponding $95\%$ confidence intervals (CI) using the 2.5th and 97.5th percentiles of the bootstrapped metric difference distribution, adjusted via Bonferroni correction to account for multiple comparisons across the 13 labels. 

While AUROC provides an aggregate measure of diagnostic performance, it may obscure model behavior at specific operating points critical for clinical decision-making. We therefore further introduce sensitivity at 95\% specificity: the true positive rate when the false positive rate is constrained to 5\%. 

To capture both the model's potential and its practical deployment risk, we evaluate this metric using both subgroup-specific thresholds, which reflect the model's intrinsic discriminative capacity, and a universal global threshold, which simulates clinical practice. For brevity, we denote these as \textbf{sensitivity at local 95\% specificity} and \textbf{sensitivity at global 95\% specificity}, respectively.

\section{Results}
\subsection{CXR Labels are Predictable from Prior Clinical Notes Alone}
\label{subsec:quantile}

We find that text classifiers trained on either LM or BoW representations are able to predict many of the MIMIC-CXR labels using only discharge summaries from prior to the CXR visit.  Performance of models using the best LM embeddings is given in \cref{table:text_performance_auroc}, with full results in \cref{table:text_performance}. \Cref{table:bow_performance} gives performance of models using BoW representations. We make three general observations: First, we note that extensive hyperparameter optimization across both types of representations resulted in similar test set performance for most of the labels. Second, the predictive signal in the prior context can be at least partially traced to medically relevant terms in the discharged summaries: words like ``clavicle'' and ``rib'' are assigned high feature importance in the Fracture classifiers trained on BoW embeddings (full list in \cref{table:feature_importances}). We reason that despite the presence of other noisier features such as ``completing'' and ``18'' that contribute to these predictions, the strong performance of both classes of text classifiers and relative importance of medically relevant terms provide orthogonal axes of evidence pointing towards the existence of medically relevant textual signal in the clinical context. Finally, we observe varying levels of text classifier calibration across different labels (refer to \cref{fig:text_calibration_plots} and \cref{table:text_performance_calibration}). 

 \begin{table}[t]
    \centering
    \caption{Discrimination performance of text-only classifiers. We report the test set AUROC for the best performing language model and classifier combination for each label, trained solely on prior clinical notes. We observe that prior context is generally sufficient to build longitudinal ML classifiers of future CXR labels. Refer to table \ref{table:bow_performance} for the full results on the BoW representations and \ref{table:xgboost_performance} for the full results on the XGBoost models.}
    \footnotesize 
    \resizebox{0.4\textwidth}{!}{%
        \begin{tabular}{lc}
            \toprule
            \textbf{Label} & \textbf{AUROC} \\
            \midrule
            Atelectasis & 0.623 \\
            Cardiomegaly & 0.701 \\
            Consolidation & 0.633 \\
            Edema & 0.751 \\
            Enlarged Cardiomediastinum & 0.566 \\
            Fracture & 0.676 \\
            Lung Lesion & 0.765 \\
            Lung Opacity & 0.628 \\
            Pleural Effusion & 0.745 \\
            Pleural Other & 0.654 \\
            Pneumonia & 0.592 \\
            Pneumothorax & 0.712 \\
            Support Devices & 0.682 \\
            \bottomrule
        \end{tabular}%
    }
    
    \label{table:text_performance_auroc}
\end{table}

\begin{figure*}[t]
\centering
\includegraphics[width=1.0\linewidth]{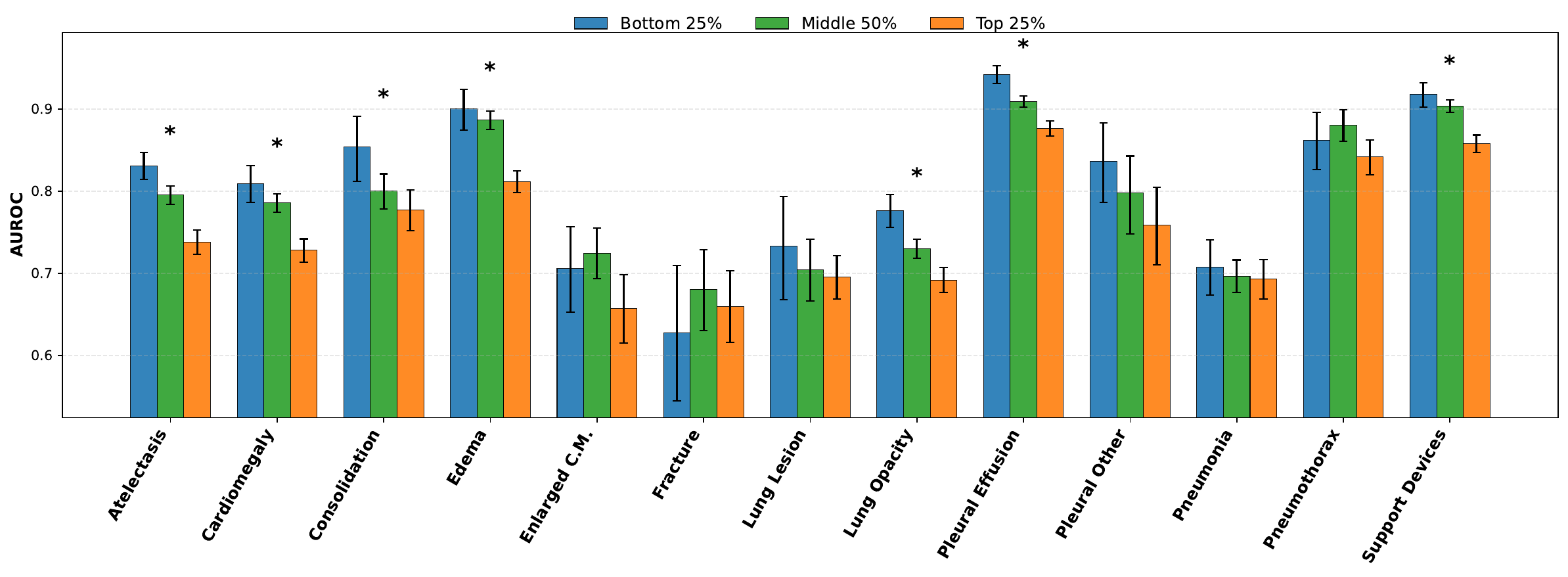}
\vspace{-25pt}
\caption{Held-out performance (in terms of \textbf{AUROC}) of CXR models across sub-populations stratified by pre-test probability (LM embeddings) of the CXR label. Vision model performance generally degrades across most labels as the prior probability of the label increases. Labels marked with an asterisk (*) indicate a statistically significant difference in AUROC between the Bottom 25\% and Top 25\% groups. The 95\% confidence intervals in parentheses were calculated using percentile bootstrapping as described in section \ref{subsec:Stratified and Matched Analyses}.}
\label{fig:quantile_lm_auroc}
\end{figure*}

\subsection{Vision Model Performance Varies Across Context-Derived Subpopulations}
\label{subsec:context-strat-results}
We observe that when stratifying our dataset based on the pre-test probabilities (based on prior notes) produced by our text classifiers, statistically significant performance gaps emerge across our subgroups across several labels, particularly when comparing the ``low risk'' (bottom 25\%) quantile against the ``higher risk'' (top 25\%) quantile of pre-test probabilities (see \cref{fig:quantile_lm_auroc,tab:quantile_lm_auroc} for results using the text classifiers trained on LM embeddings). Notably, images within the lower quantiles of the text classifier predictions tend to yield higher vision model performance, which decreases as we move to higher quantiles. For example, the vision model’s AUROC for Edema decreases from $0.9$ in the bottom ``low-risk'' quantile to $0.812$ in the top ``high-risk'' quantile, a statistically significant difference. Similarly, we observe that vision model performance for Atelectasis decreases from $0.831$ in the bottom quantile to $0.738$ in the top quantile. These examples indicate that vision models generally perform best on cases where the pre-test probability is low.

Complementing these findings, we observe a similar pattern of performance degradation in sensitivity at 95\% local specificity, where thresholds are set independently in each subpopulation to maintain 95\% specificity (see \cref{tab:quantile_lm_sens_local}). These results further reinforce the observation that vision model reliability at specific clinical operating points is significantly influenced by prior clinical context. However, this trend reverses when measuring sensitivity at 95\% global specificity, where a single global threshold is chosen (see \cref{tab:quantile_lm_sens_global}), and where both sensitivity and specificty can vary across quantiles.  In this setting, sensitivity is higher in the high-risk quantile, but as illustrated by the score distribution shifts for Edema and Atelectasis in \cref{fig:quantile_histograms}, this effect can be explained by an overall upward shift in model scores for high-risk patients (with a correspondingly lower specificity; see \cref{tab:quantile_lm_spec_local}), rather than improved discriminative performance. Consequently, applying a fixed global threshold can disproportionately limit sensitivity in low-risk cases: in clinical practice, the use of a single global threshold may systematically suppress performance in subgroups where the model’s discriminative potential is higher. We also repeated the above analyses and using BoW representations and observed similar patterns, as shown in \cref{tab:quantile_bow_auroc,tab:quantile_bow_sens_local,tab:quantile_bow_sens_global}. We also observe that the ResNet50 backbone provided similar performance trends, as indicated in \cref{tab:resnet50_quantile}. Finally, we provide representative examples of images whose discharge notes yield high and low pre-test probabilties in \cref{app:pre-test_ex}.

\subsection{Vision Model Performance Varies Across Cases with and without Prior Mentions}
\label{subsec:prior-mention-strat-results}

As described in~\cref{subsec:prior_mentions}, we considered a simpler stratification of cases based on the presence or absence of label-relevant medical terms in previous discharge summaries. 
We observe a consistent pattern in which vision models perform systematically better on images from patients without prior clinical mentions of the corresponding label. Specifically, across all labels, the ``no previous mentions'' group demonstrates higher AUROC values than the ``previous mentions" group, as shown in \cref{tab:prior_mention_auroc}. For example, the performance of the vision model for Atelectasis on the subset of images with a previous mention of the disease is $0.757$, but the performance of the same model on the subset of images without a previous mention is $0.831$. Similarly, the performance of the vision model for Cardiomegaly on the subset of images with a previous mention of the disease is $0.787$, but the performance is $0.819$ when evaluated on the images without this mention. 

These patterns persist for sensitivity at local 95\% specificity (\cref{tab:prior_mention_sens_local}), which remains consistently higher for cases without prior mentions. However, this advantage is obscured when measured via sensitivity at global 95\% specificity (\cref{tab:prior_mention_sens_global}). This contrast is similar with the pattern observed in \cref{subsec:quantile}: in clinical practice, the use of a single global threshold may systematically suppress performance in subgroups where the model’s discriminative potential is higher.

\begin{figure*}[t]
\centering
\includegraphics[width=1.0\linewidth]{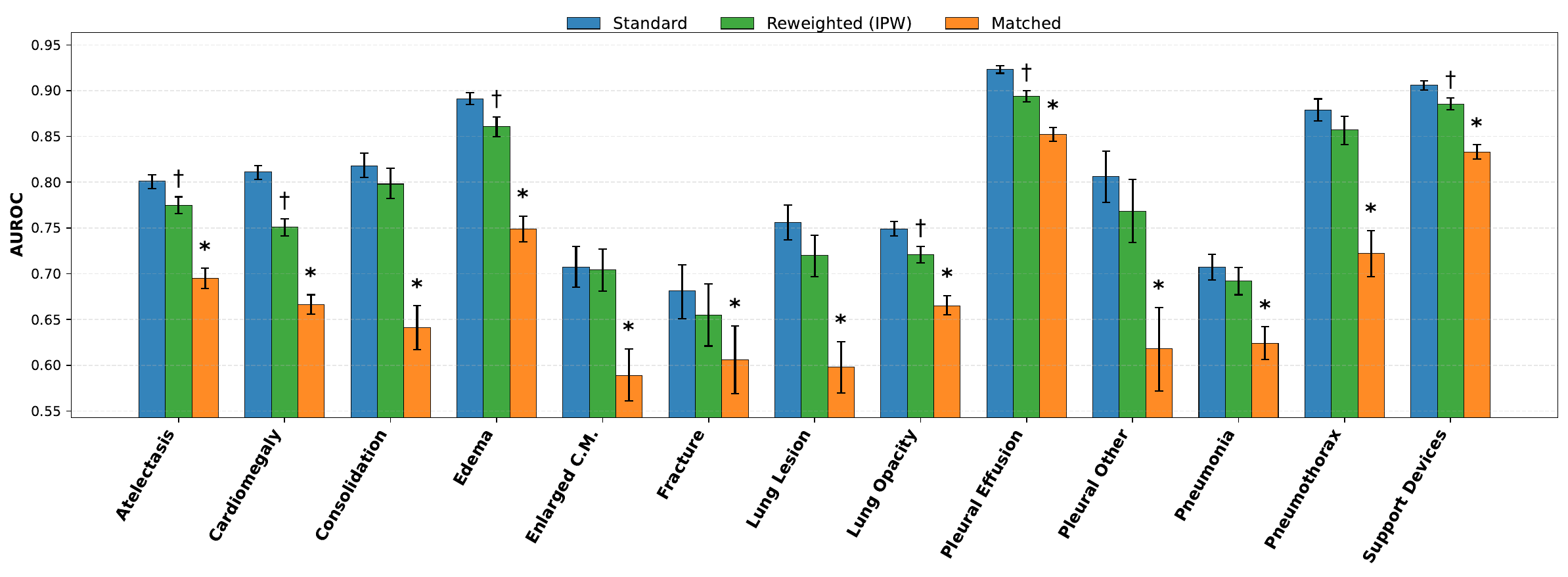}
\vspace{-25pt}
\caption{Comparison of \textbf{AUROC} across Standard, Reweighted (IPW), and Matched Neighbor settings. Metrics reported are mean (95\% CI). Labels marked with a dagger ($\dagger$) and an asterisk (*) indicate statistically significant differences in the Reweighted (IPW) and Matched settings, respectively, compared to the Standard setting. We observe that performance consistently drops across the Matched evaluation settings, and that vision model performance degrades across all labels in the Reweighted setting, though not all differences are statistically significant.}
\label{fig:reweighted_matched}
\end{figure*}

\subsection{Vision Model Performance Degrades when Controlling for Pre-Test probability} \label{subsec:matching-analysis-results}

We next employ two complementary strategies to control for the predictive signal already present in clinical notes: Inverse Probability Weighting (IPW) and Matched Neighbor analysis.

First, using IPW (\cref{par:methods_ipw}), we simulate a target population where diagnostic labels are statistically independent of clinical context, where ``surprising'' cases (e.g., low context-derived risk but positive diagnosis) are upweighted.   Here, \textbf{we find that vision model performance degrades across all labels, though not all differences are statistically significant}, with six out of 13 labels exhibiting statistically significant declines (\cref{tab:matched_neighbor_auroc}).  We note that the labels that do exhibit these difference correpsond to the labels whose associated text classifier predictions are comparably well calibrated. Given the reliance of these experiments on the text-derived pre-test probabilties, we posit that the relative levels of calibration in the text classifiers may be a factor in some labels exhibiting larger differences than others. 
For instance, while certain labels such as ``Atelectasis'' and ``Cardiomegaly'' display statistically significant differences between this reweighted set compared to the original, other labels such as ``Fracture'' and ``Lung Lesion'' do not appear to hold the same pattern.

Following the Matched Neighbor procedure (\cref{par:methods_matching}), we evaluated the model on a subset of paired positive and negative examples with effectively identical pre-test probabilities. \textbf{In this setting, we find that vision model performance drops across all labels in a statistically significant fashion.} For instance, the AUROC for Pleural Effusion falls from 0.92 in the standard set to 0.85 in the matched set, and the AUROC for Pneumothorax falls from 0.879 in the standard to 0.722 in the matched set (\cref{fig:reweighted_matched,tab:reweighted_matched}). 
We observe a parallel trend in our threshold-based and calibration metrics. Sensitivity at 95\% specificity (using subgroup-optimized thresholds) decreases notably in the matched evaluation set across all labels (see \cref{tab:sens_results} in the Appendix), and calibration metrics appear to follow the same trend (see \cref{tab:ace_results,tab:bss_results}). We hypothesize that this result is due to the negative instances with high pre-test probability representing a particularly difficult subset of images, in that the contextual information does not support the actual diagnosis as closely. As a result, the performance drop is more pronounced as compared to our previous evaluation methods. We broadly observe similar performance trends for the ResNet50 backbone as well. For full results on the ResNet50 backbone, refer to \cref{tab:resnet50_ipw}. 

\section{Discussion}
In this work, we proposed clinical context as a foundation for evaluating state-of-the-art CXR vision models. Rather than relying solely on aggregate performance metrics, we introduced context-derived measures, such as pre-test probability and prior textual mentions of the label (e.g., disease), that give us a proxy for the information available to clinicians in practice. This framework allows us to assess not only whether models perform well on average, but also whether they perform reliably across clinically meaningful subpopulations and scenarios.

We first find that clinical context alone (in this case, the content of prior discharge summaries) is capable of accurately predicting future CXR labels across several different classes of embedding representations. In addition, we trace this predictive signal to discrete, clinically relevant terms in the associated discharge summaries. Next, by stratifying the images into subpopulations based on the pre-test probability of the CXR label, we find that vision models often show stronger performance on lower quantiles of risk, and worse performance in higher risk strata. This indicates that vision models may be most useful in triaging low-risk cases, such as helping to surface unexpected positives, while being less reliable in cases where the relevant label (e.g., a particular disease) is already suspected. 
We also find that performance gaps persist when stratifying by discrete terms in the clinical text, suggesting that clinical context can be used to inform evaluations of performance across several \enquote{resolutions} of medical history.  
More broadly, clinical context can be used to define subpopulations that are readily interpretable by clinical end-users, who may be expected to have a sense of the relevant medical context for a given patient. Hence, information on performance variation across these subpopulations (e.g., low risk vs.\ high risk cases) may be useful for deciding when (and to what extent) to rely on the outputs of medical imaging models in clinical diagnostic tasks.

Using this contextual information, we also evaluate the extent to which model performance is driven by inference of pre-CXR context, as opposed to direct visual diagnostic signal.  We probe this dependence in two ways, by employing both Inverse Probability Weighting (IPW) and Matched Neighbor analysis to decouple the visual signal from the clinical prior.  
The inconsistent stability of model performance under IPW suggests that while performance is not entirely driven by recovery of our particular notion of pre-CXR probability, there exists some dependence on performance on that underlying signal.  
Furthermore, the sharp performance drop observed in the Matched Neighbor analysis reveals a limitation in that for patients with similar clinical risk profiles, models struggle to use direct visual features to resolve diagnostic uncertainty. 

While our Matched Neighbor and IPW analyses demonstrate a clear degradation in performance when the statistical link between context and label is severed, the exact mechanism driving this drop warrants careful interpretation. One hypothesis is that the vision model over-relies on non-clinically relevant visual confounders that happen to correlate with the pre-test probability. However, another explanation is that high pre-test probability negative cases simply possess clinically relevant visual features that make the classification task intrinsically harder. 


Taken together, our results argue for the integration of contextual information into future evaluation pipelines. Standard metrics like AUROC reported in aggregate may overstate the true diagnostic contribution of vision models by conflating context-inference with visual diagnosis. By grounding evaluation in clinical context, we obtain a more complete, nuanced picture of model utility. This helps us distinguish between cases where the model adds genuine diagnostic value to human decision making and those where it merely reiterates the patient's history.

Looking forward, context-based evaluation opens several directions for future work. Beyond chest X-rays, many clinical imaging tasks are embedded within rich longitudinal records that could provide analogous context signals. Benchmarking frameworks that use these signals appropriately in training and evaluation may sharpen our understanding of precisely when and how models complement clinical decision making. 

\paragraph{Limitations}

In this study, we restrict ourselves to MIMIC-CXR and MIMIC-IV, leveraging the unique longitudinal nature of the combined dataset, including the linkage of CXR images and prior discharge summaries.  However, as a result, the generalization of our results to other clinical datasets from other institutions in not guaranteed.

We were also restricted to individuals with prior discharge summaries, which may not represent the general population (e.g., our analysis does not cover new patients who present without any prior context). 

Our methodology establishes a robust statistical relationship between text-derived pre-test probabilities and vision model performance, but it does not empirically map these probabilities to specific visual artifacts within the image. Future work utilizing fine-grained multimodal interpretability techniques or bounding-box annotations is required to explicitly link a patient's textual clinical context to the specific visual features driving a model's prediction.

Finally, we choose to only use textual discharge summaries to represent our prior clinical context. However, many CXR datasets such as MIMIC-IV contain modalities other than text, such as time-series data with ICU lab results. Expanding the scope to modalities other than text may yield more complex, realistic contextual ``stories'' beyond what is captured in the discharge texts.

\clearpage   
\bibliography{references}  

\clearpage      
\onecolumn
\appendix
\renewcommand{\thefigure}{A.\arabic{figure}}
\setcounter{figure}{0}

\renewcommand{\thetable}{A.\arabic{table}}
\setcounter{table}{0}

\section{Experimental Details}
\label{sec:Experimental Details}

\paragraph{Dataset Version}
The study used data from the MIMIC-IV and its extensions, specifically MIMIC-IV v3.0, MIMIC-CXR-JPG v2.0.0, and MIMIC-IV-Note v2.2.

\paragraph{Dataset Processing}
The admission data was joined to the CXR studies and labels via the \enquote{subject\_id} columns, which was then matched with the metadata via the \enquote{subject\_id} and \enquote{study\_id} columns. Finally, the notes was added to this output by joining the dataframes on the \enquote{subject\_id} and \enquote{hadm\_id} columns.

While there exists a standardized train/validation/test split for the MIMIC-CXR database, we chose to construct our data split by randomizing the subjects and performing an 80-10-10 split based on the unique subject IDs. Based on our definition of prior context and due to the nature of each subject having a variable number of admissions, CXR images, and associated clinical notes, our splitting by subject ID ensures that there is no overlap of subjects or their associated images and clinical notes between training, validation, and test data. Following this pipeline, we have 167,291 images in our training set, 21,239 images in our validation set, and 20,770 images in our test set. For distribution of positive and negative image labels in our dataset, refer to \cref{table:dist}.

\paragraph{Vision Model Training}
In order to evaluate performance impacts on state-of-the-art models for this dataset, we developed our vision models following a standard approach, as pre-trained weights are generally unavailable. Specifically, we based our design choices on the methodology outlined in \citet{Seyyed-Kalantari2020-97}, using both DenseNet121 and ResNet50 backbones. Using our custom train/validation/test split, we trained models on NVIDIA L40S GPUs, applying multi-GPU model and data parallelism to accelerate the training process. Our training procedure closely follows that in \citet{Seyyed-Kalantari2020-97}, including a batch size of 256. This approach yielded models with state-of-the-art performance (see \cref{table:vision_performance} for details), which enables us to perform our evaluations on state-of-the-art vision models. 

We also attempted to evaluate BioMedCLIP in a zero-shot manner to avoid data leakage and use larger Vision-Language Models (VLMs), but it achieved a macro AUROC of only $0.585$ (vs. $0.803$ for DenseNet), making downstream analysis uninformative. See \cref{table:biomedclip_performance} for details.

\paragraph{Preprocessing and Embedding of Prior Clinical Notes}
We first preprocessed all of the clinical notes by implementing punctuation and stopword removal, replacement of repetitive words, and lower-casing. To embed a summarized version of the prior clinical context of a CXR, we profiled six pre-trained language model encoders publicly available on HuggingFace: Mistral, BERT, BioLinkBERT, PubMedBERT, ClinicalBERT, and RoBERTa. We processed the clinical notes using a sliding window chunking strategy to handle notes exceeding the models' context limit. Each chunk was independently encoded, mean-pooled, and the resulting embeddings were aggregated via mean pooling to produce a final note representation. To ensure consistency and improve downstream stability, the pooled embeddings were normalized to have a unit L2 norm. Finally, in cases where multiple notes occurred prior to a CXR, their embeddings were averaged to produce a single summarized representation of the patient's prior context.
    
\paragraph{Hyperparameter and Model Class Selection}
We evaluated ten distinct classification models provided in sklearn: \texttt{Perceptron, RidgeClassifierCV, PassiveAggressiveClassifier, GaussianNB, LinearDiscriminantAnalysis, RandomForestClassifier, KNeighborsClassifier, MLPClassifier, LinearSVC, and SGDClassifier}.

We performed an automated search using \texttt{GridSearchCV} over predefined parameter grids (see Supplementary Tables \cref{tab:base_classifiers} and \cref{tab:hp_configs} for specific grids). To select the optimal hyperparameters for each classifier type, we employed a 5-fold stratified group cross-validation (\texttt{StratifiedGroupKFold}) on the combined training-validation data.  
The \enquote{subject\_id} were designated as the grouping variable to ensure that all data from a single patient resided within the same fold during cross-validation, preventing patient-level data leakage between folds. The specific combination of language model embedding, classifier architecture, and hyperparameters yielding the highest mean AUROC for each label was selected as the best predictor. These final models were calibrated using \texttt{CalibratedClassifierCV} with sigmoid calibration (Platt scaling). See \cref{table:text_performance} for details on the best performing configurations.

\paragraph{BoW}
A bag-of-words (BoW) representation was constructed on these notes using the \enquote{CountVectorizer} in scikit-learn, restricted to a vocabulary of the top $8,192$ most frequent tokens (after removing English stop words), with \enquote{min\_df=5} and \enquote{max\_df=0.90} to filter rare and overly common tokens/words.

We evaluated seven distinct classification models provided in sklearn: \texttt{Perceptron, RidgeClassifierCV, PassiveAggressiveClassifier, RandomForestClassifier,  DecisionTreeClassifier, LinearSVC, and SGDClassifier}. These were chosen because of their interpretable features/coefficients.

We performed an automated search over the interpretable model types and hyperparameters in the same manner as \Cref{subsec:Clinical}. See Supplementary Tables \cref{tab:bow_classifiers} and \cref{tab:bow_hp_configs} for specific grids), and \cref{table:bow_performance} for details on the best performing classifiers. Following the selection of the best classifier, we examined the most important features by extracting the feature importances/coefficients and subsetting to the top ten contributing terms in the clinical text (refer to \cref{table:feature_importances} for results).

\paragraph{Calibration Metrics}
To evaluate the reliability of the vision and text models' probability estimates, we computed the Adaptive Calibration Error (ACE). Unlike the static Expected Calibration Error (ECE), which uses bins of equal width (e.g., 0-0.1, 0.1-0.2), ACE utilizes an adaptive binning scheme where bins are constructed to contain an equal number of samples. This approach was chosen for its robustness in imbalanced datasets where model predictions may be heavily skewed toward low probabilities.

We partition the  samples into  disjoint bins  such that the number of samples in each bin is approximately equal:
\begin{equation}
|B_k| \approx \frac{N}{K}
\end{equation}
For each bin , we compute the average predicted probability  and the observed label proportion :
\begin{equation}
\bar{p}_k = \frac{1}{|B_k|} \sum_{i \in B_k} p_i, \quad \bar{y}_k = \frac{1}{|B_k|} \sum_{i \in B_k} y_i
\end{equation}
The ACE is then calculated as the mean absolute calibration error across these equal-frequency bins:
\begin{equation}
\text{ACE} = \frac{1}{K} \sum_{k=1}^K | \bar{y}_k - \bar{p}_k |
\end{equation}

Additionally, we reported the Brier Skill Score (BSS), which measures the relative improvement in mean squared error compared to a baseline. We first compute the Brier Score (BS) as:
\begin{equation}
BS = \frac{1}{N} \sum_{i=1}^N (y_i - p_i)^2
\end{equation}
The Brier Skill Score (BSS) normalizes this metric against a baseline reference classifier that has no discriminative ability and simply predicts the sample prevalence. We define the reference Brier Score, as the Brier Score obtained if the predicted probability for every instance was fixed to the global disease prevalence. The Brier Skill Score is defined as:
\begin{equation}
BSS = 1 - \frac{BS}{BS_{\text{ref}}}
\end{equation}
A BSS of 0 indicates the model performs no better than the prevalence baseline, while a BSS of 1 indicates perfect prediction. Negative values indicate performance worse than the baseline.

\clearpage
\section{Additional Vision Model Results}
\label{sec:appendix_vision}

\begin{table}[h!]
	\centering
	\begin{tabular}{l c c c}
		\toprule
		& \multicolumn{2}{c}{Our Architectures (AUROC)} & Comparison (AUROC) \\
		\cmidrule(lr){2-3} \cmidrule(lr){4-4}
		Label                      & DenseNet121 & ResNet50 & Chexclusion \\
		\midrule
		Atelectasis                & 0.8008      & 0.7379   & 0.837       \\
		Cardiomegaly               & 0.8108      & 0.7355   & 0.828       \\
		Consolidation              & 0.8182      & 0.7246   & 0.844       \\
		Edema                      & 0.8914      & 0.8398   & 0.904       \\
		Enlarged Cardiomediastinum & 0.7074      & 0.6642   & 0.757       \\
		Fracture                   & 0.681       & 0.6524   & 0.718       \\
		Lung Lesion                & 0.7565      & 0.6326   & 0.772       \\
		Lung Opacity               & 0.7491      & 0.6718   & N/A         \\
		Pleural Effusion           & 0.9233      & 0.8322   & N/A         \\
		Pleural Other              & 0.8061      & 0.7233   & 0.848       \\
		Pneumonia                  & 0.7073      & 0.6173   & 0.748       \\
		Pneumothorax               & 0.8790      & 0.7818   & 0.903       \\
		Support Devices            & 0.906       & 0.8490   & 0.927       \\
		\bottomrule
	\end{tabular}
	\caption{Comparisons of our trained DenseNet121 and ResNet50 vision models against the state-of-the-art Chexclusion model from \citet{Seyyed-Kalantari2020-97}.}
	\label{table:vision_performance}
\end{table}

\begin{table}[h!]
	\centering
	\begin{tabular}{l c}
		\toprule
		Label                      & BioMedClip AUROC \\
		\midrule
		Atelectasis                & 0.507            \\
		Cardiomegaly               & 0.646            \\
		Consolidation              & 0.587            \\
		Edema                      & 0.698            \\
		Enlarged Cardiomediastinum & 0.551            \\
		Fracture                   & 0.499            \\
		Lung Lesion                & 0.545            \\
		Lung Opacity               & 0.563            \\
		Pleural Effusion           & 0.699            \\
		Pleural Other              & 0.444            \\
		Pneumonia                  & 0.564            \\
		Pneumothorax               & 0.576            \\
		Support Devices            & 0.727            \\
		\bottomrule
	\end{tabular}
	\caption{Performance of the BioMedClip Vision Model on Test Data}
	\label{table:biomedclip_performance}
\end{table}

\begin{table}[t]
	\centering
	\begin{tabular}{c c c c}
		\toprule
		Label                      & Negatives & Positives & Proportion of Positives \\ [0.5ex]
		\midrule
		Atelectasis                & 171488    & 37812     & 0.22                    \\
		Cardiomegaly               & 169585    & 39715     & 0.234                   \\
		Consolidation              & 200875    & 8425      & 0.042                   \\
		Edema                      & 185961    & 23339     & 0.126                   \\
		Enlarged Cardiomediastinum & 203854    & 5446      & 0.027                   \\
		Fracture                   & 205555    & 3745      & 0.018                   \\
		Lung Lesion                & 202778    & 6522      & 0.032                   \\
		Lung Opacity               & 164244    & 45056     & 0.274                   \\
		Pleural Effusion           & 161091    & 48209     & 0.299                   \\
		Pleural Other              & 207075    & 2225      & 0.011                   \\
		Pneumonia                  & 193869    & 15431     & 0.079                   \\
		Pneumothorax               & 202296    & 7004      & 0.035                   \\
		Support Devices            & 163275    & 46025     & 0.282                   \\
		\bottomrule
	\end{tabular}
	\caption{Distribution of Positive and Negative CXR Labels in Our Dataset}
	\label{table:dist}
\end{table}

\begin{figure}[htbp]
    \centering
    \includegraphics[width=1.0\textwidth]{img/vision_calibration_grid.png}
    \caption{Calibration Plots for Vision Model}
\end{figure}



\clearpage
\section{Additional Text Model Results}
\label{sec:appendix_text}

\begin{table}[ht]
	\centering
	\caption{Overview of evaluated LM text classifiers and their base configurations}
	\label{tab:base_classifiers}
	\begin{tabular}{ll}
		\toprule
		\textbf{Classifier}         & \textbf{Base Configuration}                                                 \\
		\midrule
		Perceptron                  & \texttt{penalty='l2'}, \texttt{random\_state=42}                            \\
		RidgeClassifierCV           & \texttt{alphas=np.logspace(-3, 3, 7)}                                       \\
		PassiveAggressiveClassifier & \texttt{random\_state=42}                                                   \\
		GaussianNB                  & default parameters                                                          \\
		LinearDiscriminantAnalysis  & \texttt{shrinkage='auto'}, \texttt{solver='lsqr'}                           \\
		RandomForestClassifier      & \texttt{random\_state=42}                                                   \\
		KNeighborsClassifier        & default parameters                                                           \\
		MLPClassifier               & \texttt{early\_stopping=True}, \texttt{validation\_fraction=0.1},           \\
		                            & \texttt{max\_iter=300}, \texttt{random\_state=42}                           \\
		LinearSVC                   & \texttt{max\_iter=1000}, \texttt{random\_state=42}                          \\
		SGDClassifier               & \texttt{loss='log\_loss'}, \texttt{penalty='l2'}, \texttt{random\_state=42} \\
		\bottomrule
	\end{tabular}
\end{table}

\begin{table}[ht]
	\centering
	\caption{Hyperparameter grids used for each LM classifier during cross-validation}
	\label{tab:hp_configs}
	\begin{tabular}{ll}
		\toprule
		\textbf{Classifier}         & \textbf{Hyperparameter Grid}                            \\
		\midrule
		Perceptron                  & \texttt{alpha: [1e-5, 1e-4, 1e-3]}                      \\
		RidgeClassifierCV           & \texttt{alphas: [np.logspace(-4, 4, 9)]}                \\
		PassiveAggressiveClassifier & \texttt{C: [0.01, 0.1, 0.5, 1.0]}                       \\
		GaussianNB                  & \textit{(No hyperparameters tuned)}                     \\
		LinearDiscriminantAnalysis  & \texttt{solver: ['lsqr', 'eigen']},                     \\
		                            & \texttt{shrinkage: [None, 'auto']}                      \\
		RandomForestClassifier      & \texttt{n\_estimators: [50, 100, 150]},                 \\
		                            & \texttt{max\_depth: [None, 5, 10]},                     \\
		                            & \texttt{min\_samples\_leaf: [1, 5, 10]}                 \\
		KNeighborsClassifier        & \texttt{n\_neighbors: [3, 5, 7, 9]}                     \\
		MLPClassifier               & \texttt{alpha: [1e-5, 1e-4, 1e-3]},                     \\
		                            & \texttt{hidden\_layer\_sizes: [(64,), (128,), (64,32)]} \\
		LinearSVC                   & \texttt{C: [0.1, 1.0, 10.0]}                            \\
		SGDClassifier               & \texttt{alpha: [1e-5, 1e-4, 1e-3]},                     \\
		                            & \texttt{penalty: ['l1', 'l2']}                          \\
		\bottomrule
	\end{tabular}
\end{table}

\begin{table}[t]
    \centering
    \caption{Best performing classifiers per label and Language Model}
    
    \resizebox{\textwidth}{!}{%
        \begin{filecontents*}{csv/all_LM_classifiers.csv}
Label,BERT,BioLinkBERT,ClinicalBERT,Mistral,PubMedBERT,RoBERTa
Atelectasis,0.589 (SVC),0.606 (SVC),0.608 (SVC),0.623 (SVC),0.609 (SVC),0.603 (SVC)
Cardiomegaly,0.684 (SVC),0.695 (SVC),0.683 (LinearDisc),0.701 (MLP),0.693 (SVC),0.683 (Ridge)
Consolidation,0.613 (SVC),0.624 (SVC),0.623 (Ridge),0.633 (SVC),0.621 (SVC),0.610 (SVC)
Edema,0.730 (PassAgg),0.746 (SVC),0.738 (SVC),0.751 (PassAgg),0.741 (SVC),0.730 (SVC)
Enlarged Cardiomediastinum,0.558 (RF),0.560 (Ridge),0.569 (SGD),0.569 (SVC),0.566 (Ridge),0.557 (SVC)
Fracture,0.620 (Ridge),0.655 (SVC),0.617 (SGD),0.676 (SVC),0.628 (SVC),0.614 (SVC)
Lung Lesion,0.691 (SVC),0.768 (SVC),0.742 (SVC),0.765 (MLP),0.759 (SVC),0.737 (SVC)
Lung Opacity,0.604 (Ridge),0.625 (PassAgg),0.614 (SVC),0.628 (MLP),0.620 (SVC),0.607 (SVC)
Pleural Effusion,0.696 (MLP),0.724 (Ridge),0.711 (Percep),0.745 (SVC),0.728 (MLP),0.721 (MLP)
Pleural Other,0.611 (SVC),0.652 (RF),0.682 (Ridge),0.654 (SVC),0.659 (SVC),0.622 (SVC)
Pneumonia,0.561 (SVC),0.577 (SVC),0.555 (LinearDisc),0.592 (SVC),0.580 (Ridge),0.581 (SVC)
Pneumothorax,0.689 (SVC),0.695 (PassAgg),0.705 (SVC),0.712 (SVC),0.683 (PassAgg),0.694 (PassAgg)
Support Devices,0.663 (SVC),0.672 (MLP),0.623 (GaussianNB),0.682 (SVC),0.673 (SVC),0.670 (SVC)
\end{filecontents*}
\pgfplotstabletypeset[
            col sep=comma,
            string type,
            header=true,
            every head row/.style={before row=\toprule,after row=\midrule},
            every last row/.style={after row=\bottomrule},
            column type=l
        ]{csv/all_LM_classifiers.csv}
    }%
    \label{table:text_performance}
\end{table}

\begin{table}[tbhp]
    \centering
    \caption{Calibration performance of text-only classifiers. We report the Adaptive Calibration Error (ACE), Brier Score, and Brier Skill Score (BSS) for the selected models trained on prior clinical notes.}
    
    \resizebox{\textwidth}{!}{%
        \begin{tabular}{lllccc}
            \toprule
            \textbf{Label} & \textbf{Language Model} & \textbf{Classifier} & \textbf{ACE} & \textbf{Brier Score} & \textbf{Brier Skill Score} \\
            \midrule
            Atelectasis & Mistral & LinearSVC & 0.012 & 0.144 & 0.027 \\
            Cardiomegaly & Mistral & MLPClassifier & 0.022 & 0.135 & 0.075 \\
            Consolidation & Mistral & LinearSVC & 0.006 & 0.034 & 0.007 \\
            Edema & Mistral & PassiveAggressiveClassifier & 0.018 & 0.080 & 0.079 \\
            Enlarged Cardiomediastinum & PubMedBERT & RidgeClassifierCV & 0.002 & 0.024 & 0.001 \\
            Fracture & Mistral & LinearSVC & 0.006 & 0.017 & -0.005 \\
            Lung Lesion & Mistral & MLPClassifier & 0.010 & 0.028 & 0.038 \\
            Lung Opacity & Mistral & MLPClassifier & 0.014 & 0.156 & 0.036 \\
            Pleural Effusion & Mistral & LinearSVC & 0.026 & 0.154 & 0.132 \\
            Pleural Other & Mistral & LinearSVC & 0.002 & 0.010 & 0.008 \\
            Pneumonia & Mistral & LinearSVC & 0.006 & 0.066 & 0.007 \\
            Pneumothorax & Mistral & LinearSVC & 0.010 & 0.034 & 0.033 \\
            Support Devices & Mistral & LinearSVC & 0.015 & 0.156 & 0.074 \\
            \bottomrule
        \end{tabular}%
    }
    
    \label{table:text_performance_calibration}
\end{table}

\begin{table}[tbhp]
    \centering
    \caption{Representative examples of clinical notes with high and low predicted probabilities for Pleural Effusion. We truncate the raw notes to the relevant portions for clarity, but the full set of notes may be obtained using the Subject and Study ID, which we provide for reference.}
    \label{app:pre-test_ex}
    
    \begin{minipage}{\textwidth}
        \scriptsize
        \textbf{High Predicted Probability (True Positive)} \\
        \textbf{Subject ID:} 14965197 \quad \textbf{Study ID:} 53934290 \\
        \textbf{Model Probability:} 0.9237 \quad \textbf{Reference Label:} 1 (Positive)
        
        \begin{lstlisting}[breaklines=true, basicstyle=\ttfamily\scriptsize, frame=single, escapeinside={(*@}{@*)}]
name:  ___                    unit no:   ___
service: medicine
...
chief complaint:
massive (*@\textbf{pleural effusion}@*)
 
major surgical or invasive procedure:
chest tube placed ___ by ip
(*@\textbf{thoracentesis}@*) ___ by ip

history of present illness:
... found to have massive (*@\textbf{pleural effusion}@*).
(*@\textbf{thoracentesis}@*) on ___ removed 1.5l of (*@\textbf{fluid}@*).
\end{lstlisting}

        \vspace{0.2cm}

        \textbf{Low Predicted Probability (True Negative)} \\
        \textbf{Subject ID:} 10486638 \quad \textbf{Study ID:} 57867200 \\
        \textbf{Model Probability:} 0.0003 \quad \textbf{Reference Label:} 0 (Negative)

        \begin{lstlisting}[breaklines=true, basicstyle=\ttfamily\scriptsize, frame=single, escapeinside={(*@}{@*)}]
name:  ___                     unit no:   ___
service: medicine
...
chief complaint:
chest pain and vertigo
...
physical examination:
lungs: (*@\textbf{clear}@*) to auscultation bilaterally. (*@\textbf{no wheezes}@*), rales, or rhonchi.
...
brief hospital course:
chest x-ray showed (*@\textbf{no pleural effusion}@*) or pneumothorax.
\end{lstlisting}
    \end{minipage}
\end{table}

\begin{table}[t]
    \centering
    \caption{Calibration performance of BoW classifiers. We report the Adaptive Calibration Error (ACE), Brier Score, and Brier Skill Score (BSS) for the selected models trained on prior clinical notes.}
    
    \resizebox{\textwidth}{!}{%
        \begin{tabular}{llccc}
            \toprule
            \textbf{Label} & \textbf{Classifier} & \textbf{ACE} & \textbf{Brier Score} & \textbf{Brier Skill Score} \\
            \midrule
            Atelectasis & RandomForestClassifier & 0.014 & 0.144 & 0.028 \\
            Cardiomegaly & RandomForestClassifier & 0.015 & 0.136 & 0.068 \\
            Consolidation & RandomForestClassifier & 0.009 & 0.034 & 0.003 \\
            Edema & RandomForestClassifier & 0.016 & 0.082 & 0.053 \\
            Enlarged Cardiomediastinum & RandomForestClassifier & 0.004 & 0.024 & 0.001 \\
            Fracture & DecisionTreeClassifier & 0.003 & 0.017 & 0.021 \\
            Lung Lesion & RandomForestClassifier & 0.004 & 0.027 & 0.056 \\
            Lung Opacity & RandomForestClassifier & 0.012 & 0.157 & 0.030 \\
            No Finding & RandomForestClassifier & 0.017 & 0.200 & 0.122 \\
            Pleural Effusion & RandomForestClassifier & 0.016 & 0.154 & 0.132 \\
            Pleural Other & SGDClassifier & 0.003 & 0.010 & 0.001 \\
            Pneumonia & DecisionTreeClassifier & 0.006 & 0.067 & 0.003 \\
            Pneumothorax & RandomForestClassifier & 0.006 & 0.033 & 0.043 \\
            Support Devices & RandomForestClassifier & 0.018 & 0.157 & 0.070 \\
            \bottomrule
        \end{tabular}%
    }
    \label{table:bow_performance_calibration}
\end{table}

\begin{figure}[htbp]
    \centering
    \includegraphics[width=1.0\textwidth]{img/text_calibration_grid.png}
    \caption{Calibration Plots for Text Classifiers Trained on Prior Context}
    \label{fig:text_calibration_plots}
\end{figure}

\begin{table}[ht]
	\centering
	\caption{Overview of BoW-based classifiers and their base configurations}
	\label{tab:bow_classifiers}
	\begin{tabular}{ll}
		\toprule
		\textbf{Classifier}         & \textbf{Base Configuration}                                                   \\
		\midrule
		Perceptron                  & \texttt{penalty='l2'}, \texttt{random\_state=42}, \texttt{max\_iter=1000}, \texttt{tol=1e-3} \\
		RidgeClassifierCV           & \texttt{alphas=np.logspace(-3, 3, 7)}                                         \\
		PassiveAggressiveClassifier & \texttt{random\_state=42}, \texttt{max\_iter=1000}, \texttt{tol=1e-3}         \\
		RandomForestClassifier      & \texttt{random\_state=42}                                                     \\
		LinearSVC                   & \texttt{max\_iter=2000}, \texttt{random\_state=42}, \texttt{dual='auto'}      \\
		SGDClassifier               & \texttt{loss='log\_loss'}, \texttt{penalty='l2'}, \texttt{random\_state=42}, \texttt{max\_iter=1000}, \texttt{tol=1e-3} \\
		DecisionTreeClassifier      & \texttt{random\_state=42}                                                     \\
		\bottomrule
	\end{tabular}
\end{table}

\begin{table}[ht]
	\centering
	\caption{Hyperparameter grids used for BoW classifiers during cross-validation}
	\label{tab:bow_hp_configs}
	\begin{tabular}{ll}
		\toprule
		\textbf{Classifier}         & \textbf{Hyperparameter Grid}                                \\
		\midrule
		Perceptron                  & \texttt{alpha: [1e-6, 1e-5, 1e-4]}                           \\
		RidgeClassifierCV           & \texttt{alphas: [np.logspace(-4, 4, 9)]}                     \\
		PassiveAggressiveClassifier & \texttt{C: [0.001, 0.01, 0.1, 1.0]}                          \\
		RandomForestClassifier      & \texttt{n\_estimators: [50, 100, 200]},                      \\
		                            & \texttt{max\_depth: [10, 20, None]},                         \\
		                            & \texttt{min\_samples\_leaf: [1, 5, 10]}                      \\
		LinearSVC                   & \texttt{C: [0.01, 0.1, 1.0, 10.0]}                           \\
		SGDClassifier               & \texttt{alpha: [1e-6, 1e-5, 1e-4]},                          \\
		                            & \texttt{penalty: ['l1', 'l2', 'elasticnet']}                 \\
		DecisionTreeClassifier      & \texttt{max\_depth: [5, 10, 20, None]},                      \\
		                            & \texttt{min\_samples\_leaf: [1, 5, 10, 20]}                  \\
		\bottomrule
	\end{tabular}
\end{table}

\begin{filecontents*}{top_feature_importances.csv}
feature,importance,rank,label_name,classifier_name,coefficient,abs_coefficient
metastatic,0.01863784925,1,Lung Lesion,RandomForestClassifier,,
metastasis,0.01332490948,2,Lung Lesion,RandomForestClassifier,,
metastases,0.01284911658,3,Lung Lesion,RandomForestClassifier,,
lung,0.009337958029,4,Lung Lesion,RandomForestClassifier,,
nsclc,0.007777329373,5,Lung Lesion,RandomForestClassifier,,
oncology,0.006987599988,6,Lung Lesion,RandomForestClassifier,,
il2,0.006959053939,7,Lung Lesion,RandomForestClassifier,,
tumor,0.006661712776,8,Lung Lesion,RandomForestClassifier,,
nodules,0.006192722259,9,Lung Lesion,RandomForestClassifier,,
nonsmall,0.005902637214,10,Lung Lesion,RandomForestClassifier,,
pleural,0.01459591239,1,Pleural Effusion,RandomForestClassifier,,
thoracentesis,0.01389340598,2,Pleural Effusion,RandomForestClassifier,,
effusion,0.009709947157,3,Pleural Effusion,RandomForestClassifier,,
effusions,0.008726069767,4,Pleural Effusion,RandomForestClassifier,,
atelectasis,0.004093384711,5,Pleural Effusion,RandomForestClassifier,,
lasix,0.003948792247,6,Pleural Effusion,RandomForestClassifier,,
compressive,0.003910660697,7,Pleural Effusion,RandomForestClassifier,,
tube,0.003893108582,8,Pleural Effusion,RandomForestClassifier,,
pleurx,0.003677873823,9,Pleural Effusion,RandomForestClassifier,,
cytology,0.00362098549,10,Pleural Effusion,RandomForestClassifier,,
pneumothorax,0.01113693761,1,Pneumothorax,RandomForestClassifier,,
cardiothoracic,0.00744237441,2,Pneumothorax,RandomForestClassifier,,
ptx,0.006500874755,3,Pneumothorax,RandomForestClassifier,,
apical,0.005992919101,4,Pneumothorax,RandomForestClassifier,,
tube,0.005845197797,5,Pneumothorax,RandomForestClassifier,,
nl,0.00583192852,6,Pneumothorax,RandomForestClassifier,,
spontaneous,0.005264218369,7,Pneumothorax,RandomForestClassifier,,
pigtail,0.004948041625,8,Pneumothorax,RandomForestClassifier,,
abnormal,0.003903366949,9,Pneumothorax,RandomForestClassifier,,
vats,0.00383820557,10,Pneumothorax,RandomForestClassifier,,
lasix,0.006265608956,1,Edema,RandomForestClassifier,,
diuresed,0.005575644134,2,Edema,RandomForestClassifier,,
cardiomegaly,0.005486928186,3,Edema,RandomForestClassifier,,
furosemide,0.005208072338,4,Edema,RandomForestClassifier,,
failure,0.004406526897,5,Edema,RandomForestClassifier,,
ef,0.004332732952,6,Edema,RandomForestClassifier,,
congestive,0.003941699324,7,Edema,RandomForestClassifier,,
chf,0.003867438774,8,Edema,RandomForestClassifier,,
diuresis,0.003792339753,9,Edema,RandomForestClassifier,,
diastolic,0.003573421451,10,Edema,RandomForestClassifier,,
cardiomegaly,0.01286861647,1,Cardiomegaly,RandomForestClassifier,,
lasix,0.00611186228,2,Cardiomegaly,RandomForestClassifier,,
furosemide,0.005676550433,3,Cardiomegaly,RandomForestClassifier,,
diuresed,0.005610363037,4,Cardiomegaly,RandomForestClassifier,,
fibrillation,0.005322495866,5,Cardiomegaly,RandomForestClassifier,,
congestive,0.004583380503,6,Cardiomegaly,RandomForestClassifier,,
ef,0.004460934714,7,Cardiomegaly,RandomForestClassifier,,
diastolic,0.004423489502,8,Cardiomegaly,RandomForestClassifier,,
chf,0.004335244359,9,Cardiomegaly,RandomForestClassifier,,
atrial,0.004292803245,10,Cardiomegaly,RandomForestClassifier,,
prior,0.227540441,1,Fracture,DecisionTreeClassifier,,
rib,0.2207677758,2,Fracture,DecisionTreeClassifier,,
clavicle,0.09348979581,3,Fracture,DecisionTreeClassifier,,
velcade,0.04623628326,4,Fracture,DecisionTreeClassifier,,
0945pm,0.04528351139,5,Fracture,DecisionTreeClassifier,,
acute,0.04482248189,6,Fracture,DecisionTreeClassifier,,
completing,0.04389272787,7,Fracture,DecisionTreeClassifier,,
abdomen,0.04179216696,8,Fracture,DecisionTreeClassifier,,
0220pm,0.03581871922,9,Fracture,DecisionTreeClassifier,,
18,0.03556758664,10,Fracture,DecisionTreeClassifier,,
picc,0.01237358311,1,Support Devices,RandomForestClassifier,,
placement,0.008536741876,2,Support Devices,RandomForestClassifier,,
tube,0.008512463485,3,Support Devices,RandomForestClassifier,,
transferred,0.008210591176,4,Support Devices,RandomForestClassifier,,
osh,0.007296842954,5,Support Devices,RandomForestClassifier,,
line,0.005431271076,6,Support Devices,RandomForestClassifier,,
flush,0.004235559859,7,Support Devices,RandomForestClassifier,,
pacemaker,0.004186328066,8,Support Devices,RandomForestClassifier,,
tip,0.004152788477,9,Support Devices,RandomForestClassifier,,
svc,0.003903867497,10,Support Devices,RandomForestClassifier,,
pneumonia,0.004645727031,1,Consolidation,RandomForestClassifier,,
bronchoscopy,0.004479790235,2,Consolidation,RandomForestClassifier,,
lobe,0.004085000291,3,Consolidation,RandomForestClassifier,,
pleural,0.003280321244,4,Consolidation,RandomForestClassifier,,
bronchiectasis,0.00308405866,5,Consolidation,RandomForestClassifier,,
lung,0.003011046617,6,Consolidation,RandomForestClassifier,,
tube,0.002925409912,7,Consolidation,RandomForestClassifier,,
opacities,0.00281801472,8,Consolidation,RandomForestClassifier,,
hcap,0.002626492749,9,Consolidation,RandomForestClassifier,,
effusion,0.00259852285,10,Consolidation,RandomForestClassifier,,
atelectasis,0.01757023616,1,Atelectasis,RandomForestClassifier,,
pleural,0.006271913782,2,Atelectasis,RandomForestClassifier,,
bibasilar,0.00598040783,3,Atelectasis,RandomForestClassifier,,
collapse,0.005270571751,4,Atelectasis,RandomForestClassifier,,
thoracentesis,0.005221985691,5,Atelectasis,RandomForestClassifier,,
tube,0.004693420487,6,Atelectasis,RandomForestClassifier,,
compressive,0.004159031513,7,Atelectasis,RandomForestClassifier,,
effusions,0.004023706505,8,Atelectasis,RandomForestClassifier,,
furosemide,0.003796625451,9,Atelectasis,RandomForestClassifier,,
facility,0.003533252677,10,Atelectasis,RandomForestClassifier,,
pneumonia,0.0118668549,1,Lung Opacity,RandomForestClassifier,,
opacities,0.0110167586,2,Lung Opacity,RandomForestClassifier,,
lung,0.009430481553,3,Lung Opacity,RandomForestClassifier,,
bronchoscopy,0.007126853996,4,Lung Opacity,RandomForestClassifier,,
opacity,0.006811578226,5,Lung Opacity,RandomForestClassifier,,
aspiration,0.006587325587,6,Lung Opacity,RandomForestClassifier,,
lobe,0.00626861913,7,Lung Opacity,RandomForestClassifier,,
bronchiectasis,0.005534682302,8,Lung Opacity,RandomForestClassifier,,
sputum,0.005374823027,9,Lung Opacity,RandomForestClassifier,,
copd,0.004887029102,10,Lung Opacity,RandomForestClassifier,,
icd,,1,Pleural Other,SGDClassifier,-232.8305297,232.8305297
whipple,,2,Pleural Other,SGDClassifier,-178.411282,178.411282
gastropathy,,3,Pleural Other,SGDClassifier,-162.2917335,162.2917335
crohns,,4,Pleural Other,SGDClassifier,-158.0134791,158.0134791
aml,,5,Pleural Other,SGDClassifier,-144.3296164,144.3296164
cerebral,,6,Pleural Other,SGDClassifier,-139.9081511,139.9081511
hearing,,7,Pleural Other,SGDClassifier,-127.8643743,127.8643743
t12,,8,Pleural Other,SGDClassifier,-122.4832463,122.4832463
bacteremia,,9,Pleural Other,SGDClassifier,-121.1774295,121.1774295
paroxetine,,10,Pleural Other,SGDClassifier,-105.7691966,105.7691966
pneumonia,0.2753894553,1,Pneumonia,DecisionTreeClassifier,,
bronchiectasis,0.07601937016,2,Pneumonia,DecisionTreeClassifier,,
congestion,0.07192239854,3,Pneumonia,DecisionTreeClassifier,,
ran,0.05491221813,4,Pneumonia,DecisionTreeClassifier,,
inhibitor,0.05447529222,5,Pneumonia,DecisionTreeClassifier,,
pump,0.05428538109,6,Pneumonia,DecisionTreeClassifier,,
multifocal,0.0317465746,7,Pneumonia,DecisionTreeClassifier,,
cefepime,0.03131304734,8,Pneumonia,DecisionTreeClassifier,,
poorly,0.03078622525,9,Pneumonia,DecisionTreeClassifier,,
emphysema,0.03023526459,10,Pneumonia,DecisionTreeClassifier,,
tube,0.004224720446,1,Enlarged Cardiomediastinum,RandomForestClassifier,,
cardiothoracic,0.002867359543,2,Enlarged Cardiomediastinum,RandomForestClassifier,,
pleurex,0.002791340118,3,Enlarged Cardiomediastinum,RandomForestClassifier,,
dissection,0.002583807582,4,Enlarged Cardiomediastinum,RandomForestClassifier,,
wout,0.002498580791,5,Enlarged Cardiomediastinum,RandomForestClassifier,,
jtube,0.002347047388,6,Enlarged Cardiomediastinum,RandomForestClassifier,,
pox,0.002285750495,7,Enlarged Cardiomediastinum,RandomForestClassifier,,
esophagectomy,0.002225538566,8,Enlarged Cardiomediastinum,RandomForestClassifier,,
wk,0.002188991665,9,Enlarged Cardiomediastinum,RandomForestClassifier,,
oxygenation,0.002131884975,10,Enlarged Cardiomediastinum,RandomForestClassifier,,
\end{filecontents*}
\pgfplotstabletypeset[
    begin table=\begin{longtable},
    end table=\end{longtable},
    col sep=comma,
    string type,
    header=true,
    columns={feature,rank,label_name,classifier_name}, 
    columns/feature/.style={string type, column name=Feature},
    columns/rank/.style={column name=Rank},
    columns/label_name/.style={string type, column name=Label Name},
    columns/classifier_name/.style={string type, column name=Classifier Name},
    every head row/.style={
        before row={
            \caption{Top Features Ranked by Importance by the BoW Text Classifiers. Each feature corresponds to a token from the bag-of-words vocabulary. Importance scores correspond to the magnitude of the coefficient assigned to the feature by the classifier. Across all image labels, clinically meaningful words consistently appeared in the top ten features, which suggests the importance of specific language features within the discharge notes as strong indicators of disease status.}\label{table:feature_importances}\\
            \toprule
        }, 
        after row={
            \midrule
            \endfirsthead
            \caption[]{Top Features Ranked by Importance by the BoW Text Classifiers.}\\
            \toprule
            Feature & Rank & Label Name & Classifier Name \\
            \midrule
            \endhead
            \midrule
            \multicolumn{4}{r}{\textit{Continued on next page}}\\
            \endfoot
            \bottomrule
            \endlastfoot
        }
    }
]{top_feature_importances.csv}\label{tab:feature_importances}

\begin{table}[t]
	\centering
	\caption{Best Performing BoW Text Classifiers. We train classifiers to predict the disease labels of CXR images based on prior clinical notes without looking at the image itself. All models use an BoW embedding of prior clinical notes, and are trained on top of that representation. We note the specific label, classifier type, and best test set AUROC achieved.}
	\centering
	\begin{filecontents*}{csv/bow_best_interpretable_models_test.csv}
Label,Classifier,AUROC
Lung Lesion,RandomForestClassifier,0.758957518
Pleural Effusion,RandomForestClassifier,0.734148124
Edema,RandomForestClassifier,0.717096473
Pneumothorax,RandomForestClassifier,0.702883616
Cardiomegaly,RandomForestClassifier,0.688324271
Support Devices,RandomForestClassifier,0.680903114
Fracture,DecisionTreeClassifier,0.657237787
Atelectasis,RandomForestClassifier,0.62260748
Lung Opacity,RandomForestClassifier,0.616698705
Consolidation,RandomForestClassifier,0.610689545
Pleural Other,SGDClassifier,0.577126473
Enlarged Cardiomediastinum,RandomForestClassifier,0.573320728
Pneumonia,DecisionTreeClassifier,0.550070497
\end{filecontents*}
\pgfplotstabletypeset[
            col sep=comma,
            string type,
            header=true,
            columns={{Label},{Classifier},AUROC},  
            columns/{Label}/.style={string type, column name=Label},
            columns/{Classifier}/.style={string type, column name=Classifier},
            columns/AUROC/.style={numeric type, fixed, fixed zerofill, precision=3, column name=AUROC},
            every head row/.style={before row=\toprule,after row=\midrule},
            every last row/.style={after row=\bottomrule}
        ]{csv/bow_best_interpretable_models_test.csv}
	\label{table:bow_performance}
\end{table}

\begin{table}[t]
	\centering
	\caption{XGBoost Classifier Performance. Predicting disease labels of CXR images using XGBoost. We note the specific label, classifier type, and best test set AUROC achieved.}
	\centering
	\pgfplotstabletypeset[
            col sep=comma,
            string type,
            header=true,
            columns={{Label},{Classifier},AUROC},  
            columns/{Label}/.style={string type, column name=Label},
            columns/{Classifier}/.style={string type, column name=Classifier},
            columns/AUROC/.style={numeric type, fixed, fixed zerofill, precision=3, column name=AUROC},
            every head row/.style={before row=\toprule,after row=\midrule},
            every last row/.style={after row=\bottomrule}
        ]{
            Label,Classifier,AUROC
            Atelectasis,XGBoost,0.640149
            Cardiomegaly,XGBoost,0.712454
            Consolidation,XGBoost,0.627417
            Edema,XGBoost,0.741089
            Enlarged Cardiomediastinum,XGBoost,0.573961
            Fracture,XGBoost,0.737397
            Lung Lesion,XGBoost,0.800435
            Lung Opacity,XGBoost,0.6262
            Pleural Effusion,XGBoost,0.751253
            Pleural Other,XGBoost,0.703666
            Pneumonia,XGBoost,0.581111
            Pneumothorax,XGBoost,0.735548
            Support Devices,XGBoost,0.675367
        }
	\label{table:xgboost_performance}
\end{table}

\begin{table}[h]
\centering
\caption{Adapted CheXpert phrases used for prior mention detection. Since the original CheXpert phrases were designed for radiology reports and were not fully applicable to the clinical notes we used, we adapted the original phrase list primarily by removing terms  to better align with our clinical context, where \st{strikethrough} indicates a term we removed. Due to excessive processing time, we did not perform classification of mentions (e.g., into positive or negative mentions) since the Chexpert labeler relies on a syntactic parser (BLLIP parser) to process the entire note, which generates an excessive number of possible parse tree structures. This process was not feasible to run on discharge summaries, even when broken down into individual sentences. Instead, we performed direct substring matching. To make the provided terms suitable for substring matching, we also removed underscores (e.g., in ``drain\_'').}
\label{tab:chexpert_phrases}
\begin{tabular}{|l|p{10cm}|}
\hline
\textbf{Label} & \textbf{Adapted Phrases} \\
\hline
Atelectasis & atelecta, \st{collapse} \\
\hline
Cardiomegaly & 
cardiomegaly, 
\st{the heart},
heart size,
cardiac enlargement,
cardiac size,
cardiac shadow,
cardiac contour,
cardiac silhouette,
enlarged heart
\\
\hline
Consolidation & consolidat\\
\hline
Edema & 
edema,
heart failure,
chf,
vascular congestion,
pulmonary congestion,
\st{indistinctness},
vascular prominence
\\
\hline
Enlarged Cardiomediastinum & 
\st{\_}mediastinum,
cardiomediastinum,
\st{contour},
mediastinal configuration,
mediastinal silhouette,
pericardial silhouette,
cardiac silhouette and vascularity
\\
\hline
Fracture & fracture \\
\hline
Lung Lesion 
& 
\st{mass},
nodular density,
nodular densities,
nodular opacity,
nodular opacities,
nodular opacification,
nodule,
\st{lump},
cavitary lesion,
carcinoma,
neoplasm,
tumor
\\
\hline
Lung Opacity & 
opaci,
decreased translucency,
\st{increased density},
airspace disease,
air-space disease,
air space disease,
infiltrate,
infiltration,
interstitial marking,
interstitial pattern,
interstitial lung,
reticular pattern,
reticular marking,
reticulation,
parenchymal scarring,
peribronchial thickening,
\st{wall thickening},
\st{scar} 
\\
\hline
Pleural Effusion 
&
pleural fluid,
effusion
\\
\hline
Pleural Other
&
pleural thickening,
\st{fibrosis},
fibrothorax,
pleural scar,
pleural parenchymal scar,
pleuro-parenchymal scar,
pleuro-pericardial scar
\\
\hline
Pneumonia &
pneumonia,
\st{infection},
\st{infectious process},
\st{infectious}
\\
\hline
Pneumothorax
&
pneumothorax,
pneumothoraces
\\
\hline
Support Devices
&
pacer,
\st{\_line\_},
\st{lines},
picc,
tube,
valve,
catheter,
pacemaker,
hardware,
arthroplast,
marker,
icd,
defib,
device,
drain\st{\_},
plate,
screw,
cannula,
apparatus,
coil,
\st{support},
equipment,
mediport
\\
\hline
\end{tabular}
\end{table}

\clearpage
\section{Additional Experimental Results}
\label{sec:appendix_experimental}

\begin{table*}[htbp]
\centering
\caption{Held-out performance (in terms of \textbf{AUROC}) of CXR models across sub-populations stratified by pre-test probability (LM embeddings) of the CXR label. Labels marked with an asterisk (*) indicate a statistically significant difference between the Bottom 25\% and Top 25\% groups.}
\label{tab:quantile_lm_auroc}
\footnotesize 
\setlength{\tabcolsep}{12pt} 
\begin{tabular}{@{}l ccc@{}}
\toprule
\textbf{Label} & \textbf{Bottom 25\%} & \textbf{Middle 50\%} & \textbf{Top 25\%} \\
\midrule
Atelectasis* & \num[round-mode=places,round-precision=3]{0.8307964542508125} \scriptsize(\num[round-mode=places,round-precision=3]{0.8141799837350845}--\num[round-mode=places,round-precision=3]{0.8470166206359864}) & \num[round-mode=places,round-precision=3]{0.7952889112651348} \scriptsize(\num[round-mode=places,round-precision=3]{0.784314575791359}--\num[round-mode=places,round-precision=3]{0.8061108008027077}) & \num[round-mode=places,round-precision=3]{0.738381933939457} \scriptsize(\num[round-mode=places,round-precision=3]{0.7235927537083626}--\num[round-mode=places,round-precision=3]{0.752700737118721}) \\
Cardiomegaly* & \num[round-mode=places,round-precision=3]{0.8090304223001004} \scriptsize(\num[round-mode=places,round-precision=3]{0.7867073968052865}--\num[round-mode=places,round-precision=3]{0.8311660125851631}) & \num[round-mode=places,round-precision=3]{0.7859922852754593} \scriptsize(\num[round-mode=places,round-precision=3]{0.7744529545307159}--\num[round-mode=places,round-precision=3]{0.7971635147929191}) & \num[round-mode=places,round-precision=3]{0.7281075172424316} \scriptsize(\num[round-mode=places,round-precision=3]{0.7136339977383613}--\num[round-mode=places,round-precision=3]{0.742082080245018}) \\
Consolidation* & \num[round-mode=places,round-precision=3]{0.8537479230225086} \scriptsize(\num[round-mode=places,round-precision=3]{0.8123035937547683}--\num[round-mode=places,round-precision=3]{0.8915153384208679}) & \num[round-mode=places,round-precision=3]{0.8004199338674546} \scriptsize(\num[round-mode=places,round-precision=3]{0.7786812290549279}--\num[round-mode=places,round-precision=3]{0.821283009648323}) & \num[round-mode=places,round-precision=3]{0.7775449517130851} \scriptsize(\num[round-mode=places,round-precision=3]{0.7518688857555389}--\num[round-mode=places,round-precision=3]{0.8014472216367722}) \\
Edema* & \num[round-mode=places,round-precision=3]{0.9001759154915809} \scriptsize(\num[round-mode=places,round-precision=3]{0.8742916435003281}--\num[round-mode=places,round-precision=3]{0.9243751347064972}) & \num[round-mode=places,round-precision=3]{0.8868543323218823} \scriptsize(\num[round-mode=places,round-precision=3]{0.8755547136068345}--\num[round-mode=places,round-precision=3]{0.8975969269871712}) & \num[round-mode=places,round-precision=3]{0.811701034706831} \scriptsize(\num[round-mode=places,round-precision=3]{0.7981741219758988}--\num[round-mode=places,round-precision=3]{0.8250798374414444}) \\
Enlarged Cardiomediastinum & \num[round-mode=places,round-precision=3]{0.706056204867363} \scriptsize(\num[round-mode=places,round-precision=3]{0.6526718720793724}--\num[round-mode=places,round-precision=3]{0.7565792113542557}) & \num[round-mode=places,round-precision=3]{0.7245896056771278} \scriptsize(\num[round-mode=places,round-precision=3]{0.6936694413423539}--\num[round-mode=places,round-precision=3]{0.7551746144890785}) & \num[round-mode=places,round-precision=3]{0.6575177795171737} \scriptsize(\num[round-mode=places,round-precision=3]{0.6151051163673401}--\num[round-mode=places,round-precision=3]{0.6986228883266449}) \\
Fracture & \num[round-mode=places,round-precision=3]{0.6277390860766172} \scriptsize(\num[round-mode=places,round-precision=3]{0.5447914153337479}--\num[round-mode=places,round-precision=3]{0.709368060529232}) & \num[round-mode=places,round-precision=3]{0.6805097050011158} \scriptsize(\num[round-mode=places,round-precision=3]{0.6303301706910134}--\num[round-mode=places,round-precision=3]{0.7288821905851364}) & \num[round-mode=places,round-precision=3]{0.6599817522764206} \scriptsize(\num[round-mode=places,round-precision=3]{0.6157479420304298}--\num[round-mode=places,round-precision=3]{0.7029301464557648}) \\
Lung Lesion & \num[round-mode=places,round-precision=3]{0.7330825163424015} \scriptsize(\num[round-mode=places,round-precision=3]{0.6679522082209587}--\num[round-mode=places,round-precision=3]{0.7938495144248009}) & \num[round-mode=places,round-precision=3]{0.704333219307661} \scriptsize(\num[round-mode=places,round-precision=3]{0.666653199493885}--\num[round-mode=places,round-precision=3]{0.741900771856308}) & \num[round-mode=places,round-precision=3]{0.6953035536408424} \scriptsize(\num[round-mode=places,round-precision=3]{0.6689186662435531}--\num[round-mode=places,round-precision=3]{0.7216912880539894}) \\
Lung Opacity* & \num[round-mode=places,round-precision=3]{0.7764088233172893} \scriptsize(\num[round-mode=places,round-precision=3]{0.7563624456524849}--\num[round-mode=places,round-precision=3]{0.7957751885056495}) & \num[round-mode=places,round-precision=3]{0.7303688061654567} \scriptsize(\num[round-mode=places,round-precision=3]{0.7185405373573304}--\num[round-mode=places,round-precision=3]{0.7419829130172729}) & \num[round-mode=places,round-precision=3]{0.6920138481080532} \scriptsize(\num[round-mode=places,round-precision=3]{0.6766140177845955}--\num[round-mode=places,round-precision=3]{0.7074193522334099}) \\
Pleural Effusion* & \num[round-mode=places,round-precision=3]{0.9421324301362036} \scriptsize(\num[round-mode=places,round-precision=3]{0.9309275418519972}--\num[round-mode=places,round-precision=3]{0.952572600543499}) & \num[round-mode=places,round-precision=3]{0.9092278953433036} \scriptsize(\num[round-mode=places,round-precision=3]{0.9021611049771308}--\num[round-mode=places,round-precision=3]{0.9160274311900138}) & \num[round-mode=places,round-precision=3]{0.8763795358121396} \scriptsize(\num[round-mode=places,round-precision=3]{0.8672360047698021}--\num[round-mode=places,round-precision=3]{0.8853559508919716}) \\
Pleural Other & \num[round-mode=places,round-precision=3]{0.8367779944717884} \scriptsize(\num[round-mode=places,round-precision=3]{0.7867583379149437}--\num[round-mode=places,round-precision=3]{0.8835798367857932}) & \num[round-mode=places,round-precision=3]{0.7981138274788856} \scriptsize(\num[round-mode=places,round-precision=3]{0.7482965067028999}--\num[round-mode=places,round-precision=3]{0.8428575724363327}) & \num[round-mode=places,round-precision=3]{0.7588110817790031} \scriptsize(\num[round-mode=places,round-precision=3]{0.7104740634560585}--\num[round-mode=places,round-precision=3]{0.8050803959369659}) \\
Pneumonia & \num[round-mode=places,round-precision=3]{0.7080390561163425} \scriptsize(\num[round-mode=places,round-precision=3]{0.6739625349640846}--\num[round-mode=places,round-precision=3]{0.7409050449728966}) & \num[round-mode=places,round-precision=3]{0.6966639891922474} \scriptsize(\num[round-mode=places,round-precision=3]{0.6765868633985519}--\num[round-mode=places,round-precision=3]{0.7164712101221085}) & \num[round-mode=places,round-precision=3]{0.6933436700940132} \scriptsize(\num[round-mode=places,round-precision=3]{0.6689333200454712}--\num[round-mode=places,round-precision=3]{0.7170596733689308}) \\
Pneumothorax & \num[round-mode=places,round-precision=3]{0.8622637165606022} \scriptsize(\num[round-mode=places,round-precision=3]{0.8261392772197723}--\num[round-mode=places,round-precision=3]{0.8961870849132538}) & \num[round-mode=places,round-precision=3]{0.8804438649535179} \scriptsize(\num[round-mode=places,round-precision=3]{0.8611136451363564}--\num[round-mode=places,round-precision=3]{0.8988976657390595}) & \num[round-mode=places,round-precision=3]{0.8417357900857926} \scriptsize(\num[round-mode=places,round-precision=3]{0.8202489286661148}--\num[round-mode=places,round-precision=3]{0.8621016338467598}) \\
Support Devices* & \num[round-mode=places,round-precision=3]{0.9180815553605556} \scriptsize(\num[round-mode=places,round-precision=3]{0.9026956468820572}--\num[round-mode=places,round-precision=3]{0.9321084275841712}) & \num[round-mode=places,round-precision=3]{0.9038698417425156} \scriptsize(\num[round-mode=places,round-precision=3]{0.8961236208677292}--\num[round-mode=places,round-precision=3]{0.911299140751362}) & \num[round-mode=places,round-precision=3]{0.8578011896073818} \scriptsize(\num[round-mode=places,round-precision=3]{0.8470763817429543}--\num[round-mode=places,round-precision=3]{0.8684714674949646}) \\
\bottomrule
\end{tabular}
\end{table*}

\begin{table}[htbp]
\centering
\caption{Held-out performance (in terms of \textbf{sensitivity at local 95\% specificity}) of CXR models across sub-populations stratified by pre-test probability  (LM embeddings) of the CXR label. Labels marked with an asterisk (*) indicate a statistically significant difference between the Bottom 25\% and Top 25\% groups.}
\label{tab:quantile_lm_sens_local}
\footnotesize 
\setlength{\tabcolsep}{12pt} 
\begin{tabular}{@{}l ccc@{}}
\toprule
\textbf{Label} & \textbf{Bottom 25\%} & \textbf{Middle 50\%} & \textbf{Top 25\%} \\
\midrule
\begin{filecontents*}{csv/quantile/0201/quantile_evaluation_summary_Feb01_Sens_at_Spec95.csv}
Label,bottom_25_Mean,bottom_25_CI_Lower,bottom_25_CI_Upper,middle_50_Mean,middle_50_CI_Lower,middle_50_CI_Upper,top_25_Mean,top_25_CI_Lower,top_25_CI_Upper,sig_bottom_25_vs_middle_50,sig_bottom_25_vs_top_25,sig_middle_50_vs_top_25
Atelectasis,0.3298489361702128,0.2606382978723404,0.4024822695035461,0.2762286199095022,0.2409502262443439,0.3133484162895927,0.1838924738675958,0.1430166175288126,0.2264808362369338,,*,*
Cardiomegaly,0.2932485955056179,0.2162921348314606,0.3792134831460674,0.2683272385723231,0.2254226675015654,0.3148222628967774,0.1859726011560693,0.1468208092485549,0.2236994219653179,,,*
Consolidation,0.4396527472527472,0.2857142857142857,0.6153846153846154,0.2473713864306784,0.1769911504424778,0.3274336283185841,0.2320137254901961,0.15359477124183,0.3137254901960784,,,
Edema,0.5948287037037037,0.4537037037037037,0.7314814814814815,0.4471062154696132,0.3895027624309392,0.505524861878453,0.2985743324720068,0.2448138209766116,0.3540051679586563,,*,*
Enlarged Cardiomediastinum,0.1785677083333333,0.0729166666666666,0.2916666666666667,0.2011748,0.1289153846153846,0.284,0.1071767295597484,0.0440251572327044,0.1886792452830188,,,
Fracture,0.1878799999999999,0.0273076923076923,0.3777777777777777,0.1869427419354838,0.0887096774193548,0.2983870967741935,0.1817391534391534,0.1005291005291005,0.2751322751322751,,,
Lung Lesion,0.1200568181818181,0.0,0.2954545454545454,0.1778628048780487,0.0975609756097561,0.274390243902439,0.1765955555555555,0.1185185185185185,0.2395061728395061,,,
Lung Opacity,0.28404265625,0.2234375,0.346875,0.1924366281986954,0.1575514300050175,0.227295534370296,0.1674818523153942,0.1265512178684894,0.2076164917685566,,*,
Pleural Effusion,0.6511307692307693,0.5608974358974359,0.7467948717948718,0.587639809976247,0.5416764114745113,0.6350456787867713,0.4424615514333895,0.3655143338954469,0.5193929173693086,,*,*
Pleural Other,0.2856599999999999,0.0857142857142857,0.5142857142857142,0.3055055555555556,0.1527777777777778,0.4722222222222222,0.2770419047619047,0.1428571428571428,0.419047619047619,,,
Pneumonia,0.218509505703422,0.1482889733840304,0.2965779467680608,0.174328837876614,0.1291248206599713,0.2209469153515064,0.1889148975791434,0.1378026070763501,0.2420856610800744,,,
Pneumothorax,0.4263,0.2631578947368421,0.5921052631578947,0.5086448148148147,0.4148148148148148,0.5962962962962963,0.447524504950495,0.3712871287128713,0.5321782178217822,,,
Support Devices,0.7139830303030303,0.646926961926962,0.7797979797979798,0.5951011330049262,0.545320197044335,0.639296134899583,0.4123214064115822,0.3417786970010341,0.4885062843051466,*,*,*
\end{filecontents*}
\csvreader[
  head=true, 
  late after line=\\
]{csv/quantile/0201/quantile_evaluation_summary_Feb01_Sens_at_Spec95.csv}{}{%
  \csvcoli\textsuperscript{\csvcolxii} & 
  \num[round-mode=places,round-precision=3]{\csvcolii} \scriptsize(\num[round-mode=places,round-precision=3]{\csvcoliii}--\num[round-mode=places,round-precision=3]{\csvcoliv}) &
  \num[round-mode=places,round-precision=3]{\csvcolv} \scriptsize(\num[round-mode=places,round-precision=3]{\csvcolvi}--\num[round-mode=places,round-precision=3]{\csvcolvii}) &
  \num[round-mode=places,round-precision=3]{\csvcolviii} \scriptsize(\num[round-mode=places,round-precision=3]{\csvcolix}--\num[round-mode=places,round-precision=3]{\csvcolx})%
}
\bottomrule
\end{tabular}
\end{table}

\begin{table}[htbp]
\centering
\caption{Held-out performance (in terms of \textbf{sensitivity at global 95\% specificity}) of CXR models across sub-populations stratified by pre-test probability (LM embeddings)  of the CXR label. Labels marked with an asterisk (*) indicate a statistically significant difference between the Bottom 25\% and Top 25\% groups.}
\label{tab:quantile_lm_sens_global}
\footnotesize 
\setlength{\tabcolsep}{12pt} 
\begin{tabular}{@{}l ccc@{}}
\toprule
\textbf{Label} & \textbf{Bottom 25\%} & \textbf{Middle 50\%} & \textbf{Top 25\%} \\
\midrule
\begin{filecontents*}{csv/quantile/0201/quantile_evaluation_summary_Feb01_Sens_at_Global_Spec95.csv}
Label,bottom_25_Mean,bottom_25_CI_Lower,bottom_25_CI_Upper,middle_50_Mean,middle_50_CI_Lower,middle_50_CI_Upper,top_25_Mean,top_25_CI_Lower,top_25_CI_Upper,sig_bottom_25_vs_middle_50,sig_bottom_25_vs_top_25,sig_middle_50_vs_top_25
Atelectasis,0.1917338652482269,0.1436170212765957,0.2446808510638297,0.2348963800904977,0.2014869039331709,0.269666942220675,0.3288807665505226,0.2871080139372822,0.3721254355400696,,*,*
Cardiomegaly,0.0465485955056179,0.0168539325842696,0.0842696629213483,0.193460237946149,0.1615529117094552,0.2254226675015654,0.4028903468208092,0.3658959537572254,0.4426422854602043,*,*,*
Consolidation,0.3130846153846153,0.1758241758241758,0.4615384615384615,0.2072194690265486,0.1422679827547084,0.2772861356932153,0.3955052287581699,0.3112053795877325,0.4803921568627451,,,*
Edema,0.2462166666666667,0.1296296296296296,0.3775106837606809,0.390546408839779,0.3331890140246494,0.4485789417764552,0.5206,0.4737295434969854,0.5650301464254953,,*,*
Enlarged Cardiomediastinum,0.14110625,0.0520833333333333,0.25,0.1940752,0.124,0.268,0.1433226415094339,0.0691823899371069,0.2327044025157232,,,
Fracture,0.1544133333333333,0.0222222222222222,0.3333333333333333,0.1774201612903226,0.0887096774193548,0.2741935483870967,0.2281920634920634,0.1428571428571428,0.3174603174603174,,,
Lung Lesion,0.0450727272727272,0.0,0.1590909090909091,0.1513268292682926,0.073170731707317,0.2378048780487804,0.34758,0.2795773979107313,0.4172839506172839,,*,*
Lung Opacity,0.12781875,0.090625,0.1715174278846149,0.1659109382839939,0.1406065459878806,0.1936778725539387,0.2860332290362954,0.2471839799749687,0.3265151150476555,,*,*
Pleural Effusion,0.3908849358974359,0.3108974358974359,0.4704203648915177,0.5470943942992874,0.5088973140873378,0.5866983372921615,0.6903096121416525,0.6548182319366974,0.7262947528862367,*,*,*
Pleural Other,0.1715742857142857,0.0285714285714285,0.3714285714285714,0.2232263888888889,0.0972222222222222,0.375,0.4008133333333333,0.2666666666666666,0.5406776556776527,,,
Pneumonia,0.1650764258555133,0.1026615969581749,0.2357414448669201,0.1605434720229555,0.1190817790530846,0.2062713828495746,0.2580877094972067,0.2048417132216015,0.3147113594040968,,,
Pneumothorax,0.2681065789473684,0.1184210526315789,0.4342105263157895,0.433807037037037,0.3481481481481481,0.5222222222222223,0.6005962871287128,0.5297029702970297,0.6732673267326733,,*,*
Support Devices,0.4822694949494949,0.4161616161616162,0.5494949494949495,0.5492552709359606,0.5104575596816976,0.5900350511557406,0.6419121509824197,0.6018614270941055,0.6835573940020683,,*,*
\end{filecontents*}
\csvreader[
  head=true, 
  late after line=\\
]{csv/quantile/0201/quantile_evaluation_summary_Feb01_Sens_at_Global_Spec95.csv}{}{%
  \csvcoli\textsuperscript{\csvcolxii} & 
  \num[round-mode=places,round-precision=3]{\csvcolii} \scriptsize(\num[round-mode=places,round-precision=3]{\csvcoliii}--\num[round-mode=places,round-precision=3]{\csvcoliv}) &
  \num[round-mode=places,round-precision=3]{\csvcolv} \scriptsize(\num[round-mode=places,round-precision=3]{\csvcolvi}--\num[round-mode=places,round-precision=3]{\csvcolvii}) &
  \num[round-mode=places,round-precision=3]{\csvcolviii} \scriptsize(\num[round-mode=places,round-precision=3]{\csvcolix}--\num[round-mode=places,round-precision=3]{\csvcolx})%
}
\bottomrule
\end{tabular}
\end{table}

\begin{table}[htbp]
\centering
\caption{Held-out performance (in terms of \textbf{sensitivity at local 95\% specificity}) of CXR models across sub-populations stratified by pre-test probability  (LM embeddings) of the CXR label. Labels marked with an asterisk (*) indicate a statistically significant difference between the Bottom 25\% and Top 25\% groups.}
\label{tab:quantile_lm_spec_local}
\footnotesize 
\setlength{\tabcolsep}{12pt} 
\begin{tabular}{@{}l ccc@{}}
\toprule
\textbf{Label} & \textbf{Bottom 25\%} & \textbf{Middle 50\%} & \textbf{Top 25\%} \\
\midrule
\csvreader[
  head=true, 
  late after line=\\
]{csv/quantile/0201/quantile_evaluation_summary_Feb01_Sens_at_Spec95.csv}{}{%
  \csvcoli\textsuperscript{\csvcolxii} & 
  \num[round-mode=places,round-precision=3]{\csvcolii} \scriptsize(\num[round-mode=places,round-precision=3]{\csvcoliii}--\num[round-mode=places,round-precision=3]{\csvcoliv}) &
  \num[round-mode=places,round-precision=3]{\csvcolv} \scriptsize(\num[round-mode=places,round-precision=3]{\csvcolvi}--\num[round-mode=places,round-precision=3]{\csvcolvii}) &
  \num[round-mode=places,round-precision=3]{\csvcolviii} \scriptsize(\num[round-mode=places,round-precision=3]{\csvcolix}--\num[round-mode=places,round-precision=3]{\csvcolx})%
}
\bottomrule
\end{tabular}
\end{table}

\begin{table}[htbp]
\centering
\caption{Held-out performance (in terms of \textbf{specificity at global 95\% specificity}) of CXR models across sub-populations stratified by pre-test probability (LM embeddings)  of the CXR label. Labels marked with an asterisk (*) indicate a statistically significant difference between the Bottom 25\% and Top 25\% groups.}
\label{tab:quantile_lm_spec_global}
\footnotesize 
\setlength{\tabcolsep}{12pt} 
\begin{tabular}{@{}l ccc@{}}
\toprule
\textbf{Label} & \textbf{Bottom 25\%} & \textbf{Middle 50\%} & \textbf{Top 25\%} \\
\midrule
\begin{filecontents*}{csv/quantile/0201/quantile_evaluation_summary_Feb01_Spec_at_Global_Spec95.csv}
Label,bottom_25_Mean,bottom_25_CI_Lower,bottom_25_CI_Upper,middle_50_Mean,middle_50_CI_Lower,middle_50_CI_Upper,top_25_Mean,top_25_CI_Lower,top_25_CI_Upper,sig_bottom_25_vs_middle_50,sig_bottom_25_vs_top_25,sig_middle_50_vs_top_25
Atelectasis,0.9777323916325212,0.9717489756307958,0.98361009273237,0.961899128109742,0.9572192513368984,0.9666356661241572,0.8886038257173221,0.8775847093926266,0.8993092454835282,*,*,*
Cardiomegaly,0.9929863871101012,0.989258417682297,0.996075191076224,0.967268598428783,0.9628828418535808,0.971649777980189,0.8461278371354317,0.8343138452653325,0.8573491192607565,*,*,*
Consolidation,0.9771223614646564,0.9712159780693166,0.9825729391031915,0.9594040342663612,0.9548759836637116,0.9638410200219144,0.9023502659574468,0.8932078559738135,0.911620294599018,*,*,*
Edema,0.9912934526150216,0.9874164372788046,0.9946913094769956,0.9614318836077456,0.9569224396810604,0.965804199424889,0.8706087010411502,0.8597493421303536,0.8815071888943976,*,*,*
Enlarged Cardiomediastinum,0.96776101196313,0.9611688566385566,0.974112571092371,0.952514244817374,0.9480750246791708,0.957035651909788,0.9269461477362988,0.9187847498014295,0.9350675138999206,*,*,*
Fracture,0.9646484777971688,0.957727360868722,0.9714950552646888,0.9579895219512194,0.9536585365853658,0.962439024390244,0.9185275724275724,0.9094905094905096,0.9274268039652654,,*,*
Lung Lesion,0.9807684016313848,0.9751408040396192,0.9858224898038456,0.9685280751541246,0.9644779332615716,0.97260005871416,0.8774149300480268,0.8682397160158697,0.8872415953226144,*,*,*
Lung Opacity,0.9802942675159236,0.9743026575884032,0.9855040632549968,0.9575761745766754,0.952897686620558,0.9624373956594324,0.894105361111111,0.8833333333333333,0.905,*,*,*
Pleural Effusion,0.9842405366653012,0.979311757476444,0.9889389594428512,0.9572680923039748,0.9525190286335627,0.9621567178743692,0.8693644932671863,0.8550673281360737,0.8830616583982991,*,*,*
Pleural Other,0.9821912790697672,0.976937984496124,0.987015503875969,0.9678749757516976,0.9636275460717748,0.9722377452809072,0.8811730542452829,0.872248427672956,0.8903301886792453,*,*,*
Pneumonia,0.9661350101419878,0.959026369168357,0.9724137931034482,0.9538380446004544,0.9493082799917406,0.9583935577121618,0.9249275713978956,0.9160403693364828,0.9336482714193688,*,*,*
Pneumothorax,0.9737574247753028,0.9675654552559594,0.9796348483482128,0.9684365373193426,0.9644624826766977,0.972359053043571,0.8858071458333334,0.8766666666666667,0.8947916666666667,,*,*
Support Devices,0.98257564293305,0.9770456960680128,0.987672688629118,0.9581150832634476,0.9531568228105908,0.9628334915354204,0.8821350721080086,0.869898741945382,0.8943759293789977,*,*,*
\end{filecontents*}
\csvreader[
  head=true, 
  late after line=\\
]{csv/quantile/0201/quantile_evaluation_summary_Feb01_Spec_at_Global_Spec95.csv}{}{%
  \csvcoli\textsuperscript{\csvcolxii} & 
  \num[round-mode=places,round-precision=3]{\csvcolii} \scriptsize(\num[round-mode=places,round-precision=3]{\csvcoliii}--\num[round-mode=places,round-precision=3]{\csvcoliv}) &
  \num[round-mode=places,round-precision=3]{\csvcolv} \scriptsize(\num[round-mode=places,round-precision=3]{\csvcolvi}--\num[round-mode=places,round-precision=3]{\csvcolvii}) &
  \num[round-mode=places,round-precision=3]{\csvcolviii} \scriptsize(\num[round-mode=places,round-precision=3]{\csvcolix}--\num[round-mode=places,round-precision=3]{\csvcolx})%
}
\bottomrule
\end{tabular}
\end{table}

\begin{table}[htbp]
\centering
\caption{Held-out performance (in terms of \textbf{Adaptive Calibration Error (ACE)}) of CXR models across sub-populations stratified by pre-test probability (LM embeddings) of the CXR label.}
\label{tab:quantile_lm_ace}
\footnotesize 
\setlength{\tabcolsep}{12pt} 
\begin{tabular}{@{}l ccc@{}}
\toprule
\textbf{Label} & \textbf{Bottom 25\%} & \textbf{Middle 50\%} & \textbf{Top 25\%} \\
\midrule
Atelectasis & \num[round-mode=places,round-precision=3]{0.0393393961261354} \scriptsize(\num[round-mode=places,round-precision=3]{0.0322882737855663}--\num[round-mode=places,round-precision=3]{0.0467592420108043}) & \num[round-mode=places,round-precision=3]{0.0298768781215185} \scriptsize(\num[round-mode=places,round-precision=3]{0.0239172873411977}--\num[round-mode=places,round-precision=3]{0.0359378614763952}) & \num[round-mode=places,round-precision=3]{0.0457786527413866} \scriptsize(\num[round-mode=places,round-precision=3]{0.0356347141800705}--\num[round-mode=places,round-precision=3]{0.0561594195290378}) \\
Cardiomegaly & \num[round-mode=places,round-precision=3]{0.0309824360436313} \scriptsize(\num[round-mode=places,round-precision=3]{0.0265262253201263}--\num[round-mode=places,round-precision=3]{0.0365646668850878}) & \num[round-mode=places,round-precision=3]{0.0267946370075221} \scriptsize(\num[round-mode=places,round-precision=3]{0.0210425488997308}--\num[round-mode=places,round-precision=3]{0.0327366386506498}) & \num[round-mode=places,round-precision=3]{0.0595598939057585} \scriptsize(\num[round-mode=places,round-precision=3]{0.0492024606068857}--\num[round-mode=places,round-precision=3]{0.0702028348481505}) \\
Consolidation & \num[round-mode=places,round-precision=3]{0.0134856738998889} \scriptsize(\num[round-mode=places,round-precision=3]{0.0111461917881599}--\num[round-mode=places,round-precision=3]{0.0159306179602142}) & \num[round-mode=places,round-precision=3]{0.0142130690944892} \scriptsize(\num[round-mode=places,round-precision=3]{0.0119429725908911}--\num[round-mode=places,round-precision=3]{0.0165385144192}) & \num[round-mode=places,round-precision=3]{0.0179764217132426} \scriptsize(\num[round-mode=places,round-precision=3]{0.0147590631652693}--\num[round-mode=places,round-precision=3]{0.0213145565907646}) \\
Edema & \num[round-mode=places,round-precision=3]{0.0244735142159366} \scriptsize(\num[round-mode=places,round-precision=3]{0.0230344350216926}--\num[round-mode=places,round-precision=3]{0.0260883447334712}) & \num[round-mode=places,round-precision=3]{0.028645006318598} \scriptsize(\num[round-mode=places,round-precision=3]{0.0243723015758964}--\num[round-mode=places,round-precision=3]{0.0329668404502607}) & \num[round-mode=places,round-precision=3]{0.0646583098215294} \scriptsize(\num[round-mode=places,round-precision=3]{0.0564602180071118}--\num[round-mode=places,round-precision=3]{0.0729225600859241}) \\
Enlarged Cardiomediastinum & \num[round-mode=places,round-precision=3]{0.0076741382223488} \scriptsize(\num[round-mode=places,round-precision=3]{0.0055125144405029}--\num[round-mode=places,round-precision=3]{0.0104215576500927}) & \num[round-mode=places,round-precision=3]{0.0071443341216903} \scriptsize(\num[round-mode=places,round-precision=3]{0.0050968734631835}--\num[round-mode=places,round-precision=3]{0.0093925239198153}) & \num[round-mode=places,round-precision=3]{0.0090093912969809} \scriptsize(\num[round-mode=places,round-precision=3]{0.0053041016797788}--\num[round-mode=places,round-precision=3]{0.0131136675993642}) \\
Fracture & \num[round-mode=places,round-precision=3]{0.0076884841808668} \scriptsize(\num[round-mode=places,round-precision=3]{0.006602321840649}--\num[round-mode=places,round-precision=3]{0.0095140780090805}) & \num[round-mode=places,round-precision=3]{0.0049159652629936} \scriptsize(\num[round-mode=places,round-precision=3]{0.0040276963592248}--\num[round-mode=places,round-precision=3]{0.0063324581741944}) & \num[round-mode=places,round-precision=3]{0.0165535333774983} \scriptsize(\num[round-mode=places,round-precision=3]{0.0156710457844315}--\num[round-mode=places,round-precision=3]{0.0178092160003685}) \\
Lung Lesion & \num[round-mode=places,round-precision=3]{0.0177399838046737} \scriptsize(\num[round-mode=places,round-precision=3]{0.0172158765464161}--\num[round-mode=places,round-precision=3]{0.0184400508360268}) & \num[round-mode=places,round-precision=3]{0.0128545729296692} \scriptsize(\num[round-mode=places,round-precision=3]{0.0123331518017986}--\num[round-mode=places,round-precision=3]{0.0139832950270486}) & \num[round-mode=places,round-precision=3]{0.0332443220513995} \scriptsize(\num[round-mode=places,round-precision=3]{0.0306074274599055}--\num[round-mode=places,round-precision=3]{0.0363620115957155}) \\
Lung Opacity & \num[round-mode=places,round-precision=3]{0.0325105382378014} \scriptsize(\num[round-mode=places,round-precision=3]{0.0277446721994525}--\num[round-mode=places,round-precision=3]{0.0388925104625959}) & \num[round-mode=places,round-precision=3]{0.027107603297427} \scriptsize(\num[round-mode=places,round-precision=3]{0.0208844580161891}--\num[round-mode=places,round-precision=3]{0.033777264729384}) & \num[round-mode=places,round-precision=3]{0.0477497447238358} \scriptsize(\num[round-mode=places,round-precision=3]{0.0379101684524956}--\num[round-mode=places,round-precision=3]{0.058082759846237}) \\
Pleural Effusion & \num[round-mode=places,round-precision=3]{0.0392138639491689} \scriptsize(\num[round-mode=places,round-precision=3]{0.032893144858392}--\num[round-mode=places,round-precision=3]{0.0452066345703474}) & \num[round-mode=places,round-precision=3]{0.0371164087977818} \scriptsize(\num[round-mode=places,round-precision=3]{0.0317319183395117}--\num[round-mode=places,round-precision=3]{0.0426986983885061}) & \num[round-mode=places,round-precision=3]{0.0571937894840381} \scriptsize(\num[round-mode=places,round-precision=3]{0.0491800170536685}--\num[round-mode=places,round-precision=3]{0.0655144638135613}) \\
Pleural Other & \num[round-mode=places,round-precision=3]{0.0069042818275546} \scriptsize(\num[round-mode=places,round-precision=3]{0.0055100986994827}--\num[round-mode=places,round-precision=3]{0.0083655808686802}) & \num[round-mode=places,round-precision=3]{0.004958592209172} \scriptsize(\num[round-mode=places,round-precision=3]{0.0039685844867487}--\num[round-mode=places,round-precision=3]{0.0061675747583206}) & \num[round-mode=places,round-precision=3]{0.0085733482309521} \scriptsize(\num[round-mode=places,round-precision=3]{0.0065372087315636}--\num[round-mode=places,round-precision=3]{0.0109648393642324}) \\
Pneumonia & \num[round-mode=places,round-precision=3]{0.0154116692007114} \scriptsize(\num[round-mode=places,round-precision=3]{0.0121989238901113}--\num[round-mode=places,round-precision=3]{0.0198285747771063}) & \num[round-mode=places,round-precision=3]{0.0158560253402716} \scriptsize(\num[round-mode=places,round-precision=3]{0.0115667283430379}--\num[round-mode=places,round-precision=3]{0.0202345080817741}) & \num[round-mode=places,round-precision=3]{0.0242003051474866} \scriptsize(\num[round-mode=places,round-precision=3]{0.0184688787485874}--\num[round-mode=places,round-precision=3]{0.030891447465981}) \\
Pneumothorax & \num[round-mode=places,round-precision=3]{0.0140785791185261} \scriptsize(\num[round-mode=places,round-precision=3]{0.0121155935648532}--\num[round-mode=places,round-precision=3]{0.0166068768432559}) & \num[round-mode=places,round-precision=3]{0.0172698782407574} \scriptsize(\num[round-mode=places,round-precision=3]{0.0150520229995836}--\num[round-mode=places,round-precision=3]{0.0195582773692128}) & \num[round-mode=places,round-precision=3]{0.0318383818452401} \scriptsize(\num[round-mode=places,round-precision=3]{0.0274654438954959}--\num[round-mode=places,round-precision=3]{0.036076829160613}) \\
Support Devices & \num[round-mode=places,round-precision=3]{0.0297868542321805} \scriptsize(\num[round-mode=places,round-precision=3]{0.0250877010898234}--\num[round-mode=places,round-precision=3]{0.0346677525251244}) & \num[round-mode=places,round-precision=3]{0.0319290595833355} \scriptsize(\num[round-mode=places,round-precision=3]{0.0266565124383231}--\num[round-mode=places,round-precision=3]{0.0372335177611505}) & \num[round-mode=places,round-precision=3]{0.0442701640765471} \scriptsize(\num[round-mode=places,round-precision=3]{0.0345080966831078}--\num[round-mode=places,round-precision=3]{0.0541776837083878}) \\
\bottomrule
\end{tabular}
\end{table}

\begin{table}[htbp]
\centering
\caption{Held-out performance (in terms of \textbf{Brier Skill Score (BSS)}) of CXR models across sub-populations stratified by pre-test probability (LM embeddings) of the CXR label.}
\label{tab:quantile_lm_bss}
\footnotesize 
\setlength{\tabcolsep}{12pt} 
\begin{tabular}{@{}l ccc@{}}
\toprule
\textbf{Label} & \textbf{Bottom 25\%} & \textbf{Middle 50\%} & \textbf{Top 25\%} \\
\midrule
Atelectasis & \num[round-mode=places,round-precision=3]{0.157268428172797} \scriptsize(\num[round-mode=places,round-precision=3]{0.1283027238649078}--\num[round-mode=places,round-precision=3]{0.1861423015203717}) & \num[round-mode=places,round-precision=3]{0.1670856564287572} \scriptsize(\num[round-mode=places,round-precision=3]{0.1501494340245197}--\num[round-mode=places,round-precision=3]{0.1841054133521962}) & \num[round-mode=places,round-precision=3]{0.1233732625228761} \scriptsize(\num[round-mode=places,round-precision=3]{0.1007817034242864}--\num[round-mode=places,round-precision=3]{0.1455258814474867}) \\
Cardiomegaly & \num[round-mode=places,round-precision=3]{0.0835289041535096} \scriptsize(\num[round-mode=places,round-precision=3]{0.058615791579411}--\num[round-mode=places,round-precision=3]{0.1091211884014581}) & \num[round-mode=places,round-precision=3]{0.142485946885212} \scriptsize(\num[round-mode=places,round-precision=3]{0.12473525929081}--\num[round-mode=places,round-precision=3]{0.1602253055577322}) & \num[round-mode=places,round-precision=3]{0.122366064182581} \scriptsize(\num[round-mode=places,round-precision=3]{0.0994682141187675}--\num[round-mode=places,round-precision=3]{0.1451001988978451}) \\
Consolidation & \num[round-mode=places,round-precision=3]{0.0358044025098319} \scriptsize(\num[round-mode=places,round-precision=3]{-0.0193626279058895}--\num[round-mode=places,round-precision=3]{0.091124022777649}) & \num[round-mode=places,round-precision=3]{0.0283774849446042} \scriptsize(\num[round-mode=places,round-precision=3]{0.0101114300174018}--\num[round-mode=places,round-precision=3]{0.0467268207718514}) & \num[round-mode=places,round-precision=3]{0.0469220882151383} \scriptsize(\num[round-mode=places,round-precision=3]{0.0198671614188475}--\num[round-mode=places,round-precision=3]{0.075013320394175}) \\
Edema & \num[round-mode=places,round-precision=3]{0.0853744308611514} \scriptsize(\num[round-mode=places,round-precision=3]{0.02711651612343}--\num[round-mode=places,round-precision=3]{0.1457443462191687}) & \num[round-mode=places,round-precision=3]{0.1903763781545727} \scriptsize(\num[round-mode=places,round-precision=3]{0.1608970602554532}--\num[round-mode=places,round-precision=3]{0.2197220396680435}) & \num[round-mode=places,round-precision=3]{0.1874931143879398} \scriptsize(\num[round-mode=places,round-precision=3]{0.1592400725400789}--\num[round-mode=places,round-precision=3]{0.2155212533739916}) \\
Enlarged Cardiomediastinum & \num[round-mode=places,round-precision=3]{-0.0066584713832251} \scriptsize(\num[round-mode=places,round-precision=3]{-0.0230842878968945}--\num[round-mode=places,round-precision=3]{0.009071325489183}) & \num[round-mode=places,round-precision=3]{0.0053172073085443} \scriptsize(\num[round-mode=places,round-precision=3]{-0.0140550362060889}--\num[round-mode=places,round-precision=3]{0.0247910043013898}) & \num[round-mode=places,round-precision=3]{-0.0271146159605042} \scriptsize(\num[round-mode=places,round-precision=3]{-0.0548048805284156}--\num[round-mode=places,round-precision=3]{-0.0036547432058617}) \\
Fracture & \num[round-mode=places,round-precision=3]{-0.006915313564144} \scriptsize(\num[round-mode=places,round-precision=3]{-0.033985067656195}--\num[round-mode=places,round-precision=3]{0.0269466348918542}) & \num[round-mode=places,round-precision=3]{0.0018864690875213} \scriptsize(\num[round-mode=places,round-precision=3]{-0.0085774243569376}--\num[round-mode=places,round-precision=3]{0.0131006953034323}) & \num[round-mode=places,round-precision=3]{-0.0032366199095136} \scriptsize(\num[round-mode=places,round-precision=3]{-0.0162217648132063}--\num[round-mode=places,round-precision=3]{0.0099312381879192}) \\
Lung Lesion & \num[round-mode=places,round-precision=3]{-0.0522387866327843} \scriptsize(\num[round-mode=places,round-precision=3]{-0.0871465123736382}--\num[round-mode=places,round-precision=3]{-0.0112594215230246}) & \num[round-mode=places,round-precision=3]{-0.0120679276669673} \scriptsize(\num[round-mode=places,round-precision=3]{-0.0264388747751114}--\num[round-mode=places,round-precision=3]{0.0023049395073853}) & \num[round-mode=places,round-precision=3]{-0.0145298801894083} \scriptsize(\num[round-mode=places,round-precision=3]{-0.0372989099090037}--\num[round-mode=places,round-precision=3]{0.007803647803063}) \\
Lung Opacity & \num[round-mode=places,round-precision=3]{0.1197646922714776} \scriptsize(\num[round-mode=places,round-precision=3]{0.0956288595452595}--\num[round-mode=places,round-precision=3]{0.1431557599062704}) & \num[round-mode=places,round-precision=3]{0.0996442531701193} \scriptsize(\num[round-mode=places,round-precision=3]{0.0851245599894072}--\num[round-mode=places,round-precision=3]{0.1136327544396551}) & \num[round-mode=places,round-precision=3]{0.0854210690041237} \scriptsize(\num[round-mode=places,round-precision=3]{0.066056985442736}--\num[round-mode=places,round-precision=3]{0.1049424761430061}) \\
Pleural Effusion & \num[round-mode=places,round-precision=3]{0.2929997258518066} \scriptsize(\num[round-mode=places,round-precision=3]{0.2379503171806706}--\num[round-mode=places,round-precision=3]{0.348029927814087}) & \num[round-mode=places,round-precision=3]{0.4363456748883654} \scriptsize(\num[round-mode=places,round-precision=3]{0.4142742337154179}--\num[round-mode=places,round-precision=3]{0.457674845876903}) & \num[round-mode=places,round-precision=3]{0.4049768272285514} \scriptsize(\num[round-mode=places,round-precision=3]{0.3789380904055869}--\num[round-mode=places,round-precision=3]{0.4305519283425166}) \\
Pleural Other & \num[round-mode=places,round-precision=3]{-0.0068549486167413} \scriptsize(\num[round-mode=places,round-precision=3]{-0.0177073779552888}--\num[round-mode=places,round-precision=3]{0.0021107469949182}) & \num[round-mode=places,round-precision=3]{0.0106290625654258} \scriptsize(\num[round-mode=places,round-precision=3]{-0.0457957980215173}--\num[round-mode=places,round-precision=3]{0.0711726472108444}) & \num[round-mode=places,round-precision=3]{-0.0032587645415976} \scriptsize(\num[round-mode=places,round-precision=3]{-0.0413311974082967}--\num[round-mode=places,round-precision=3]{0.0357545416061217}) \\
Pneumonia & \num[round-mode=places,round-precision=3]{0.0189168038682186} \scriptsize(\num[round-mode=places,round-precision=3]{-0.0082926835938754}--\num[round-mode=places,round-precision=3]{0.0468980740971163}) & \num[round-mode=places,round-precision=3]{0.0284074396559477} \scriptsize(\num[round-mode=places,round-precision=3]{0.0129806664434254}--\num[round-mode=places,round-precision=3]{0.0445403618792407}) & \num[round-mode=places,round-precision=3]{0.0350744983384786} \scriptsize(\num[round-mode=places,round-precision=3]{0.0115149600320125}--\num[round-mode=places,round-precision=3]{0.0590584187823744}) \\
Pneumothorax & \num[round-mode=places,round-precision=3]{-0.0260973735483594} \scriptsize(\num[round-mode=places,round-precision=3]{-0.1086991820118635}--\num[round-mode=places,round-precision=3]{0.0543644928629265}) & \num[round-mode=places,round-precision=3]{0.0686604747057979} \scriptsize(\num[round-mode=places,round-precision=3]{0.0188506650940498}--\num[round-mode=places,round-precision=3]{0.1179971885427476}) & \num[round-mode=places,round-precision=3]{0.1491637254058964} \scriptsize(\num[round-mode=places,round-precision=3]{0.1063704727077746}--\num[round-mode=places,round-precision=3]{0.1914868322098602}) \\
Support Devices & \num[round-mode=places,round-precision=3]{0.4274307779825664} \scriptsize(\num[round-mode=places,round-precision=3]{0.3850215908796674}--\num[round-mode=places,round-precision=3]{0.4691118099791366}) & \num[round-mode=places,round-precision=3]{0.4339814475170674} \scriptsize(\num[round-mode=places,round-precision=3]{0.4110942145588455}--\num[round-mode=places,round-precision=3]{0.4565250895535172}) & \num[round-mode=places,round-precision=3]{0.3728714347609793} \scriptsize(\num[round-mode=places,round-precision=3]{0.344760505782691}--\num[round-mode=places,round-precision=3]{0.4009868934347563}) \\
\bottomrule
\end{tabular}
\end{table}

\begin{figure}[htbp]
    \centering
    \includegraphics[width=\linewidth]{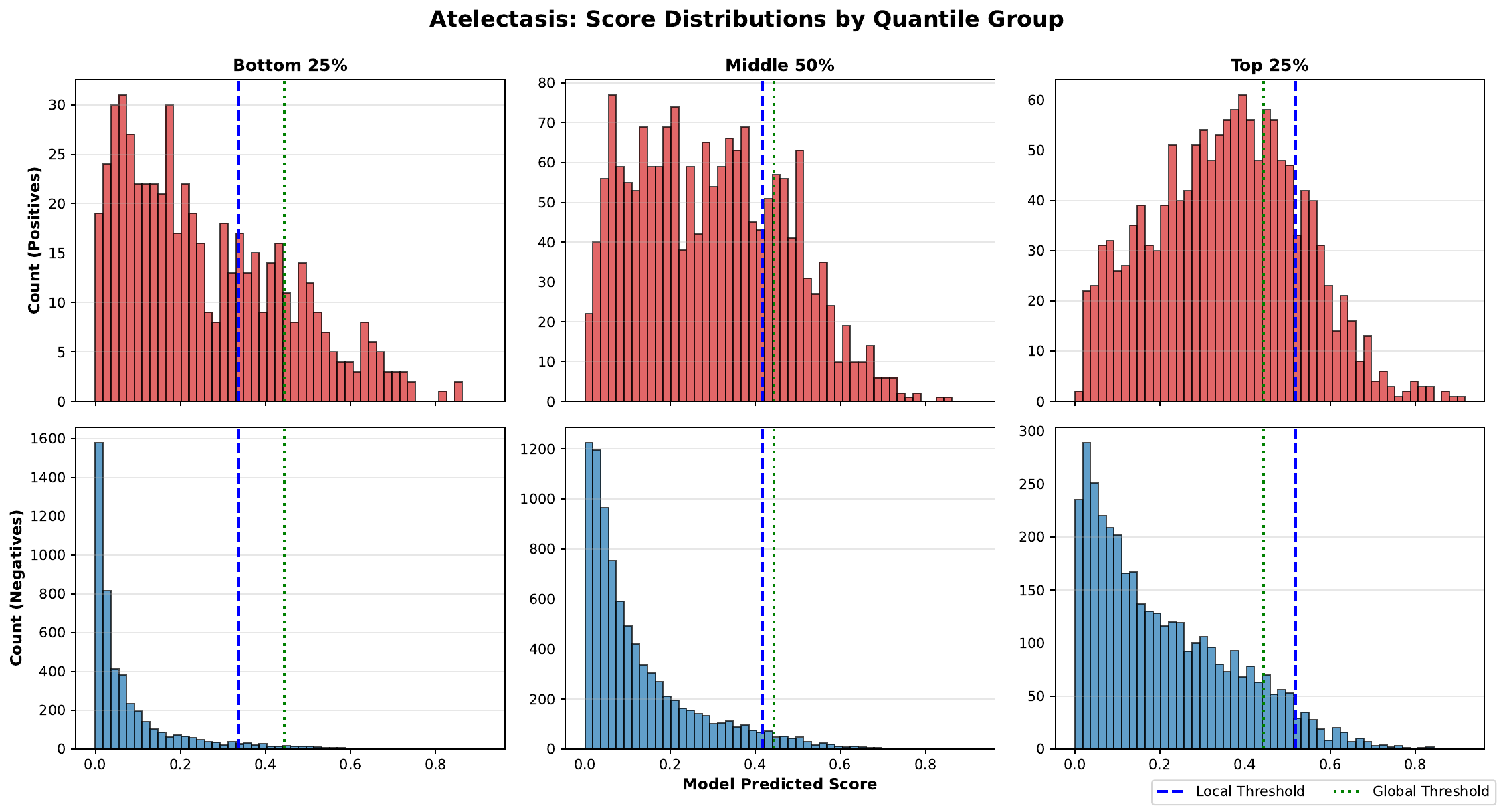}

    \vspace{0.8em}

    \includegraphics[width=\linewidth]{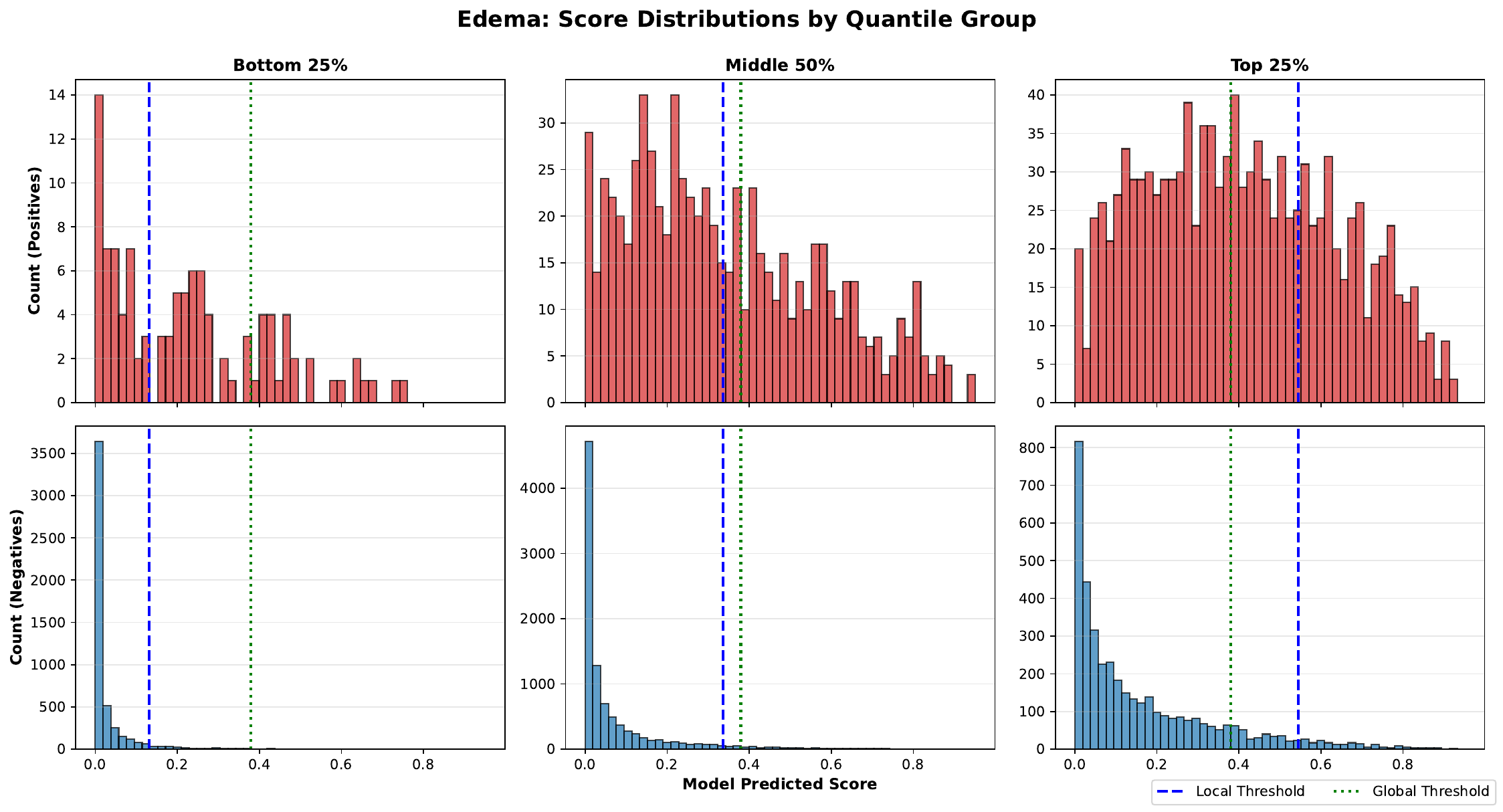}

\caption{Score distributions of the vision model predictions stratified by pre-test probability (LM embeddings) for two representative labels (Atelectasis and Edema).
Histograms show the distribution of predicted scores for positive (top row) and negative (bottom row) cases across low-, medium-, and high-risk quantile groups defined by the text-based pre-test probability. Blue dashed lines indicate subgroup-specific (local) thresholds selected at 95\% specificity within each quantile, while green dotted lines indicate a universal global threshold computed using all samples.}
    \label{fig:quantile_histograms}
\end{figure}

\begin{table}[htbp]
\centering
\caption{Held-out performance (in terms of \textbf{AUROC}) of CXR DenseNet121 models across sub-populations stratified by pre-test probability (BoW Representations) of the CXR label. Labels marked with an asterisk (*) indicate a statistically significant difference between the Bottom 25\% and Top 25\% groups.}
\label{tab:quantile_bow_auroc}
\footnotesize 
\setlength{\tabcolsep}{12pt} 
\begin{tabular}{@{}l ccc@{}}
\toprule
\textbf{Label} & \textbf{Bottom 25\%} & \textbf{Middle 50\%} & \textbf{Top 25\%} \\
\midrule
\begin{filecontents*}{csv/quantile_bow/bow_quantile_AUROC.csv}
Label,bottom_25_Mean,bottom_25_CI_Lower,bottom_25_CI_Upper,middle_50_Mean,middle_50_CI_Lower,middle_50_CI_Upper,top_25_Mean,top_25_CI_Lower,top_25_CI_Upper,sig_bottom_25_vs_middle_50,sig_bottom_25_vs_top_25,sig_middle_50_vs_top_25
Atelectasis,0.8480348587036133,0.8235766539206871,0.8718999419074791,0.7946402430534363,0.7789717459907899,0.8105898723006248,0.7246769666671753,0.7034625816803712,0.7459497259213373,*,*,*
Cardiomegaly,0.8195676803588867,0.7882294597534033,0.8495946377515793,0.7825609445571899,0.7657599472082578,0.7987763967651587,0.7421844601631165,0.7208423283237677,0.7621772842911573,,*,*
Consolidation,0.8622269630432129,0.8181142968627123,0.9002257289794775,0.8246236443519592,0.7913906700336016,0.8561141057656361,0.7506839036941528,0.7128258239764433,0.787322872533248,,*,*
Edema,0.90756756067276,0.8731596419444451,0.937791928075827,0.8887970447540283,0.8734669974217049,0.9031538579326408,0.8257596492767334,0.805489635352905,0.8450701718146985,,*,*
Enlarged Cardiomediastinum,0.7751579284667969,0.7097491001853576,0.8363641451184565,0.6957768797874451,0.6440972805023193,0.7455180533803426,0.654037594795227,0.5951067705567067,0.7105902678691421,,,
Fracture,0.6588984131813049,0.595078099003205,0.719291838315817,,,,0.6806759238243103,0.6373995641103157,0.722657505594767,,,
Lung Lesion,0.6762468814849854,0.5582798993358246,0.7819526342245247,0.7138588428497314,0.6585644131669631,0.7675375267863273,0.7065790295600891,0.6665758143250758,0.7436406246744669,,,
Lung Opacity,0.7780014872550964,0.7502247927280573,0.805139823945669,0.7330787777900696,0.7162032002439865,0.7501237911673693,0.696986198425293,0.6747336007081546,0.7192720676843937,*,*,
Pleural Effusion,0.9507032036781312,0.9357649125731908,0.9639839173509523,0.9139318466186525,0.9041223911138682,0.923487456716024,0.8695277571678162,0.8551680750571764,0.8830073813979442,*,*,*
Pleural Other,0.738192081451416,0.6172874871354836,0.8362653759809638,0.8525621891021729,0.7956556346554022,0.9028798949259976,0.782571017742157,0.7151304037525104,0.8430834293365478,,,
Pneumonia,0.7212163805961609,0.6896892465077914,0.7531831287420713,0.6828678846359253,0.6459232486211337,0.7197920026687474,0.7028828859329224,0.6621934400154994,0.7411170768050045,,,
Pneumothorax,0.8275972604751587,0.7636218440074187,0.8857287644193722,0.879091203212738,0.8513032659888268,0.904641455411911,0.8515167236328125,0.8201992290524336,0.8802382834828816,,,
Support Devices,0.935628056526184,0.9155613030378636,0.9533017924198738,0.900791883468628,0.8897037110649623,0.9116406397177624,0.8544577360153198,0.8383424080335177,0.8702489018440246,*,*,*
\end{filecontents*}
\csvreader[
  head=true, 
  late after line=\\
]{csv/quantile_bow/bow_quantile_AUROC.csv}{}{%
  \csvcoli\textsuperscript{\csvcolxii} & 
  \num[round-mode=places,round-precision=3]{\csvcolii} \scriptsize(\num[round-mode=places,round-precision=3]{\csvcoliii}--\num[round-mode=places,round-precision=3]{\csvcoliv}) &
  \num[round-mode=places,round-precision=3]{\csvcolv} \scriptsize(\num[round-mode=places,round-precision=3]{\csvcolvi}--\num[round-mode=places,round-precision=3]{\csvcolvii}) &
  \num[round-mode=places,round-precision=3]{\csvcolviii} \scriptsize(\num[round-mode=places,round-precision=3]{\csvcolix}--\num[round-mode=places,round-precision=3]{\csvcolx})%
}
\bottomrule
\end{tabular}
\end{table}

\begin{table*}[htbp]
\centering
\caption{Held-out performance (in terms of \textbf{AUROC}) of CXR ResNet50 models across sub-populations stratified by pre-test probability (LM embeddings) of the CXR label. Labels marked with an asterisk (*) indicate a statistically significant difference between the Bottom 25\% and Top 25\% groups.}
\label{tab:resnet50_quantile}
\footnotesize 
\setlength{\tabcolsep}{12pt} 
\begin{tabular}{@{}l ccc@{}}
\toprule
\textbf{Label} & \textbf{Bottom 25\%} & \textbf{Middle 50\%} & \textbf{Top 25\%} \\
\midrule
Atelectasis* & \num[round-mode=places,round-precision=3]{0.775509} \scriptsize(\num[round-mode=places,round-precision=3]{0.745145}--\num[round-mode=places,round-precision=3]{0.803472}) & \num[round-mode=places,round-precision=3]{0.732707} \scriptsize(\num[round-mode=places,round-precision=3]{0.714670}--\num[round-mode=places,round-precision=3]{0.750584}) & \num[round-mode=places,round-precision=3]{0.651757} \scriptsize(\num[round-mode=places,round-precision=3]{0.627925}--\num[round-mode=places,round-precision=3]{0.675471}) \\
Cardiomegaly* & \num[round-mode=places,round-precision=3]{0.757433} \scriptsize(\num[round-mode=places,round-precision=3]{0.721824}--\num[round-mode=places,round-precision=3]{0.792871}) & \num[round-mode=places,round-precision=3]{0.701504} \scriptsize(\num[round-mode=places,round-precision=3]{0.681455}--\num[round-mode=places,round-precision=3]{0.720719}) & \num[round-mode=places,round-precision=3]{0.628169} \scriptsize(\num[round-mode=places,round-precision=3]{0.604090}--\num[round-mode=places,round-precision=3]{0.650915}) \\
Consolidation & \num[round-mode=places,round-precision=3]{0.772317} \scriptsize(\num[round-mode=places,round-precision=3]{0.698351}--\num[round-mode=places,round-precision=3]{0.837585}) & \num[round-mode=places,round-precision=3]{0.696846} \scriptsize(\num[round-mode=places,round-precision=3]{0.659497}--\num[round-mode=places,round-precision=3]{0.734798}) & \num[round-mode=places,round-precision=3]{0.684637} \scriptsize(\num[round-mode=places,round-precision=3]{0.639823}--\num[round-mode=places,round-precision=3]{0.727427}) \\
Edema* & \num[round-mode=places,round-precision=3]{0.860103} \scriptsize(\num[round-mode=places,round-precision=3]{0.809766}--\num[round-mode=places,round-precision=3]{0.906856}) & \num[round-mode=places,round-precision=3]{0.823006} \scriptsize(\num[round-mode=places,round-precision=3]{0.800528}--\num[round-mode=places,round-precision=3]{0.844322}) & \num[round-mode=places,round-precision=3]{0.756136} \scriptsize(\num[round-mode=places,round-precision=3]{0.733786}--\num[round-mode=places,round-precision=3]{0.778102}) \\
Fracture & \num[round-mode=places,round-precision=3]{0.669201} \scriptsize(\num[round-mode=places,round-precision=3]{0.558687}--\num[round-mode=places,round-precision=3]{0.766499}) & \num[round-mode=places,round-precision=3]{0.619504} \scriptsize(\num[round-mode=places,round-precision=3]{0.550367}--\num[round-mode=places,round-precision=3]{0.690039}) & \num[round-mode=places,round-precision=3]{0.640239} \scriptsize(\num[round-mode=places,round-precision=3]{0.577702}--\num[round-mode=places,round-precision=3]{0.701701}) \\
Lung Lesion & \num[round-mode=places,round-precision=3]{0.603903} \scriptsize(\num[round-mode=places,round-precision=3]{0.468978}--\num[round-mode=places,round-precision=3]{0.734301}) & \num[round-mode=places,round-precision=3]{0.589861} \scriptsize(\num[round-mode=places,round-precision=3]{0.519538}--\num[round-mode=places,round-precision=3]{0.654864}) & \num[round-mode=places,round-precision=3]{0.566134} \scriptsize(\num[round-mode=places,round-precision=3]{0.523984}--\num[round-mode=places,round-precision=3]{0.606874}) \\
Lung Opacity* & \num[round-mode=places,round-precision=3]{0.710305} \scriptsize(\num[round-mode=places,round-precision=3]{0.679787}--\num[round-mode=places,round-precision=3]{0.740469}) & \num[round-mode=places,round-precision=3]{0.643738} \scriptsize(\num[round-mode=places,round-precision=3]{0.624415}--\num[round-mode=places,round-precision=3]{0.662612}) & \num[round-mode=places,round-precision=3]{0.608246} \scriptsize(\num[round-mode=places,round-precision=3]{0.584428}--\num[round-mode=places,round-precision=3]{0.632315}) \\
Pleural Effusion* & \num[round-mode=places,round-precision=3]{0.860249} \scriptsize(\num[round-mode=places,round-precision=3]{0.830496}--\num[round-mode=places,round-precision=3]{0.886894}) & \num[round-mode=places,round-precision=3]{0.813719} \scriptsize(\num[round-mode=places,round-precision=3]{0.798847}--\num[round-mode=places,round-precision=3]{0.828260}) & \num[round-mode=places,round-precision=3]{0.747308} \scriptsize(\num[round-mode=places,round-precision=3]{0.728388}--\num[round-mode=places,round-precision=3]{0.766423}) \\
Pleural Other & \num[round-mode=places,round-precision=3]{0.749264} \scriptsize(\num[round-mode=places,round-precision=3]{0.629206}--\num[round-mode=places,round-precision=3]{0.850901}) & \num[round-mode=places,round-precision=3]{0.700964} \scriptsize(\num[round-mode=places,round-precision=3]{0.615819}--\num[round-mode=places,round-precision=3]{0.781896}) & \num[round-mode=places,round-precision=3]{0.684555} \scriptsize(\num[round-mode=places,round-precision=3]{0.606101}--\num[round-mode=places,round-precision=3]{0.759720}) \\
Pneumonia & \num[round-mode=places,round-precision=3]{0.607185} \scriptsize(\num[round-mode=places,round-precision=3]{0.553667}--\num[round-mode=places,round-precision=3]{0.661923}) & \num[round-mode=places,round-precision=3]{0.612005} \scriptsize(\num[round-mode=places,round-precision=3]{0.579007}--\num[round-mode=places,round-precision=3]{0.643976}) & \num[round-mode=places,round-precision=3]{0.601690} \scriptsize(\num[round-mode=places,round-precision=3]{0.562062}--\num[round-mode=places,round-precision=3]{0.639481}) \\
Pneumothorax & \num[round-mode=places,round-precision=3]{0.764921} \scriptsize(\num[round-mode=places,round-precision=3]{0.695748}--\num[round-mode=places,round-precision=3]{0.826214}) & \num[round-mode=places,round-precision=3]{0.778718} \scriptsize(\num[round-mode=places,round-precision=3]{0.740443}--\num[round-mode=places,round-precision=3]{0.816542}) & \num[round-mode=places,round-precision=3]{0.735095} \scriptsize(\num[round-mode=places,round-precision=3]{0.697694}--\num[round-mode=places,round-precision=3]{0.769323}) \\
Support Devices* & \num[round-mode=places,round-precision=3]{0.879369} \scriptsize(\num[round-mode=places,round-precision=3]{0.853154}--\num[round-mode=places,round-precision=3]{0.902446}) & \num[round-mode=places,round-precision=3]{0.841180} \scriptsize(\num[round-mode=places,round-precision=3]{0.826695}--\num[round-mode=places,round-precision=3]{0.855135}) & \num[round-mode=places,round-precision=3]{0.784572} \scriptsize(\num[round-mode=places,round-precision=3]{0.765034}--\num[round-mode=places,round-precision=3]{0.802918}) \\
\bottomrule
\end{tabular}
\end{table*}

\begin{table}[htbp]
\centering
\caption{Held-out performance (in terms of \textbf{sensitivity at local 95\% specificity}) of CXR models across sub-populations stratified by pre-test probability  (BoW Representations) of the CXR label. Labels marked with an asterisk (*) indicate a statistically significant difference between the Bottom 25\% and Top 25\% groups.}
\label{tab:quantile_bow_sens_local}
\footnotesize 
\setlength{\tabcolsep}{12pt} 
\begin{tabular}{@{}l ccc@{}}
\toprule
\textbf{Label} & \textbf{Bottom 25\%} & \textbf{Middle 50\%} & \textbf{Top 25\%} \\
\midrule
\begin{filecontents*}{csv/quantile_bow/bow_quantile_Sens_at_Spec95.csv}
Label,bottom_25_Mean,bottom_25_CI_Lower,bottom_25_CI_Upper,middle_50_Mean,middle_50_CI_Lower,middle_50_CI_Upper,top_25_Mean,top_25_CI_Lower,top_25_CI_Upper,sig_bottom_25_vs_middle_50,sig_bottom_25_vs_top_25,sig_middle_50_vs_top_25
Atelectasis,0.3922992523364486,0.319626168224299,0.4654205607476635,0.2872749456521739,0.2467391304347826,0.3254191053511704,0.1650994971264368,0.1222908377541998,0.2040229885057471,,*,*
Cardiomegaly,0.2925880681818181,0.21875,0.3693181818181818,0.2579951980792317,0.2120221165389232,0.3023836457659985,0.1997971771771771,0.1585585585585585,0.2401027951027949,,,
Consolidation,0.3721957264957265,0.2393162393162393,0.5194115713346457,0.3643345454545454,0.2818181818181818,0.4515151515151515,0.183259169550173,0.1176470588235294,0.259515570934256,,,*
Edema,0.6145507352941176,0.4796238687782805,0.7483173076923054,0.4565591911764706,0.3936628016591252,0.5171568627450981,0.3230170989433237,0.2718539865513928,0.377521613832853,,*,*
Enlarged Cardiomediastinum,0.1954680851063829,0.0851063829787234,0.3297872340425531,0.1748592274678111,0.1030042918454935,0.2532188841201717,0.118393820224719,0.0561797752808988,0.196629213483146,,,
Fracture,0.1909220430107526,0.1129032258064516,0.2849462365591398,,,,0.2007486033519553,0.1424581005586592,0.2625698324022346,,,
Lung Lesion,0.0771878048780487,0.0,0.2439024390243902,0.1900346368715083,0.106145251396648,0.2793296089385474,0.1721569974554707,0.1119592875318066,0.2340966921119593,,,
Lung Opacity,0.283305598755832,0.2177293934681182,0.3452566096423017,0.1913001984126984,0.1613238324175824,0.2217261904761904,0.1533928117048346,0.116412213740458,0.1921119592875318,,*,
Pleural Effusion,0.7095935028248589,0.6242937853107344,0.7966101694915254,0.6041501197891711,0.5588063469831558,0.6514337842320589,0.4410457410562181,0.3753104114794915,0.5186418883501506,,*,*
Pleural Other,0.2146431818181818,0.0454545454545454,0.4090909090909091,0.3847707317073171,0.2317073170731707,0.5487804878048781,0.2350837209302325,0.1162790697674418,0.3720930232558139,,,
Pneumonia,0.2192937943262411,0.1648936170212765,0.2783687943262411,0.1203132743362832,0.0685840707964601,0.1747787610619469,0.2416376299376299,0.1808731808731808,0.3035343035343035,,,*
Pneumothorax,0.3418333333333333,0.1884057971014492,0.5217391304347826,0.4959567857142857,0.4071428571428571,0.5857142857142857,0.453382294264339,0.3740648379052369,0.5361596009975063,,,
Support Devices,0.7825334710743802,0.7252066115702479,0.8383701525746974,0.5763530817009077,0.52651696129957,0.6230291447682752,0.4174789054197662,0.3534690999754761,0.4796871168151719,*,*,*
\end{filecontents*}
\csvreader[
  head=true, 
  late after line=\\
]{csv/quantile_bow/bow_quantile_Sens_at_Spec95.csv}{}{%
  \csvcoli\textsuperscript{\csvcolxii} & 
  \num[round-mode=places,round-precision=3]{\csvcolii} \scriptsize(\num[round-mode=places,round-precision=3]{\csvcoliii}--\num[round-mode=places,round-precision=3]{\csvcoliv}) &
  \num[round-mode=places,round-precision=3]{\csvcolv} \scriptsize(\num[round-mode=places,round-precision=3]{\csvcolvi}--\num[round-mode=places,round-precision=3]{\csvcolvii}) &
  \num[round-mode=places,round-precision=3]{\csvcolviii} \scriptsize(\num[round-mode=places,round-precision=3]{\csvcolix}--\num[round-mode=places,round-precision=3]{\csvcolx})%
}
\bottomrule
\end{tabular}
\end{table}

\begin{table}[htbp]
\centering
\caption{Held-out performance (in terms of \textbf{sensitivity at global 95\% specificity}) of CXR models across sub-populations stratified by pre-test probability (BoW Representations)  of the CXR label. Labels marked with an asterisk (*) indicate a statistically significant difference between the Bottom 25\% and Top 25\% groups.}
\label{tab:quantile_bow_sens_global}
\footnotesize 
\setlength{\tabcolsep}{12pt} 
\begin{tabular}{@{}l ccc@{}}
\toprule
\textbf{Label} & \textbf{Bottom 25\%} & \textbf{Middle 50\%} & \textbf{Top 25\%} \\
\midrule
\begin{filecontents*}{csv/quantile_bow/bow_quantile_Sens_at_Global_Spec95.csv}
Label,bottom_25_Mean,bottom_25_CI_Lower,bottom_25_CI_Upper,middle_50_Mean,middle_50_CI_Lower,middle_50_CI_Upper,top_25_Mean,top_25_CI_Lower,top_25_CI_Upper,sig_bottom_25_vs_middle_50,sig_bottom_25_vs_top_25,sig_middle_50_vs_top_25
Atelectasis,0.2369588785046728,0.1813084112149532,0.2948993529834645,0.2551934782608696,0.2195652173913043,0.2902173913043478,0.2866760775862069,0.2435344827586207,0.3290229885057471,,,
Cardiomegaly,0.0697204545454545,0.034090909090909,0.1164772727272727,0.1980269507803121,0.1668667466986794,0.2286914765906362,0.4012001801801801,0.3616990066990067,0.445045045045045,*,*,*
Consolidation,0.2433239316239316,0.1282051282051282,0.358974358974359,0.3369136363636363,0.2606060606060606,0.417488344988344,0.2775581314878892,0.2006920415224913,0.3598615916955017,,,
Edema,0.2673948529411765,0.1617647058823529,0.3823529411764705,0.4003664215686274,0.3455882352941176,0.4558823529411764,0.5294459173871277,0.4786060001477869,0.5761490430798785,,*,*
Enlarged Cardiomediastinum,0.1565893617021276,0.0556260229132569,0.2659574468085106,0.1690884120171674,0.1030042918454935,0.2446351931330472,0.1725342696629213,0.0898876404494382,0.2584269662921348,,,
Fracture,0.1805483870967742,0.1021505376344086,0.2688172043010752,,,,0.2044715083798882,0.1452513966480447,0.2681564245810056,,,
Lung Lesion,0.048290243902439,0.0,0.1707317073170731,0.1586826815642458,0.0837988826815642,0.2458100558659217,0.3489997455470738,0.2798982188295165,0.4223918575063613,,*,*
Lung Opacity,0.1345760497667185,0.0948678071539657,0.1753828209115918,0.1817619047619047,0.1553714514652014,0.2103174603174603,0.2654606234096692,0.227735368956743,0.307106331963202,,*,*
Pleural Effusion,0.4534491525423729,0.3757062146892655,0.5310734463276836,0.5471469094393867,0.5089740518226383,0.5867476318602334,0.6850847529812606,0.6465076660988075,0.7217935067487877,,*,*
Pleural Other,0.118215909090909,0.0,0.2727272727272727,0.3741756097560976,0.2195121951219512,0.524390243902439,0.3300093023255814,0.1860465116279069,0.4767441860465116,,,
Pneumonia,0.1916241134751773,0.1400709219858156,0.2442750954719034,0.0972807522123894,0.0575221238938053,0.1415929203539823,0.2948744282744283,0.237006237006237,0.3555093555093555,,,*
Pneumothorax,0.1941376811594203,0.0579710144927536,0.3478260869565217,0.4313471428571428,0.3464285714285714,0.5214285714285715,0.6132633416458853,0.5411471321695761,0.6907730673316709,,*,*
Support Devices,0.5362518595041322,0.4669421487603306,0.6033057851239669,0.5558744386048734,0.517439082656474,0.5953177257525084,0.622740701381509,0.5818278427205101,0.6663124335812965,,,
\end{filecontents*}
\csvreader[
  head=true, 
  late after line=\\
]{csv/quantile_bow/bow_quantile_Sens_at_Global_Spec95.csv}{}{%
  \csvcoli\textsuperscript{\csvcolxii} & 
  \num[round-mode=places,round-precision=3]{\csvcolii} \scriptsize(\num[round-mode=places,round-precision=3]{\csvcoliii}--\num[round-mode=places,round-precision=3]{\csvcoliv}) &
  \num[round-mode=places,round-precision=3]{\csvcolv} \scriptsize(\num[round-mode=places,round-precision=3]{\csvcolvi}--\num[round-mode=places,round-precision=3]{\csvcolvii}) &
  \num[round-mode=places,round-precision=3]{\csvcolviii} \scriptsize(\num[round-mode=places,round-precision=3]{\csvcolix}--\num[round-mode=places,round-precision=3]{\csvcolx})%
}
\bottomrule
\end{tabular}
\end{table}

\begin{table}[htbp]
\centering
\caption{Held-out performance (in terms of \textbf{specificity at global 95\% specificity}) of CXR models across sub-populations stratified by pre-test probability (BoW Representations)  of the CXR label. Labels marked with an asterisk (*) indicate a statistically significant difference between the Bottom 25\% and Top 25\% groups.}
\label{tab:quantile_bow_spec_global}
\footnotesize 
\setlength{\tabcolsep}{12pt} 
\begin{tabular}{@{}l ccc@{}}
\toprule
\textbf{Label} & \textbf{Bottom 25\%} & \textbf{Middle 50\%} & \textbf{Top 25\%} \\
\midrule
\begin{filecontents*}{csv/quantile_bow/bow_quantile_Spec_at_Global_Spec95.csv}
Label,bottom_25_Mean,bottom_25_CI_Lower,bottom_25_CI_Upper,middle_50_Mean,middle_50_CI_Lower,middle_50_CI_Upper,top_25_Mean,top_25_CI_Lower,top_25_CI_Upper,sig_bottom_25_vs_middle_50,sig_bottom_25_vs_top_25,sig_middle_50_vs_top_25
Atelectasis,0.9786884129363892,0.9725851360034268,0.9841507817519812,0.957715058685446,0.9528169014084508,0.9625586854460094,0.8976112218143683,0.8869952805453593,0.9084950183534348,*,*,*
Cardiomegaly,0.993044972124716,0.9894693371876936,0.9960768118934544,0.963696317117944,0.9592703074804956,0.9679899036255164,0.8570918083900227,0.8460884353741497,0.8681972789115646,*,*,*
Consolidation,0.9714456855791964,0.9645390070921984,0.9775413711583923,0.9569012878107218,0.9526804432464808,0.9614426628577571,0.9139834648856509,0.9050799433313096,0.9230924913985024,*,*,*
Edema,0.9869950563575244,0.9820051413881749,0.991101443543603,0.9594551518324608,0.9549738219895288,0.9638743455497382,0.8835190647482015,0.8733812949640288,0.8932853717026379,*,*,*
Enlarged Cardiomediastinum,0.9702844283192782,0.9639144930378506,0.9764659737203372,0.9517816251974724,0.9472748815165876,0.9561611374407584,0.9258779475982531,0.9168775002290286,0.9342539011207132,*,*,*
Fracture,0.953773061760841,0.9500174486048336,0.957449471247106,,,,0.9470502915666194,0.944197032692402,0.9500171509776058,,*,
Lung Lesion,0.9856324544397054,0.980612640558356,0.9901124466847616,0.9641693451501724,0.9598227474150663,0.9682914820285574,0.8823845375722543,0.8732452518579686,0.8912056151940545,*,*,*
Lung Opacity,0.9793875383267632,0.973718791064389,0.9848883048620236,0.9538547992810064,0.948831635710006,0.958777711204314,0.9042128169790518,0.8936052921719956,0.915380374862183,*,*,*
Pleural Effusion,0.9818745825602968,0.9764996907854052,0.9868068439497012,0.9609432037662966,0.9560598744567842,0.965810134457527,0.8637363668306394,0.8496134926212228,0.8777231201686577,*,*,*
Pleural Other,0.9731211844660193,0.9668405526512324,0.9790291262135924,0.9550838082137943,0.9505796816663392,0.9597170367459226,0.9173822562141491,0.908413001912046,0.925812619502868,*,*,*
Pneumonia,0.9582569447736432,0.9530694477364304,0.9633528269549838,0.955681125380196,0.9491382223724232,0.96214937478878,0.9289832418141144,0.9204799674598332,0.9375635550132194,,*,*
Pneumothorax,0.9815626123541786,0.9762860967680244,0.9864218779881432,0.964231176235247,0.9598148475849017,0.9685708295505256,0.8858116628416441,0.876486542875026,0.895263926559566,*,*,*
Support Devices,0.9847232587324828,0.979711357456599,0.9891236143066304,0.953753742241694,0.949036004156486,0.9585945179318672,0.8905285541805011,0.8786598249320857,0.9022034409900392,*,*,*
\end{filecontents*}
\csvreader[
  head=true, 
  late after line=\\
]{csv/quantile_bow/bow_quantile_Spec_at_Global_Spec95.csv}{}{%
  \csvcoli\textsuperscript{\csvcolxii} & 
  \num[round-mode=places,round-precision=3]{\csvcolii} \scriptsize(\num[round-mode=places,round-precision=3]{\csvcoliii}--\num[round-mode=places,round-precision=3]{\csvcoliv}) &
  \num[round-mode=places,round-precision=3]{\csvcolv} \scriptsize(\num[round-mode=places,round-precision=3]{\csvcolvi}--\num[round-mode=places,round-precision=3]{\csvcolvii}) &
  \num[round-mode=places,round-precision=3]{\csvcolviii} \scriptsize(\num[round-mode=places,round-precision=3]{\csvcolix}--\num[round-mode=places,round-precision=3]{\csvcolx})%
}
\bottomrule
\end{tabular}
\end{table}

\begin{table}[htbp]
\centering
\caption{Held-out performance (in terms of \textbf{Adaptive Calibration Error (ACE)}) of CXR models across sub-populations stratified by pre-test probability (BoW representations) of the CXR label.}
\label{tab:quantile_bow_ace}
\footnotesize
\setlength{\tabcolsep}{10pt}
\begin{tabular}{@{}l ccc@{}}
\toprule
\textbf{Label} & \textbf{Bottom 25\%} & \textbf{Middle 50\%} & \textbf{Top 25\%} \\
\midrule
Atelectasis & 0.011 (0.005--0.019) & 0.018 (0.014--0.023) & 0.036 (0.024--0.051) \\
Cardiomegaly & 0.013 (0.007--0.021) & 0.012 (0.006--0.021) & 0.030 (0.019--0.044) \\
Consolidation & 0.005 (0.002--0.009) & 0.006 (0.002--0.010) & 0.013 (0.008--0.021) \\
Edema & 0.014 (0.010--0.018) & 0.017 (0.013--0.021) & 0.019 (0.009--0.031) \\
Enlarged Cardiomediastinum & 0.005 (0.003--0.009) & 0.006 (0.003--0.009) & 0.009 (0.004--0.016) \\
Fracture & 0.006 (0.005--0.008) & $-$ & 0.003 (0.001--0.006) \\
Lung Lesion & 0.016 (0.015--0.018) & 0.013 (0.012--0.015) & 0.022 (0.019--0.029) \\
Lung Opacity & 0.014 (0.006--0.023) & 0.017 (0.009--0.026) & 0.055 (0.048--0.065) \\
Pleural Effusion & 0.013 (0.006--0.022) & 0.011 (0.006--0.019) & 0.037 (0.027--0.048) \\
Pleural Other & 0.004 (0.002--0.008) & 0.004 (0.002--0.006) & 0.007 (0.002--0.011) \\
Pneumonia & 0.008 (0.003--0.014) & 0.013 (0.006--0.021) & 0.015 (0.008--0.024) \\
Pneumothorax & 0.005 (0.002--0.009) & 0.006 (0.005--0.008) & 0.020 (0.016--0.025) \\
Support Devices & 0.011 (0.006--0.017) & 0.015 (0.008--0.022) & 0.035 (0.022--0.049) \\
\bottomrule
\end{tabular}
\end{table}

\begin{table}[htbp]
\centering
\caption{Held-out performance (in terms of \textbf{Brier Skill Score (BSS)}) of CXR models across sub-populations stratified by pre-test probability (BoW representations) of the CXR label.}
\label{tab:quantile_bow_bss}
\footnotesize
\setlength{\tabcolsep}{10pt}
\begin{tabular}{@{}l ccc@{}}
\toprule
\textbf{Label} & \textbf{Bottom 25\%} & \textbf{Middle 50\%} & \textbf{Top 25\%} \\
\midrule
Atelectasis & 0.209 (0.166--0.250) & 0.178 (0.155--0.201) & 0.113 (0.084--0.140) \\
Cardiomegaly & 0.110 (0.067--0.154) & 0.150 (0.126--0.174) & 0.153 (0.125--0.183) \\
Consolidation & 0.061 (0.023--0.101) & 0.072 (0.050--0.095) & 0.045 (0.023--0.069) \\
Edema & 0.126 (0.036--0.216) & 0.220 (0.182--0.260) & 0.234 (0.194--0.272) \\
Enlarged Cardiomediastinum & 0.019 ($-$0.002--0.042) & 0.013 ($-$0.002--0.028) & 0.005 ($-$0.013--0.025) \\
Fracture & 0.004 ($-$0.004--0.013) & $-$ & 0.012 (0.006--0.019) \\
Lung Lesion & $-$0.091 ($-$0.124--$-$0.059) & $-$0.014 ($-$0.042--0.015) & 0.027 (0.002--0.051) \\
Lung Opacity & 0.130 (0.097--0.164) & 0.111 (0.090--0.131) & 0.086 (0.059--0.111) \\
Pleural Effusion & 0.387 (0.312--0.458) & 0.457 (0.426--0.487) & 0.406 (0.370--0.441) \\
Pleural Other & $-$0.006 ($-$0.031--0.029) & 0.039 (0.003--0.086) & 0.019 ($-$0.011--0.053) \\
Pneumonia & 0.045 (0.021--0.069) & 0.020 ($-$0.001--0.041) & 0.072 (0.041--0.103) \\
Pneumothorax & 0.010 ($-$0.055--0.081) & 0.133 (0.078--0.188) & 0.185 (0.128--0.243) \\
Support Devices & 0.502 (0.445--0.560) & 0.439 (0.407--0.471) & 0.372 (0.331--0.409) \\
\bottomrule
\end{tabular}
\end{table}

\begin{table}[htbp]
\centering
\caption{Held-out performance (in terms of \textbf{AUROC}) of CXR models across sub-populations stratified by the presence of label-relevant terms in prior discharge summaries. Statistically significant differences (as evaluated by bootstrapping) are highlighted with an asterisk (*) next to the label.}
\label{tab:prior_mention_auroc}
\footnotesize 
\setlength{\tabcolsep}{18pt} 
\begin{tabular}{@{}l cc@{}}
\toprule
\textbf{Label} & \textbf{Previous Mentions} & \textbf{No Previous Mentions} \\
\midrule
\begin{filecontents*}{csv/prior_mention/prior_mention_AUROC.csv}
Label,Previous Mentions_Mean,Previous Mentions_CI_Lower,Previous Mentions_CI_Upper,No Previous Mentions_Mean,No Previous Mentions_CI_Lower,No Previous Mentions_CI_Upper,sig
Atelectasis,0.7572678237375079,0.7402841203985341,0.7732582996488686,0.8308511107932031,0.8156268803195513,0.8455553313274163,*
Cardiomegaly,0.7869851009002694,0.7717630432694701,0.8022014135065058,0.8192523613014655,0.8040045263380861,0.8338676349456722,*
Consolidation,0.8019171635112361,0.7696295415124912,0.8320224434128203,0.828322043766711,0.8014195535759076,0.8541425539140831,
Edema,0.8887239240828804,0.8785183191523718,0.8982275792239754,0.8999748700373779,0.8380617212444482,0.9450270505497266,
Enlarged Cardiomediastinum,0.6989536647132596,0.6208883732568575,0.7696902011264579,0.7104737861541559,0.6740644039600668,0.7454032076196011,
Fracture,0.6507639028086527,0.5949518562223879,0.7049107939158595,0.7020454599528665,0.6296855700359276,0.7707350782176946,
Lung Lesion,0.7367793741751377,0.702685728371114,0.7694377080167715,0.7501973299958876,0.6906514058913598,0.8031154647865125,
Lung Opacity,0.724389697135967,0.70874935203676,0.7396834385654886,0.7678919333266699,0.7501163401048334,0.7846521679262546,*
Pleural Effusion,0.9180990910194552,0.9109865216599288,0.924981888783234,0.929517305247028,0.917906574296094,0.9404038549886384,
Pleural Other,0.704338910331753,0.5820454167127311,0.8133317109988923,0.8175215370685666,0.7738874949125261,0.8595174515239747,
Pneumonia,0.696427300286611,0.6698174337166736,0.7225157761760381,0.7139022494202654,0.6822426990436733,0.7444673127551429,
Pneumothorax,0.8666046578939015,0.8401253083660423,0.8907030691479605,0.8918299290030601,0.8658556250995438,0.9150948211758612,
Support Devices,0.9010395670711876,0.8933675557682621,0.9087930165566126,0.932262863878804,0.9005995240971995,0.9593173736880768,
\end{filecontents*}
\csvreader[
  head=true, 
  late after line=\\
]{csv/prior_mention/prior_mention_AUROC.csv}{}{%
  \csvcoli\textsuperscript{\csvcolviii} & 
  \num[round-mode=places,round-precision=3]{\csvcolii} \scriptsize(\num[round-mode=places,round-precision=3]{\csvcoliii}--\num[round-mode=places,round-precision=3]{\csvcoliv}) &
  \num[round-mode=places,round-precision=3]{\csvcolv} \scriptsize(\num[round-mode=places,round-precision=3]{\csvcolvi}--\num[round-mode=places,round-precision=3]{\csvcolvii})
}
\bottomrule
\end{tabular}
\end{table}

\begin{table}[htbp]
\centering
\caption{Held-out performance (in terms of \textbf{sensitivity at local 95\% specificity}) of CXR models across sub-populations stratified by the presence of label-relevant terms in prior discharge summaries. Statistically significant differences (as evaluated by bootstrapping) are highlighted with an asterisk (*) next to the label.}
\label{tab:prior_mention_sens_local}
\footnotesize 
\setlength{\tabcolsep}{18pt} 
\begin{tabular}{@{}l cc@{}}
\toprule
\textbf{Label} & \textbf{Previous Mentions} & \textbf{No Previous Mentions} \\
\midrule
\begin{filecontents*}{csv/prior_mention/prior_mention_Sens_at_Spec95.csv}
Label,Previous Mentions_Mean,Previous Mentions_CI_Lower,Previous Mentions_CI_Upper,No Previous Mentions_Mean,No Previous Mentions_CI_Lower,No Previous Mentions_CI_Upper,sig
Atelectasis,0.2006040391171437,0.1643941490282439,0.2393055121180815,0.3330808558646243,0.2892765466697196,0.3794176291860094,*
Cardiomegaly,0.2639914995656537,0.225010020063173,0.303129868912801,0.2983353294035332,0.2535407183777945,0.3447712567367291,
Consolidation,0.2748272272478101,0.2055471787781906,0.3487511836635275,0.3154405067199497,0.2487765637266544,0.390536108699805,
Edema,0.4520949773376073,0.4133279808860235,0.4912440727570602,0.4758096413578088,0.2549056678088936,0.6965468474614811,
Enlarged Cardiomediastinum,0.164865423222172,0.0634009954314155,0.2853937728937724,0.1697777983880378,0.1155553928638538,0.2295927783971261,
Fracture,0.1843843225112671,0.1134161494805524,0.2643421082123004,0.2309272698675075,0.1300813008130081,0.3427944302874114,
Lung Lesion,0.2527849980276213,0.1930336922560005,0.3136050290017915,0.2055503834062314,0.1140485603528697,0.312292711817168,
Lung Opacity,0.1773129476399031,0.1494801776100333,0.2061790199909106,0.2376022027396013,0.2,0.2745563099587891,
Pleural Effusion,0.5879961444799797,0.5522029261560507,0.6240971902035374,0.6229364026591129,0.5545235859035365,0.6781936912774814,
Pleural Other,0.1146037495763059,0.0,0.2748878205128204,0.3470520331106833,0.2488592398725534,0.4588247037496863,
Pneumonia,0.1950564007933844,0.1525317603897012,0.2359404473493451,0.1951468684339166,0.1474535110511163,0.2469959200835536,
Pneumothorax,0.4903228220229652,0.4189332361207361,0.5670588353552063,0.5273591094883688,0.4366247271190817,0.6156723641091534,
Support Devices,0.5622560881031599,0.5273599607204219,0.5945295299183286,0.7927679718470206,0.7020223944151902,0.8723513291665745,*
\end{filecontents*}
\csvreader[
  head=true, 
  late after line=\\
]{csv/prior_mention/prior_mention_Sens_at_Spec95.csv}{}{%
  \csvcoli\textsuperscript{\csvcolviii} & 
  \num[round-mode=places,round-precision=3]{\csvcolii} \scriptsize(\num[round-mode=places,round-precision=3]{\csvcoliii}--\num[round-mode=places,round-precision=3]{\csvcoliv}) &
  \num[round-mode=places,round-precision=3]{\csvcolv} \scriptsize(\num[round-mode=places,round-precision=3]{\csvcolvi}--\num[round-mode=places,round-precision=3]{\csvcolvii})
}
\bottomrule
\end{tabular}
\end{table}

\begin{table}[htbp]
\centering
\caption{Held-out performance (in terms of \textbf{sensitivity at global 95\% specificity}) of CXR models across sub-populations stratified by the presence of label-relevant terms in prior discharge summaries. Statistically significant differences (as evaluated by bootstrapping) are highlighted with an asterisk (*) next to the label.}
\label{tab:prior_mention_sens_global}
\footnotesize 
\setlength{\tabcolsep}{18pt} 
\begin{tabular}{@{}l cc@{}}
\toprule
\textbf{Label} & \textbf{Previous Mentions} & \textbf{No Previous Mentions} \\
\midrule
\begin{filecontents*}{csv/prior_mention/prior_mention_Sens_at_Global_Spec95.csv}
Label,Previous Mentions_Mean,Previous Mentions_CI_Lower,Previous Mentions_CI_Upper,No Previous Mentions_Mean,No Previous Mentions_CI_Lower,No Previous Mentions_CI_Upper,sig
Atelectasis,0.26401218703183,0.2303240385500752,0.2979485487391962,0.2641839121045947,0.2249653242441012,0.3037421014951207,
Cardiomegaly,0.3363964306810267,0.3009266825632853,0.371556417941895,0.1981017234805211,0.1648289441979277,0.2327531895178771,*
Consolidation,0.296580473338203,0.2254349945828819,0.3749114372469634,0.3000584318149412,0.2359264113748846,0.3665911398832189,
Edema,0.4634570391684625,0.4251667890540442,0.502342971300951,0.315440367347148,0.1540940911742645,0.4891352338903053,
Enlarged Cardiomediastinum,0.1680158085537616,0.0680116606694276,0.284313725490196,0.1680155448757783,0.115283454413849,0.2252166620854161,
Fracture,0.1944048480223531,0.1223501880138999,0.2748349681358244,0.2125637057115299,0.1177530551893683,0.3271526058828777,
Lung Lesion,0.3124569861501513,0.250533961202751,0.3765175583403413,0.1480435663562299,0.071535409035409,0.2438996840571641,*
Lung Opacity,0.2262785232751469,0.1968223061861184,0.2594037760683579,0.1705345914121116,0.1421238518236761,0.1999421375085091,
Pleural Effusion,0.6286077574341254,0.5955591108993946,0.6611000200380484,0.524573405929815,0.4730964566215426,0.5758771396872358,*
Pleural Other,0.1510096658539048,0.0218496841322928,0.3260869565217391,0.3440151442239181,0.2460796128217209,0.4534718441316014,
Pneumonia,0.2092731925977248,0.1669362459082335,0.2515739243827381,0.176742102576435,0.1315395211966708,0.2257056915090277,
Pneumothorax,0.4966558750735836,0.4222930599567313,0.5727822718270668,0.5190641656096876,0.4350453172205438,0.602376561217527,
Support Devices,0.5829140479056707,0.549735915301462,0.6177322151837127,0.5703947789750929,0.4721009085402786,0.6636407318813208,
\end{filecontents*}
\csvreader[
  head=true, 
  late after line=\\
]{csv/prior_mention/prior_mention_Sens_at_Global_Spec95.csv}{}{%
  \csvcoli\textsuperscript{\csvcolviii} & 
  \num[round-mode=places,round-precision=3]{\csvcolii} \scriptsize(\num[round-mode=places,round-precision=3]{\csvcoliii}--\num[round-mode=places,round-precision=3]{\csvcoliv}) &
  \num[round-mode=places,round-precision=3]{\csvcolv} \scriptsize(\num[round-mode=places,round-precision=3]{\csvcolvi}--\num[round-mode=places,round-precision=3]{\csvcolvii})
}
\bottomrule
\end{tabular}
\end{table}

\begin{table}[htbp]
\centering
\caption{Held-out performance (in terms of \textbf{specificity at global 95\% specificity}) of CXR models across sub-populations stratified by the presence of label-relevant terms in prior discharge summaries. Statistically significant differences (as evaluated by bootstrapping) are highlighted with an asterisk (*) next to the label.}
\label{tab:prior_mention_spec_global}
\footnotesize 
\setlength{\tabcolsep}{18pt} 
\begin{tabular}{@{}l cc@{}}
\toprule
\textbf{Label} & \textbf{Previous Mentions} & \textbf{No Previous Mentions} \\
\midrule
\begin{filecontents*}{csv/prior_mention/prior_mention_Spec_at_Global_Spec95.csv}
Label,Previous Mentions_Mean,Previous Mentions_CI_Lower,Previous Mentions_CI_Upper,No Previous Mentions_Mean,No Previous Mentions_CI_Lower,No Previous Mentions_CI_Upper,sig
Atelectasis,0.931391093872479,0.9259796142874984,0.936943569087965,0.963796011945292,0.9596899144751656,0.967771452175167,*
Cardiomegaly,0.9237970369495196,0.9179350086417122,0.9294394593515328,0.9684029501515268,0.964467079941797,0.972433096173862,*
Consolidation,0.9421035665861708,0.9362276906717832,0.9476267722489972,0.9551120328498208,0.9515565247112074,0.9588984044365804,*
Edema,0.947501075122425,0.946480396335638,0.948663948851208,0.97469022419031,0.9634780563165892,0.9845183486238532,*
Enlarged Cardiomediastinum,0.9479030327234644,0.940534326379015,0.9548346466225778,0.950830829607352,0.9480802742748576,0.953803372267854,
Fracture,0.9355057233119678,0.929080424132858,0.9419379349763886,0.9577191110965084,0.9542947837511188,0.9612532286647184,*
Lung Lesion,0.928729618932085,0.924175845925497,0.9333609237412192,0.96881860467617,0.9647259804877722,0.972757910503651,*
Lung Opacity,0.9312408112210542,0.9267289112099832,0.9358898452743544,0.9690589110127606,0.9643745486416216,0.9735575180654672,*
Pleural Effusion,0.9406978081365808,0.9375799695578424,0.9439562713497512,0.9677440788595576,0.9615431401551452,0.9736261646263652,*
Pleural Other,0.9084535142275152,0.8832295758650115,0.9315993040521192,0.9522767898252352,0.9510048486958056,0.9536696929595908,*
Pneumonia,0.9425084271542638,0.9380428214245248,0.9468257343688522,0.9580341331259368,0.9533678218418256,0.9628968128731598,*
Pneumothorax,0.9482890526263668,0.943638949985214,0.952913668408158,0.9515467413593708,0.9473289502100096,0.955806620152298,
Support Devices,0.9454274890215189,0.9442489981358204,0.9467571106460846,0.981047247460208,0.9721797696593156,0.988931721516493,*
\end{filecontents*}
\csvreader[
  head=true, 
  late after line=\\
]{csv/prior_mention/prior_mention_Spec_at_Global_Spec95.csv}{}{%
  \csvcoli\textsuperscript{\csvcolviii} & 
  \num[round-mode=places,round-precision=3]{\csvcolii} \scriptsize(\num[round-mode=places,round-precision=3]{\csvcoliii}--\num[round-mode=places,round-precision=3]{\csvcoliv}) &
  \num[round-mode=places,round-precision=3]{\csvcolv} \scriptsize(\num[round-mode=places,round-precision=3]{\csvcolvi}--\num[round-mode=places,round-precision=3]{\csvcolvii})
}
\bottomrule
\end{tabular}
\end{table}

\begin{table}[htbp]
\centering
\caption{Comparison of DenseNet121 AUROC across Standard, Reweighted (IPW), and Matched Neighbor settings. Metrics reported are mean (95\% CI). Statistically significant differences in the reweighted/matched sets compared to the standard are bolded.}
\label{tab:reweighted_matched}

\footnotesize
\setlength{\tabcolsep}{12pt}
\begin{tabular}{@{}l ccc@{}}
\toprule
\textbf{Label} & \textbf{Standard} & \textbf{Reweighted (IPW)} & \textbf{Matched} \\
\midrule
Atelectasis & 0.801 \scriptsize(0.793--0.808) & \textbf{0.775 \scriptsize(0.766--0.784)} & \textbf{0.695 \scriptsize(0.684--0.706)} \\
Cardiomegaly & 0.811 \scriptsize(0.803--0.818) & \textbf{0.751 \scriptsize(0.741--0.760)} & \textbf{0.666 \scriptsize(0.656--0.677)} \\
Consolidation & 0.818 \scriptsize(0.805--0.832) & 0.798 \scriptsize(0.782--0.815) & \textbf{0.641 \scriptsize(0.617--0.665)} \\
Edema & 0.891 \scriptsize(0.885--0.898) & \textbf{0.861 \scriptsize(0.850--0.871)} & \textbf{0.749 \scriptsize(0.735--0.763)} \\
Enlarged Cardiomediastinum & 0.707 \scriptsize(0.685--0.730) & 0.704 \scriptsize(0.681--0.727) & \textbf{0.589 \scriptsize(0.561--0.618)} \\
Fracture & 0.681 \scriptsize(0.651--0.710) & 0.655 \scriptsize(0.621--0.689) & \textbf{0.606 \scriptsize(0.569--0.643)} \\
Lung Lesion & 0.756 \scriptsize(0.737--0.775) & 0.720 \scriptsize(0.697--0.742) & \textbf{0.598 \scriptsize(0.570--0.626)} \\
Lung Opacity & 0.749 \scriptsize(0.741--0.757) & \textbf{0.721 \scriptsize(0.712--0.730)} & \textbf{0.665 \scriptsize(0.655--0.676)} \\
Pleural Effusion & 0.923 \scriptsize(0.919--0.927) & \textbf{0.894 \scriptsize(0.888--0.900)} & \textbf{0.852 \scriptsize(0.845--0.860)} \\
Pleural Other & 0.806 \scriptsize(0.778--0.834) & 0.768 \scriptsize(0.734--0.803) & \textbf{0.618 \scriptsize(0.572--0.663)} \\
Pneumonia & 0.707 \scriptsize(0.693--0.721) & 0.692 \scriptsize(0.677--0.707) & \textbf{0.624 \scriptsize(0.606--0.642)} \\
Pneumothorax & 0.879 \scriptsize(0.867--0.891) & 0.857 \scriptsize(0.841--0.872) & \textbf{0.722 \scriptsize(0.697--0.747)} \\
Support Devices & 0.906 \scriptsize(0.901--0.911) & \textbf{0.885 \scriptsize(0.879--0.892)} & \textbf{0.833 \scriptsize(0.825--0.841)} \\
\bottomrule
\end{tabular}
\label{tab:matched_neighbor_auroc}
\end{table}

\begin{table}[htbp]
\centering
\caption{Comparison of ResNet50 AUROC across Standard, Reweighted (IPW), and Matched Neighbor settings. Metrics reported are mean (95\% CI). Statistically significant differences in the reweighted/matched sets compared to the standard are bolded.}
\label{tab:reweighted_matched_updated}

\footnotesize
\setlength{\tabcolsep}{12pt}
\begin{tabular}{@{}l ccc@{}}
\toprule
\textbf{Label} & \textbf{Standard} & \textbf{Reweighted (IPW)} & \textbf{Matched} \\
\midrule
Atelectasis  & 0.801 \scriptsize(0.793--0.808)  & \textbf{0.712 \scriptsize(0.698--0.725)}  & \textbf{0.632 \scriptsize(0.615--0.650)}  \\
Cardiomegaly  & 0.811 \scriptsize(0.803--0.818)  & \textbf{0.689 \scriptsize(0.674--0.704)}  & \textbf{0.597 \scriptsize(0.580--0.614)}  \\
Consolidation  & 0.818 \scriptsize(0.805--0.832)  & \textbf{0.704 \scriptsize(0.674--0.733)}  & \textbf{0.574 \scriptsize(0.537--0.613)}  \\
Edema  & 0.891 \scriptsize(0.885--0.898)  & \textbf{0.810 \scriptsize(0.788--0.829)}  & \textbf{0.694 \scriptsize(0.672--0.717)}  \\
Enlarged Cardiomediastinum  & 0.707 \scriptsize(0.685--0.730)  & 0.660 \scriptsize(0.624--0.696)  & \textbf{0.565 \scriptsize(0.519--0.612)}  \\
Fracture  & 0.681 \scriptsize(0.651--0.710)  & 0.635 \scriptsize(0.589--0.682)  & \textbf{0.570 \scriptsize(0.514--0.626)}  \\
Lung Lesion  & 0.756 \scriptsize(0.737--0.775)  & \textbf{0.602 \scriptsize(0.561--0.639)}  & \textbf{0.536 \scriptsize(0.488--0.580)}  \\
Lung Opacity  & 0.749 \scriptsize(0.741--0.757)  & \textbf{0.644 \scriptsize(0.631--0.658)}  & \textbf{0.593 \scriptsize(0.576--0.609)}  \\
Pleural Effusion  & 0.923 \scriptsize(0.919--0.927)  & \textbf{0.798 \scriptsize(0.785--0.810)}  & \textbf{0.739 \scriptsize(0.725--0.753)}  \\
Pleural Other  & 0.806 \scriptsize(0.778--0.834)  & \textbf{0.689 \scriptsize(0.631--0.748)}  & \textbf{0.597 \scriptsize(0.519--0.674)}  \\
Pneumonia  & 0.707 \scriptsize(0.693--0.721)  & \textbf{0.603 \scriptsize(0.579--0.626)}  & \textbf{0.554 \scriptsize(0.526--0.582)}  \\
Pneumothorax  & 0.879 \scriptsize(0.867--0.891)  & \textbf{0.764 \scriptsize(0.737--0.790)}  & \textbf{0.652 \scriptsize(0.613--0.691)}  \\
Support Devices  & 0.906 \scriptsize(0.901--0.911)  & \textbf{0.825 \scriptsize(0.813--0.837)}  & \textbf{0.763 \scriptsize(0.749--0.777)}  \\
\bottomrule
\end{tabular}
\label{tab:resnet50_ipw}
\end{table}

\begin{table}[htbp]
\centering
\caption{Comparison of Sensitivity (@Global 95\% Spec.) across Standard, Reweighted (IPW), and Matched Neighbor settings. Metrics reported are mean (95\% CI). Statistically significant differences as compared to the standard set are highlighted in bold.}
\label{tab:sens_results}
\footnotesize 
\setlength{\tabcolsep}{10pt}
\begin{tabular}{@{}l ccc@{}}
\toprule
\textbf{Label} & \textbf{Standard} & \textbf{Reweighted (IPW)} & \textbf{Matched} \\
\midrule
Atelectasis & 0.264 \scriptsize(0.235--0.292) & 0.232 \scriptsize(0.204--0.260) & \textbf{0.147 \scriptsize(0.118--0.178)} \\
Cardiomegaly & 0.638 \scriptsize(0.609--0.665) & \textbf{0.444 \scriptsize(0.413--0.476)} & \textbf{0.380 \scriptsize(0.345--0.413)} \\
Consolidation & 0.167 \scriptsize(0.133--0.203) & 0.158 \scriptsize(0.125--0.193) & \textbf{0.046 \scriptsize(0.024--0.071)} \\
Edema & 0.505 \scriptsize(0.472--0.537) & \textbf{0.382 \scriptsize(0.350--0.414)} & \textbf{0.258 \scriptsize(0.222--0.294)} \\
Enlarged Cardiomediastinum & 0.052 \scriptsize(0.038--0.068) & 0.055 \scriptsize(0.040--0.071) & \textbf{0.016 \scriptsize(0.007--0.027)} \\
Fracture & 0.049 \scriptsize(0.034--0.065) & 0.066 \scriptsize(0.049--0.084) & \textbf{0.008 \scriptsize(0.001--0.018)} \\
Lung Lesion & 0.063 \scriptsize(0.045--0.083) & 0.040 \scriptsize(0.025--0.057) & \textbf{0.005 \scriptsize(0.000--0.014)} \\
Lung Opacity & 0.264 \scriptsize(0.244--0.284) & 0.237 \scriptsize(0.217--0.257) & \textbf{0.149 \scriptsize(0.130--0.168)} \\
No Finding & 0.306 \scriptsize(0.288--0.324) & 0.323 \scriptsize(0.305--0.341) & \textbf{0.210 \scriptsize(0.187--0.234)} \\
Pleural Effusion & 0.608 \scriptsize(0.576--0.639) & \textbf{0.488 \scriptsize(0.443--0.531)} & \textbf{0.393 \scriptsize(0.345--0.442)} \\
Pleural Other & 0.303 \scriptsize(0.213--0.401) & 0.223 \scriptsize(0.136--0.323) & \textbf{0.096 \scriptsize(0.033--0.179)} \\
Pneumonia & 0.136 \scriptsize(0.111--0.162) & 0.118 \scriptsize(0.093--0.142) & \textbf{0.040 \scriptsize(0.023--0.059)} \\
Pneumothorax & 0.213 \scriptsize(0.171--0.257) & 0.147 \scriptsize(0.110--0.187) & \textbf{0.058 \scriptsize(0.033--0.086)} \\
Support Devices & 0.707 \scriptsize(0.687--0.727) & \textbf{0.627 \scriptsize(0.603--0.650)} & \textbf{0.490 \scriptsize(0.461--0.518)} \\
\bottomrule
\end{tabular}%
\end{table}

\begin{table}[htbp]
\centering
\caption{Comparison of Specificity across Standard, Reweighted (IPW), and Matched Neighbor settings. Metrics reported are mean (95\% CI). Statistically significant differences as compared to the standard set are highlighted in bold.}
\label{tab:spec_results}
\footnotesize 
\setlength{\tabcolsep}{10pt}
\begin{tabular}{@{}l ccc@{}}
\toprule
\textbf{Label} & \textbf{Standard} & \textbf{Reweighted (IPW)} & \textbf{Matched} \\
\midrule
Atelectasis & 0.950 \scriptsize(0.950--0.950) & 0.950 \scriptsize(0.950--0.950) & 0.950 \scriptsize(0.950--0.951) \\
Cardiomegaly & 0.950 \scriptsize(0.950--0.950) & 0.950 \scriptsize(0.950--0.950) & 0.950 \scriptsize(0.950--0.950) \\
Consolidation & 0.950 \scriptsize(0.950--0.950) & 0.950 \scriptsize(0.950--0.950) & 0.950 \scriptsize(0.950--0.951) \\
Edema & 0.950 \scriptsize(0.950--0.950) & 0.950 \scriptsize(0.950--0.950) & 0.950 \scriptsize(0.950--0.951) \\
Enlarged Cardiomediastinum & 0.950 \scriptsize(0.950--0.950) & 0.950 \scriptsize(0.950--0.950) & 0.950 \scriptsize(0.949--0.951) \\
Fracture & 0.950 \scriptsize(0.950--0.950) & 0.950 \scriptsize(0.950--0.950) & 0.950 \scriptsize(0.948--0.952) \\
Lung Lesion & 0.950 \scriptsize(0.950--0.950) & 0.950 \scriptsize(0.950--0.950) & 0.950 \scriptsize(0.949--0.951) \\
Lung Opacity & 0.950 \scriptsize(0.950--0.950) & 0.950 \scriptsize(0.950--0.950) & 0.950 \scriptsize(0.950--0.951) \\
No Finding & 0.950 \scriptsize(0.950--0.950) & 0.950 \scriptsize(0.950--0.950) & 0.950 \scriptsize(0.950--0.951) \\
Pleural Effusion & 0.950 \scriptsize(0.950--0.950) & 0.950 \scriptsize(0.950--0.950) & 0.950 \scriptsize(0.950--0.951) \\
Pleural Other & 0.950 \scriptsize(0.950--0.950) & 0.950 \scriptsize(0.950--0.950) & 0.949 \scriptsize(0.939--0.962) \\
Pneumonia & 0.950 \scriptsize(0.950--0.950) & 0.950 \scriptsize(0.950--0.950) & 0.950 \scriptsize(0.948--0.951) \\
Pneumothorax & 0.950 \scriptsize(0.950--0.950) & 0.950 \scriptsize(0.950--0.950) & 0.950 \scriptsize(0.948--0.951) \\
Support Devices & 0.950 \scriptsize(0.950--0.950) & 0.950 \scriptsize(0.950--0.950) & 0.950 \scriptsize(0.950--0.951) \\
\bottomrule
\end{tabular}%
\end{table}

\begin{table}[htbp]
\centering
\caption{Comparison of Adaptive Calibration Error (ACE) across Standard, Reweighted (IPW), and Matched Neighbor settings. Metrics reported are mean (95\% CI).}
\label{tab:ace_results}
\footnotesize 
\setlength{\tabcolsep}{10pt}
\begin{tabular}{@{}l ccc@{}}
\toprule
\textbf{Label} & \textbf{Standard} & \textbf{Reweighted (IPW)} & \textbf{Matched} \\
\midrule
Atelectasis & 0.017 \scriptsize(0.014--0.021) & 0.024 \scriptsize(0.018--0.030) & \textbf{0.245 \scriptsize(0.239--0.251)} \\
Cardiomegaly & 0.017 \scriptsize(0.014--0.021) & \textbf{0.033 \scriptsize(0.026--0.039)} & \textbf{0.170 \scriptsize(0.165--0.175)} \\
Consolidation & 0.009 \scriptsize(0.007--0.011) & \textbf{0.015 \scriptsize(0.012--0.017)} & \textbf{0.158 \scriptsize(0.154--0.163)} \\
Edema & 0.012 \scriptsize(0.010--0.015) & \textbf{0.025 \scriptsize(0.021--0.029)} & \textbf{0.178 \scriptsize(0.174--0.183)} \\
Enlarged Cardiomediastinum & 0.020 \scriptsize(0.015--0.025) & 0.019 \scriptsize(0.014--0.024) & \textbf{0.354 \scriptsize(0.347--0.362)} \\
Fracture & 0.010 \scriptsize(0.007--0.013) & 0.015 \scriptsize(0.011--0.019) & \textbf{0.238 \scriptsize(0.233--0.244)} \\
Lung Lesion & 0.011 \scriptsize(0.008--0.015) & 0.010 \scriptsize(0.007--0.013) & \textbf{0.171 \scriptsize(0.167--0.176)} \\
Lung Opacity & 0.020 \scriptsize(0.016--0.023) & 0.022 \scriptsize(0.018--0.027) & \textbf{0.261 \scriptsize(0.257--0.266)} \\
No Finding & 0.020 \scriptsize(0.014--0.026) & 0.029 \scriptsize(0.020--0.038) & \textbf{0.444 \scriptsize(0.437--0.450)} \\
Pleural Effusion & 0.011 \scriptsize(0.006--0.015) & \textbf{0.025 \scriptsize(0.018--0.033)} & \textbf{0.098 \scriptsize(0.090--0.106)} \\
Pleural Other & 0.002 \scriptsize(0.001--0.003) & 0.003 \scriptsize(0.002--0.004) & \textbf{0.062 \scriptsize(0.059--0.065)} \\
Pneumonia & 0.015 \scriptsize(0.012--0.018) & 0.016 \scriptsize(0.013--0.019) & \textbf{0.222 \scriptsize(0.218--0.226)} \\
Pneumothorax & 0.011 \scriptsize(0.008--0.015) & 0.009 \scriptsize(0.006--0.012) & \textbf{0.155 \scriptsize(0.151--0.159)} \\
Support Devices & 0.014 \scriptsize(0.010--0.017) & 0.023 \scriptsize(0.017--0.029) & \textbf{0.138 \scriptsize(0.134--0.143)} \\
\bottomrule
\end{tabular}%
\end{table}

\begin{table}[htbp]
\centering
\caption{Comparison of Brier Skill Score (BSS) across Standard, Reweighted (IPW), and Matched Neighbor settings. Metrics reported are mean (95\% CI).}
\label{tab:bss_results}
\footnotesize 
\setlength{\tabcolsep}{10pt}
\begin{tabular}{@{}l ccc@{}}
\toprule
\textbf{Label} & \textbf{Standard} & \textbf{Reweighted (IPW)} & \textbf{Matched} \\
\midrule
Atelectasis & 0.180 \scriptsize(0.163--0.197) & 0.147 \scriptsize(0.128--0.165) & \textbf{-0.137 \scriptsize(-0.161---0.114)} \\
Cardiomegaly & 0.490 \scriptsize(0.468--0.510) & \textbf{0.292 \scriptsize(0.264--0.320)} & \textbf{0.003 \scriptsize(-0.030--0.035)} \\
Consolidation & 0.156 \scriptsize(0.125--0.190) & 0.131 \scriptsize(0.098--0.165) & \textbf{-0.279 \scriptsize(-0.315---0.245)} \\
Edema & 0.359 \scriptsize(0.334--0.383) & \textbf{0.226 \scriptsize(0.198--0.254)} & \textbf{-0.049 \scriptsize(-0.081---0.016)} \\
Enlarged Cardiomediastinum & 0.054 \scriptsize(0.029--0.078) & 0.047 \scriptsize(0.023--0.071) & \textbf{-0.669 \scriptsize(-0.710---0.627)} \\
Fracture & 0.094 \scriptsize(0.061--0.128) & 0.081 \scriptsize(0.050--0.113) & \textbf{-0.366 \scriptsize(-0.395---0.335)} \\
Lung Lesion & 0.047 \scriptsize(0.022--0.074) & 0.033 \scriptsize(0.007--0.059) & \textbf{-0.229 \scriptsize(-0.258---0.200)} \\
Lung Opacity & 0.210 \scriptsize(0.195--0.224) & 0.183 \scriptsize(0.167--0.199) & \textbf{-0.216 \scriptsize(-0.244---0.189)} \\
No Finding & 0.207 \scriptsize(0.187--0.226) & 0.212 \scriptsize(0.192--0.231) & \textbf{-1.218 \scriptsize(-1.267---1.171)} \\
Pleural Effusion & 0.492 \scriptsize(0.473--0.513) & \textbf{0.393 \scriptsize(0.363--0.421)} & \textbf{0.324 \scriptsize(0.296--0.353)} \\
Pleural Other & 0.054 \scriptsize(0.019--0.098) & 0.047 \scriptsize(0.009--0.094) & \textbf{-0.163 \scriptsize(-0.213---0.122)} \\
Pneumonia & 0.111 \scriptsize(0.086--0.136) & 0.099 \scriptsize(0.076--0.122) & \textbf{-0.400 \scriptsize(-0.436---0.365)} \\
Pneumothorax & 0.191 \scriptsize(0.150--0.233) & 0.146 \scriptsize(0.106--0.188) & \textbf{-0.220 \scriptsize(-0.264---0.178)} \\
Support Devices & 0.584 \scriptsize(0.569--0.598) & \textbf{0.498 \scriptsize(0.479--0.517)} & \textbf{0.145 \scriptsize(0.121--0.169)} \\
\bottomrule
\end{tabular}%
\end{table}

\clearpage
\section{Mathematical Proofs}
\label{sec:proofs}

To rigorously assess whether the vision model utilizes visual signals independent of clinical context, we define a target distribution $Q$ where the diagnostic label $Y$ is statistically independent of the prior clinical context $C$ ($Y \perp C$). This is meant to represent a setting where the clinical history does not provide discriminative information about the label. We estimate the model's performance on this target distribution using Inverse Probability Weighting (IPW). By reweighting our observed evaluation set using the ratio of marginal disease prevalence to the context-conditional pre-test probability, we remove the correlation between context and label. 

\subsection{Formal Derivation of the Inverse Probability Metric Weights}\label{app:proof}
Let $P(X, Y, C)$ denote the observed data distribution, where $X$ is the image, $Y \in \{0, 1\}$ is the diagnostic label, and $C$ is the prior clinical context. In the observational setting, $Y$ and $C$ are correlated. We define a target distribution $Q(X, Y, C)$ that satisfies two conditions:
\begin{enumerate}
\item \textbf{Independence of Context and Label:} $Q(Y, C) = Q(Y)Q(C) = P(Y)P(C)$. This represents a scenario where clinical history provides no information about the current diagnosis.
\item \textbf{Invariance of Imaging Mechanism:} $Q(X \mid Y, C) = P(X \mid Y, C)$. The physical process generating the X-ray from the patient's state remains unchanged.\end{enumerate}

We seek to evaluate performance metrics defined under $Q$ using data sampled from $P$. The importance weight $w(x, y, c)$ is given by:
\begin{equation}
w(x,y,c) = \frac{Q(X \mid Y, C)Q(Y)Q(C)}{P(X \mid Y, C)P(Y \mid C)P(C)} = \frac{P(Y)P(C)}{P(Y \mid C)P(C)} = \frac{P(Y)}{P(Y \mid C)}
\end{equation}
Under the assumption that $Q(X|Y,C) = P(X|Y,C)$.
Thus, the weight is the ratio of the marginal label prevalence to the conditional pre-test probability.

\begin{restatable}{lemma}{Reweighting}
\label{lemma:reweighting}
Let $P$ be defined as the observed distribution ,and $Q$ refer to the target distribution. where $Q(X, Y, C) > 0 \implies P(X, Y, C) > 0$. Then for any measurable function $g(X, Y, C)$,
\begin{equation*}
E_P[w(X, Y, C) g(X, Y, C)] = E_Q[g(X, Y, C)]
\end{equation*}
\end{restatable}
\begin{proof} Using $p, q$ to denote the PDF / PMF of $P$ and $Q$ respectively,
\begin{align*}
E_Q[g(X, Y, X)] &= \int_{x, y, c} g(x, y, c) q(x, y, c) dx dy dc \\
&= \int_{x, y, c} g(x, y, c) p(x, y, c) \frac{q(x, y, c)}{p(x, y, c)} dx dy dc \\
&= \int_{x, y, c} w(x, y, c) g(x, y, c) p(x, y, c) dx dy dc \\
&= E_P[w(X, Y, C) g(X, Y, C)]
\end{align*}
\end{proof}

\subsection{Weighted Sensitivity and Specificity}
We define the Sensitivity (True Positive Rate) and Specificity (True Negative Rate) of a classifier $f(X)$ at a decision threshold $\tau$ as conditional probabilities under the target distribution $Q$. We utilize the importance sampling identity $\mathbb{E}_Q[g(X, Y, C)] = \mathbb{E}_P[w(X, Y, C) \cdot g(X, Y, C)]$ to express these metrics in terms of the observed distribution $P$.

\subsubsection{Sensitivity (True Positive Rate)}
\begin{proof}
Sensitivity under a distribution $Q$, denoted $\text{Sens}_Q(\tau)$, is defined as the conditional probability of a positive prediction given a positive label:
\begin{equation}
\text{Sens}_Q(\tau) = P_Q(f(X) > \tau \mid Y = 1) = \frac{\mathbb{E}_Q[\mathbb{I}(f(X) > \tau) \cdot \mathbb{I}(Y = 1)]}{\mathbb{E}_Q[\mathbb{I}(Y=1)]}
\end{equation}
We apply lemma \cref{lemma:reweighting} to both the numerator and the denominator. For the numerator, we set $g(X, Y, C) = \mathbb{I}(f(X) > \tau) \cdot \mathbb{I}(Y = 1)$, and for the denominator, we set $g(X, Y, C) = \mathbb{I}(Y = 1)$. This yields:
\begin{equation}
\text{Sens}_Q(\tau) = \frac{\mathbb{E}_P \left[ w(X,Y,C) \cdot \mathbb{I}(f(X) > \tau) \cdot \mathbb{I}(Y = 1) \right]}{\mathbb{E}_P \left[ w(X,Y,C) \cdot \mathbb{I}(Y = 1) \right]}
\end{equation}
\end{proof}

\subsubsection{Specificity (True Negative Rate)}
Similarly, the Specificity in the target distribution, denoted $\text{Spec}_Q(\tau)$, is defined as:
\begin{equation}
\text{Spec}_Q(\tau) = P_Q(f(X) \leq \tau \mid Y = 0) = \frac{\mathbb{E}_Q[\mathbb{I}(f(X) \leq \tau) \cdot \mathbb{I}(Y = 0)]}{\mathbb{E}_Q[\mathbb{I}(Y=0)]}
\end{equation}
Applying the same lemma \cref{lemma:reweighting} to the numerator (with $g = \mathbb{I}(f(X) \leq \tau) \cdot \mathbb{I}(Y = 0)$) and the denominator (with $g = \mathbb{I}(Y = 0)$), we express Specificity in terms of the observed distribution $P$:
\begin{equation}
\text{Spec}_Q(\tau) = \frac{\mathbb{E}_P \left[ w(X,Y,C) \cdot \mathbb{I}(f(X) \leq \tau) \cdot \mathbb{I}(Y = 0) \right]}{\mathbb{E}_P \left[ w(X,Y,C) \cdot \mathbb{I}(Y = 0) \right]}
\end{equation}

\subsection{Area Under the Curve (AUC)}\label{app:auroc_proof}
We define the AUROC in the target distribution, denoted $\tau_Q$, as the probability that the classifier $f$ ranks a randomly sampled positive instance higher than a randomly sampled negative instance. We consider two independent samples $(X, Y, C)$ and $(X', Y', C')$ drawn from the target joint distribution $Q$.
\\
\begin{proof}
Formally, $\tau_Q$ is the conditional probability of a correct ranking given that the first sample is positive ($Y=1$) and the second is negative ($Y'=0$). We express this as a ratio of expectations over the independent joint distributions:
\begin{equation}
\tau_Q = P_{Q}(f(X) > f(X') \mid Y = 1, Y' = 0) = \frac{\mathbb{E}_{Q, Q'} \left[ \mathbb{I}(f(X) > f(X')) \cdot \mathbb{I}(Y=1) \cdot \mathbb{I}(Y'=0) \right]}{\mathbb{E}_{Q, Q'} \left[ \mathbb{I}(Y=1) \cdot \mathbb{I}(Y'=0) \right]}
\end{equation}
We apply the reweighting identity from \cref{lemma:reweighting} to the independent samples in both the numerator and the denominator. For the numerator, we define the function $g$ over the pair of samples as $g(\cdot) = \mathbb{I}(f(X) > f(X')) \cdot \mathbb{I}(Y=1) \cdot \mathbb{I}(Y'=0)$. The weight for the pair is the product of individual weights $w(X, Y, C) \cdot w(X', Y', C')$. This yields:
\begin{equation}
\tau_Q = \frac{\mathbb{E}_{P, P'} \left[ w(X, Y, C) w(X', Y', C') \cdot \mathbb{I}(f(X) > f(X')) \cdot \mathbb{I}(Y=1) \mathbb{I}(Y'=0) \right]}{\mathbb{E}_{P, P'} \left[ w(X, Y, C) w(X', Y', C') \cdot \mathbb{I}(Y=1) \mathbb{I}(Y'=0) \right]}
\end{equation}
The denominator factors into the product of the weighted marginals for each class. Thus, we can estimate the target AUROC using pairs sampled from the observed distribution $P$:
\begin{equation}
\tau_Q = \frac{\mathbb{E}_{P, P'} \left[ w(X, Y, C) w(X', Y', C') \cdot \mathbb{I}(f(X) > f(X')) \cdot \mathbb{I}(Y=1) \mathbb{I}(Y'=0) \right]}{\mathbb{E}_{P} \left[ w(X, Y, C) \cdot \mathbb{I}(Y=1) \right] \cdot \mathbb{E}_{P} \left[ w(X', Y', C') \cdot \mathbb{I}(Y'=0) \right]}
\end{equation}
\end{proof}
\subsection{Adaptive Calibration Error (ACE)}
We define calibration error under the target distribution $Q$. Let predictions be partitioned into $K$ disjoint bins $\{B_1, \dots, B_K\}$. In the adaptive setting, these bins are chosen such that each bin contains an equal amount of probability mass under $Q$:
\begin{equation}
Q(X \in B_k) = \frac{1}{K} \quad \forall k \in {1, \dots, K}
\end{equation}
Applying importance sampling, this constraint implies that the bins must satisfy an equal weight condition in the observed distribution $P$:
\begin{equation}
\mathbb{E}_P [ w(Y,C) \cdot \mathbb{I}(X \in B_k) ] = \frac{1}{K} \mathbb{E}_P [ w(Y,C) ]
\end{equation}
Within each bin $B_k$, the Calibration Error is the difference between the expected confidence and the expected label under $Q$:
\begin{equation}
\text{Error}_k = \left| \mathbb{E}_Q[Y \mid X \in B_k] - \mathbb{E}_Q[f(X) \mid X \in B_k] \right|
\end{equation}
Expanding the conditional expectations using weighted estimators from $P$:
\begin{equation}
\mathbb{E}_Q[Y \mid X \in B_k] = \frac{\mathbb{E}_Q[Y \cdot \mathbb{I}(X \in B_k)]}{Q(X \in B_k)} = \frac{\mathbb{E}_P[w(Y,C) \cdot Y \cdot \mathbb{I}(X \in B_k)]}{\mathbb{E}_P[w(Y,C) \cdot \mathbb{I}(X \in B_k)]}
\end{equation}
Similarly for the prediction $f(X)$:
\begin{equation}
\mathbb{E}_Q[f(X) \mid X \in B_k] = \frac{\mathbb{E}_P[w(Y,C) \cdot f(X) \cdot \mathbb{I}(X \in B_k)]}{\mathbb{E}_P[w(Y,C) \cdot \mathbb{I}(X \in B_k)]}
\end{equation}
The overall Adaptive Calibration Error (ACE) under $Q$ is the expected error across all bins:
\begin{equation}
\text{ACE}_Q = \sum_{k=1}^K Q(X \in B_k) \cdot \text{Error}_k = \frac{1}{K} \sum_{k=1}^K \left| \frac{\mathbb{E}_P[w \cdot (Y - f(X)) \cdot \mathbb{I}(X \in B_k)]}{\mathbb{E}_P[w \cdot \mathbb{I}(X \in B_k)]} \right|
\end{equation}

\subsection{Brier Skill Score (BSS)}
We define the Brier Score (BS) under the target distribution $Q$ as the expected mean squared error between the predicted probability $f(X)$ and the true label $Y$.
\begin{equation}
BS_Q = \mathbb{E}_Q \left[ (Y - f(X))^2 \right]
\end{equation}
Applying the importance sampling identity $\mathbb{E}_Q[\cdot] = \mathbb{E}_P[w \cdot]$, we express this in terms of the observed distribution $P$:
\begin{equation}
BS_Q = \mathbb{E}_P \left[ w(Y,C) \cdot (Y - f(X))^2 \right]
\end{equation}
The Brier Skill Score requires a reference baseline, $BS_{\text{ref}}$, corresponding to a "no-skill" classifier that we chose to simply predict the marginal prevalence of the target distribution, $Q(Y=1)$. Since our target distribution preserves the marginal prevalence of the observed distribution ($Q(Y)=P(Y)$), this baseline prediction is simply $\pi = P(Y=1)$.
\begin{equation}
BS_{\text{ref, Q}} = \mathbb{E}_Q \left[ (Y - \pi)^2 \right] = \mathbb{E}_P \left[ w(Y,C) \cdot (Y - \pi)^2 \right]
\end{equation}
The Weighted Brier Skill Score is therefore defined as:
\begin{equation}BSS_Q = 1 - \frac{BS_Q}{BS_{\text{ref}, Q}} = 1 - \frac{\mathbb{E}_P \left[ w(Y,C) \cdot (Y - f(X))^2 \right]}{\mathbb{E}_P \left[ w(Y,C) \cdot (Y - P(Y=1))^2 \right]}
\end{equation}

\end{document}